\documentclass[journal,final]{IEEEtran}

\usepackage{amsmath}
\usepackage{amssymb}
\usepackage{graphicx}
\usepackage{multirow}
\usepackage[T1]{fontenc}
\usepackage{url}
\usepackage{cite}
\usepackage{mdwmath}
\usepackage{mdwtab}
\usepackage{comment}
\usepackage{enumerate}
\usepackage{scrextend}
\usepackage{subcaption}
\usepackage{color,soul}
\usepackage{hhline}
\usepackage{placeins}

\DeclareMathOperator*{\argmin}{arg\,min}
\let\vec\relax
\DeclareMathOperator*{\vec}{vec}
\DeclareMathOperator*{\sign}{sign}

\newcommand{\revised}[1]{#1} 

\newcommand{\removed}[1]{}

\begin{document}

\title{Robust Photometric Stereo via Dictionary Learning}

\author{Andrew~J.~Wagenmaker, \emph{Student Member, IEEE}, Brian~E.~Moore, \emph{Member, IEEE},\\ and Raj~Rao~Nadakuditi, \emph{Member, IEEE}
\thanks{This work was supported in part by the following grants: ONR grant N00014-15-1-2141, DARPA Young Faculty Award D14AP00086, and ARO MURI grants W911NF-11-1-0391 and 2015-05174-05.}
\thanks{A.~J.~Wagenmaker, B.~E.~Moore, and R.~R.~Nadakuditi are with the Department of Electrical Engineering and Computer Science, University of Michigan, Ann Arbor, MI 48109 USA (email: ajwagen@umich.edu; brimoor@umich.edu; rajnrao@umich.edu).}
}

\markboth{IEEE TRANSACTIONS ON COMPUTATIONAL IMAGING}{}

\maketitle

\begin{abstract}
Photometric stereo is a method that seeks to reconstruct the normal vectors of an object from a set of images of the object illuminated under different light sources. While effective in some situations, classical photometric stereo relies on a diffuse surface model that cannot handle objects with complex reflectance patterns, and it is sensitive to non-idealities in the images. In this work, we propose a novel approach to photometric stereo that relies on dictionary learning to produce robust normal vector reconstructions. Specifically, we develop \revised{two} formulations for applying dictionary learning to photometric stereo. We \revised{propose} a model that applies dictionary learning to regularize and reconstruct the normal vectors from the images under the classic Lambertian reflectance model. \revised{We then generalize this} model to explicitly model non-Lambertian objects. We investigate \revised{both} approaches through extensive experimentation on synthetic and real benchmark datasets and observe state-of-the-art performance compared to existing robust photometric stereo methods.
\end{abstract}


\section{Introduction} \label{sec:intro}

\IEEEPARstart{P}{hotometric} stereo \cite{woodham1980} is a method that seeks to reconstruct the normal vectors of an object from a set of images of the object illuminated under different light sources. Concretely, we have images $I_1, \ldots, I_d$ of the three-dimensional object and, in each image, the object is illuminated by a (distant) light source with light incident on the object in directions $\ell_1, \ldots, \ell_d \in \mathbb{R}^3$. Given $I_1, \ldots, I_d$ and $\ell_1, \ldots, \ell_d$, the goal is to estimate the normal vector map of the object, which can then be numerically integrated to obtain a three-dimensional representation of the object. The appeal of photometric stereo is its simplicity: it requires only a camera and a movable light source to generate a three-dimensional representation.

\subsection{Background} \label{subsec:background}

Since its introduction by Woodham \cite{woodham1980}, significant work has been performed to increase the generality and robustness of photometric stereo \cite{barsky2003,sato2007,okabe2009,goldman2010,wu2010,chandraker2011,wu2011,ackermann2012,shi2012,ikehata2012,chandraker2013,ikehata2014journal,ikehata2014,shi2014}. \revised{This body of work typically seeks to weaken assumptions found in Woodham's original model. The so-called \textit{uncalibrated photometric stereo} problem attempts to weaken the assumption that the position of the object relative to the position of the light source is known and instead estimate the normal vectors of the object without any knowledge of the lighting directions \cite{hayakawa1994,belhumeur1999,yuille1999,georghiades2003,chandraker2005,alldrin2007,shi2010,favaro2012,wu2013,queau2015}. Alternatively, many works have sought to weaken the assumption that the object follows the Lambertian reflectance model. Works in this category generally attempt to either reconstruct the normal vectors of objects whose reflectance properties deviate from the Lambertian model, or they try to develop methods that are robust to corruptions in the observed images. In this work, our focus is primarily on the design of robust photometric stereo algorithms, but we also incorporate a non-Lambertian model for increased generality.}

The Lambertian reflectance model holds that the intensity of light reflected by a point on a surface is linearly proportional to the inner product of the direction of illumination and the normal vector of the surface at that point \cite{woodham1980}. Given a set of images of a Lambertian object illuminated under several (known) lighting directions, a simple system of equations can be solved to determine the normal vector at each point on the surface. In practice, while this is a reasonable model for some objects, the reflectance properties of many real-world objects differ significantly from the Lambertian model. Furthermore, shadows, specularities, and other non-idealities can cause additional deviations from linear reflectance. Performing classical photometric stereo on such non-Lambertian data typically yields large errors in the estimated normal vectors. As such, developing photometric stereo methods for objects that are inherently non-Lambertian and improving robustness to other imperfections in the data are essential to extending the applicability and accuracy of photometric stereo.

Two primary approaches have found success addressing these problems. Several works assume the Lambertian model is fundamentally correct and seek to account for deviations from the model through explicit outlier removal \cite{wu2011,ikehata2012}---often assuming that non-idealities are sparse. While achieving some level of success, these approaches can place overly restrictive assumptions on the data, which may result in falsely rejecting useful data as outliers, and they make no attempt to model the true reflectance properties of objects. In turn, other works propose more complex reflectance models that enable non-Lambertian photometric stereo \cite{ikehata2014journal,ikehata2014,shi2014}. Approaches in this class are able to accurately model a wider range of objects, but they still break down when their modeling assumptions fail. Furthermore, they often fail when the data contains corruptions not accounted for by their reflectance models. In addition to the aforementioned difficulties, state-of-the-art methods in both categories typically rely on a large number of images to accurately estimate the normal vectors, which may be infeasible to gather in practice.

\subsection{Contributions} \label{subsec:contributions}

In this work, we propose a novel approach to photometric stereo that relies on dictionary learning \cite{elad2006image,aharon2006rm} to robustly handle a wide range of non-idealities in the data. Dictionary learning seeks to represent local patches of the data as sparse with respect to a learned collection of atoms. Such models effectively act as dynamic regularization that adapts to the underlying structure of the data and removes spurious corruptions. Inspired by recent successes applying dictionary learning to a variety of imaging problems \cite{ravishankar2011mr,ravishankar2016low,ravishankar2016lassi,dinokat2016}, we adopt this methodology to improve the robustness of photometric stereo. Our approach is data-driven and adapts to the underlying structure of the data without imposing additional explicit constraints. Furthermore, we incorporate an existing non-Lambertian reflectance model into our method to better handle non-Lambertian surfaces. In total we present \revised{two} dictionary learning-based formulations of robust photometric stereo. We investigate the performance of \revised{both methods} in a variety of different scenarios. In particular, we evaluate their performance on the benchmark DiLiGenT dataset \cite{shi2016} and their ability to handle general, non-sparse corruptions.

This work is an extension of our recent work \cite{wagenmaker2017}. Here, we substantially build on the work by performing an extensive numerical study to evaluate the performance of our proposed dictionary learning-based approaches on real and synthetic data, and we extend our methods to incorporate the aforementioned non-Lambertian reflectance model.

\subsection{Organization} \label{subsec:organization}

The remainder of this work is organized as follows. In Section~\ref{sec:related}, we provide a brief overview of related works on photometric stereo. In Section~\ref{sec:prelims}, we carefully define the photometric stereo problem and the non-Lambertian reflectance model we will incorporate into our method. We present our proposed dictionary learning-based methods in Section~\ref{sec:dl}, and in Section~\ref{sec:dlalgos} we present the associated algorithms for solving them. Finally, Section~\ref{sec:experiments} provides an extensive numerical study of the performance of our proposed methods compared to state-of-the-art methods.

\section{Related Work} \label{sec:related}

Lambertian photometric stereo was originally proposed by Woodham \cite{woodham1980} in 1980. Since then, much work has been done extending it to more general settings where the Lambertian model does not hold exactly. This body of work has generally proceeded by either treating non-Lambertian effects as outliers or directly accounting for non-Lambertian effects in the reflectance model.

A variety of approaches have been proposed that perform robust photometric stereo via outlier rejection. In general, these methods assume that the data is inherently Lambertian, seek to isolate non-Lambertian effects as outliers, and then reject the outliers to increase the accuracy of the computed normal vectors. Early works in robust photometric stereo---typically referred to as \textit{4-source photometric stereo}---utilized four images to identify and reject specularities \cite{coleman1982,solomon1996,barsky2003}. Recently, more complex methods have been developed that rely on maximum likelihood estimation \cite{verbiest2008}, expectation maximization \cite{wu2010}, and a maximum feasible subsystem framework \cite{yu2010}. Other approaches include a graph cuts-based algorithm to identify shadows \cite{chandraker2007}, a method that seeks to map color images into a two-dimensional subspace invariant to specularities \cite{zickler2008}, and several methods that utilize RANSAC-based algorithms \cite{sunkavalli2010,mukaigawa2007} to detect and reject outliers.

The most recent works on robust photometric stereo via outlier rejection---and the current state-of-the-art in this area---are those by Wu et al. \cite{wu2011} and Ikehata et al. \cite{ikehata2012}. These works rely on the observation that images of a Lambertian object lie in a three-dimensional subspace. This observation, together with modeling non-Lambertian effects as sparse corruptions to the underlying Lambertian data, motivates the authors to propose rank minimization-based approaches, which are shown to effectively separate the Lambertian portion of the data from non-Lambertian effects.

Regardless of their robustness to outliers, approaches that rely on the Lambertian reflection model as the underlying model of the data are inherently limited in scope due to the wide variety of non-Lambertian surfaces that exist in the real world. As a result, another body of work has been incorporating more general reflectance models into photometric stereo \cite{oren1995}. For example, uncalibrated photometric stereo based on the Torrance and Sparrow reflectance model has been proposed \cite{georghiades2003} as well as calibrated photometric stereo based on the Ward reflectance model \cite{chung2008,goldman2010,ackermann2012}. 

A large amount of work has also been done developing photometric stereo algorithms that incorporate reflectance models based on general reflectance properties exhibited by materials. In particular, the property of isotropy has been successfully utilized in a variety of works \cite{alldrin2007_2,alldrin2008,higo2010,shi2012,chandraker2013}. The current state-of-the-art in this category are the works of Shi et al. \cite{shi2014} and Ikehata et al. \cite{ikehata2014}. Ikehata et al. models the reflectance function using a sum-of-lobes representation \cite{chandraker2011}, utilizing Bernstein polynomials as a basis for the inverse reflectance function and performing bivariate regression to determine the normal vectors. Shi et al. instead model the low-frequency reflectance component using polynomials of up to order three while discarding the high-frequency reflectance components. Of particular interest to this paper is another work by Ikehata et al. \cite{ikehata2014journal} that models the reflectance function as piecewise-linear. We explore this method in more detail in the following section. 

\revised{Another work of interest in this category is \cite{hui2015dictionary}. This work uses a dictionary-based approach as well, in this case attempting to simultaneously estimate the normal vectors of an object while learning to represent the BRDF of the object using a fixed dictionary of BRDFs from the MERL database \cite{matusik2003data}. While this work applies a dictionary to the problem of photometric stereo, it does so in a very different sense than we do---it uses a fixed rather than a learned dictionary and seeks to represent the BRDFs in the dictionary rather than the normal vectors.}

In addition to the aforementioned approaches, a variety of other robust photometric stereo methods have been proposed \cite{hertzmann2005,holroyd2008,sato2007,okabe2009,wu2006dense,chakrabarti2016single}. The recent work of Shi et al. \cite{shi2016} seeks to standardize future work in photometric stereo by introducing an extensive dataset to facilitate future testing and evaluation. Furthermore, they compare a variety of existing approaches on this dataset, providing a benchmark for future work.

\section{Problem Formulation} \label{sec:prelims}

\subsection{Basis of Photometric Stereo} \label{sec2a}

The Lambertian reflectance model states that, to an observer, the brightness of a point on a Lambertian surface is independent of the observer's viewing angle. Surfaces that follow this model are matte in appearance. Indeed, consider an image taken of a Lambertian object. The light intensity measured at pixel $(x,y)$ of the image satisfies the relationship
\begin{equation} \label{ps1}
I_{xy} = \rho_{xy} \ell^T n_{xy},
\end{equation}
where $I_{xy}$ is the image intensity at pixel $(x,y)$, $\ell \in \mathbb{R}^3$ is the direction of the light source incident on the surface of the object, $\|\ell\|_2$ is the light source intensity, $n_{xy} \in \mathbb{R}^3$ is the (unit) normal vector of the surface at $(x,y)$, and $\rho_{xy} \in \mathbb{R}$ is the surface albedo at $(x,y)$---a measure of the reflectivity of the surface. 

Suppose we fix the position of a camera facing the surface and vary the position of the light source over $d$ unique locations. Then we can write $d$ equations of the form~\eqref{ps1} and stack them into the matrix equation
\begin{equation} \label{ps2}
\begin{bmatrix}
I_{xy}^1 \\
\vdots \\
I_{xy}^d
\end{bmatrix} = 
\begin{bmatrix}
	\ell_1^T \\ \vdots \\ \ell_d^T \\
\end{bmatrix} 
\rho_{xy} n_{xy},
\end{equation}
where $I_{xy}^k$ denotes the image intensity at $(x,y)$ in the $k$th image. Assuming each of our $d$ images has dimension $m_1 \times m_2$, \eqref{ps2} can be solved $m_1 m_2$ times to obtain the normal vector of the object at each point on the surface. We may also combine these $m_1 m_2$ equations into a single matrix equation. Indeed, define the observation matrix
\begin{equation} \label{ddef}
Y \triangleq \left [\mathrm{vec}(I^1) \ldots \mathrm{vec}(I^d) \right ] \in \mathbb{R}^{m_1 m_2 \times d},
\end{equation}
where $\mathrm{vec}(I^k)$ is the vector formed by stacking the columns of $I^k$. Then, assuming the light sources are ideal (i.e., the incident light rays are parallel and of equal intensity at each point on the surface), we can collect \eqref{ps2} into the single equation
\begin{equation} \label{ps3}
Y = NL,
\end{equation}
where $N = [\rho_{11} n_{11}~\ldots~\rho_{m_1 m_2} n_{m_1 m_2}]^T \in \mathbb{R}^{m_1 m_2 \times 3}$ and $L = [\ell_1~\ldots~\ell_d] \in \mathbb{R}^{3 \times d}$. To avoid scaling ambiguity, we assume all light sources have intensity $\|\ell_k\|_2 = 1$. \revised{Note that, henceforth, when we refer to the normal vectors $N$, we are in fact referring to the normal vectors multiplied by the surface albedos.}

Each normal vector $n_{xy}$ contains three unknown components. Thus, given $d \geq 3$ images and the corresponding light directions, we can solve \eqref{ps3} to obtain the normal vector at each point on the object. Once computed, we can integrate the normal vectors to produce a full three-dimensional model of our surface \cite{simchony1990}.

\subsection{Deviations From the Lambertian Model}
While the Lambertian reflectance model is a good approximation of the reflectance properties of some surfaces, it is a poor approximation for many real-world objects. Lambertian objects are matte in appearance and thus any non-matte objects necessarily deviate from the Lambertian reflectance model. The latter class includes any object that exhibits specularities---bright points observed when light reflects off a shiny surface. Furthermore, even if an object is Lambertian, shadows (both self-cast and those produced by other objects) can cause the Lambertian model to break down.

One approach to modeling these effects is to modify \eqref{ps3} to
\begin{equation} \label{eq:error_model}
Y = NL + E,
\end{equation}
where $E$ is an additive error matrix accounting for non-Lambertian effects. Under this model, a simple, naive approach for estimating $N$ is to solve the least squares problem
\begin{equation} \label{eq:ls}
\min_{N} \ \left \| Y - NL \right \|_F^2,
\end{equation} 
which has solution $\hat{N} = YL^\dagger$, where $\dagger$ denotes the Moore-Penrose pseudoinverse. In this setting, one typically gathers $d > 3$ images so that the problem is overdetermined and thus provides some robustness to the non-Lambertian effects.

Several works apply further constraints to \eqref{eq:error_model}---such as constraining $E$ to be \emph{sparse}---allowing them to (in cases where their assumptions hold) derive more accurate estimates of $N$ than those obtained by \eqref{eq:ls} \cite{wu2011,ikehata2012}. In Section \ref{sec:dl}, \revised{we propose a novel approach that applies dictionary learning to the model defined by \eqref{eq:error_model}.}

\subsection{Piecewise Linear Reflectance Model} \label{sec:piecewise_reflectance}
Regardless of the constraints imposed on the additive error $E$, the model \eqref{eq:error_model} fundamentally relies on the Lambertian reflectance model, thus limiting its generality. Recent works have sought to move beyond the Lambertian assumption and utilize more general reflectance models that can accurately model the normal vectors of a wider range of objects \cite{ikehata2014journal,ikehata2014,shi2014}. In this work, we are particularly interested in the model presented in \cite{ikehata2014journal}, which we briefly summarize here.

A simple extension to the Lambertian model is to assume that the image intensity is related to the inner product of $\ell$ and $n$ through a nonlinear function. In other words, we modify \eqref{ps1} to read
\begin{equation}\label{eq:dr}
I_{xy} = f_{xy}(\ell^T n_{xy})
\end{equation}
for some nonlinear function $f_{xy}$. Assuming the reflectance function at each pixel, $f_{xy}$, is monotonically increasing, a unique inverse is guaranteed to exist. We can thus invert \eqref{eq:dr} and write
\begin{equation}\label{eq:pl_normal}
f_{xy}^{-1}(I_{xy}) = g_{xy}(I_{xy}) = \ell^T n_{xy}.
\end{equation}

Given a set of lighting vectors and corresponding images, our task is then to jointly estimate $g_{xy}(.)$ and $n_{xy}$ for each pixel. This is a highly underdetermined problem and so, to solve it in practice, further constraints must be imposed. A natural possibility is to assume $g_{xy}$ is piecewise linear. That is, we let
\begin{equation}\label{eq:pl_form}
g_{xy}(t) = \sum_{k=1}^p a_{xy}^k g_{xy}^k(t),
\end{equation}
where
\begin{equation}
g_{xy}^k(t) = \begin{cases}
0 & \text{if } t < b_{xy}^{k-1}, \\
t - b_{xy}^{k-1} & \text{if } b_{xy}^{k-1} \leq t \leq b_{xy}^k, \\
b_{xy}^k - b_{xy}^{k-1} & \text{if } t > b_{xy}^k.
\end{cases}
\end{equation}
Here, $p$ is a design parameter that determines the number of piecewise segments in $g_{xy}$, $b_{xy}^k$ are the inflection points of $g_{xy}$ (a strictly increasing sequence), and $a_{xy}^k > 0$ are the slopes of the segments. For simplicity, we set $b_{xy}^0 = 0$ and choose the remaining values of $b_{xy}^k$ to be equally spaced along the range of intensity values among the $d$ images at pixel $(x,y)$. Under these assumptions, model \eqref{eq:pl_normal} reduces to the problem of estimating the slopes $a_{xy}^1, \ldots, a_{xy}^p$ and the normal vector $n_{xy}$ at each pixel. Note that the case $a_{xy}^1 = \ldots = a_{xy}^p$ reduces to the Lambertian model \eqref{ps1}.

To simplify notation, let
\begin{equation}
a_{xy} = \begin{bmatrix} a_{xy}^1 & \ldots & a_{xy}^p \end{bmatrix}^T \in \mathbb{R}^{p}
\end{equation}
and 
\begin{equation}
\bar{g}_{xy}(t) = \begin{bmatrix} g_{xy}^1(t) & \ldots & g_{xy}^p (t) \end{bmatrix}^T \in \mathbb{R}^{p}
\end{equation}
and rewrite \eqref{eq:pl_form} as the vector product
\begin{equation}
g_{xy}(t) = \bar{g}_{xy}(t)^T a_{xy}.
\end{equation}
Similarly, \eqref{eq:pl_normal} can be written as
\begin{equation}
\bar{g}_{xy}(I_{xy})^T a_{xy} = \ell^T n_{xy}.
\end{equation}
Given $d$ images, let $C_{xy} \in \mathbb{R}^{d \times p}$ be the matrix whose $j$th row is $\bar{g}_{xy}(I_{xy}^j)^T$. Then we can collect the data from the $d$ images at pixel $(x,y)$ into the single equation
\begin{equation} \label{eq:dr1}
C_{xy} a_{xy} = L^T n_{xy},
\end{equation}
which is the analogue of \eqref{ps2} for the Lambertian model. Equation~\eqref{eq:dr1} can be solved for $n_{xy}$ and $a_{xy}$ to determine the normal vector and the corresponding nonlinear reflectance function at $(x,y)$. To avoid \revised{the trivial solution $a_{xy} = n_{xy} = 0$,} one can constrain $1^T a_{xy} = 1$ and then normalize $n_{xy}$ to unit norm after solving \eqref{eq:dr1}.

As in the Lambertian case, the model \eqref{eq:pl_normal} only accounts for the reflectance properties of the object. While significantly more general than the Lambertian model, non-idealities present in the images that do not conform to these reflectance properties---or, more explicitly, that do not follow a piecewise linear relationship between $I_{xy}$ and $\ell^T n_{xy}$---will prevent \eqref{eq:dr1} from holding exactly. Thus, analogous to  \eqref{eq:error_model}, we modify \eqref{eq:dr1} to yield the model
\begin{equation} \label{eq:dr2}
C_{xy} a_{xy} = L^T n_{xy} + e,
\end{equation}
where $e \in \mathbb{R}^d$ accounts for any corruptions in the data not captured by the piecewise reflectance model. A simple approach to fiting model \eqref{eq:dr2} to data is to solve the constrained least squares problem\footnote{Note that we do not explicitly constrain values of $a_{xy}$ to be positive, although this is strictly required to interpret $g_{xy}(.)$ as the inverse of a reflectance model $f_{xy}(.)$.}
\begin{equation}\label{eq:dr3}
\min_{n_{xy},~a_{xy}} \ \left \| C_{xy} a_{xy} - L^T n_{xy} \right \|_2^2 \\
\ \ \text{s.t.} \ \ 1^T a_{xy} = 1.
\end{equation}
\revised{In the sequel, we denote this model as the Piecewise-linear Least Squares (PLS) method for photometric stereo.} In practice, one can improve robustness by applying some regularization to the modeling error $e$ from \eqref{eq:dr2}. In particular, \cite{ikehata2014journal} utilizes this reflectance model and assumes the corruptions to the data are \emph{sparse}. In the next section we propose an alternative model based on dictionary learning to robustly solve \eqref{eq:dr3}.

\section{Dictionary Learning Approaches} \label{sec:dl}

Dictionary learning refers to a class of algorithms that seek to sparsely represent some data of interest with respect to a learned ``dictionary''---a collection of basis or atom elements. Intuitively, dictionary learning methods allow one to uncover structure present in data without a priori knowledge of the form of the structure. In this section, we propose \revised{two} adaptive dictionary learning algorithms for photometric stereo. \revised{First, however, we motivate our study of dictionary learning for photometric stereo by examining the ability of sparse dictionary models to capture the structure of surface normal vectors.}

\revised{
\subsection{Motivation for Dictionary Learning} \label{subsec:dlmotivation}

Given a set of signals that are stored as columns of a matrix $P$, dictionary learning seeks to learn a dictionary $D$ and a matrix $B$ of sparse codes such that $P \approx DB$. Traditionally, the sparsity of $B$ is achieved by a constraint of the form $\|b_i\|_0 \leq s \ \forall i$, where $b_i$ denotes the $i$th column of $B$ and the $\ell_0$ ``norm" counts the number of non-zero entries in a matrix. A seminal example of this approach is the K-SVD method \cite{aharon2006ksvd}, and various alternatives exist that replace the $\ell_0$ ``norm" with other sparsity-promoting functions or enforce additional structure on the dictionary \cite{barchiesi2013learning,ramirez2010classification,rubinstein2010double}.

Inspired by recent successes \cite{ravishankar2016low,ravishankar2016lassi,dinokat2016} in other inverse problem settings, in this work we consider the following dictionary learning formulation:
\begin{equation} \label{eq:sai:dl}
\begin{array}{rl}
\displaystyle\min_{D,B} & \|P - DB\|_F^2 + \mu^2 \|B\|_0 \\
\text{s.t.} & \|B\|_{\infty} \leq q, ~ \|d_i\|_2 = 1, ~ \forall i,
\end{array}
\end{equation}
where the regularization parameter $\mu > 0$ controls the overall sparsity of the matrix $B$ and the constraints are included to make the problem well-posed \cite{ravishankar2017efficient}. Note that \eqref{eq:sai:dl} penalizes the number of non-zeros in the entire matrix $B$, which allows for variable sparsity levels in each column. Variable sparsity is a useful model for imaging applications, where the matrix $P$ is usually constructed from vectorized image or tensor patches. Patches from different regions of an image typically contain different amounts of information, and the model \eqref{eq:sai:dl} allows the sparse codes to adapt to these local complexities \cite{ravishankar2017efficient}.

To justify the adoption of a dictionary learning framework for photometric stereo, it is important to first demonstrate that dictionary models can capture the local structure of the normals of real objects. To this end, we considered the three objects shown in Figure~\ref{fig:dl_dilgent_normals} from the recent DiLiGenT dataset \cite{shi2016}. For each dataset, we constructed a patch matrix $P \in \mathbb{R}^{192 \times w}$ by extracting overlapping $8 \times 8 \times 3$ patches from the normals,\footnote{Note that the normal vectors used to construct $P$ are not the ground truth normal vectors found in the DiLiGenT dataset, but rather the surface normal vectors computed from the DiLiGenT surfaces using the traditional least squares model. In order to justify the use of dictionary learning, it is important to demonstrate that the learned dictionaries can capture both geometry and texture (albedo). The ground truth normal vectors in the DiLiGenT dataset do not express texture (each normal vector in the ground truth normals is normalized to have unit norm), so we used the unnormalized vectors produced by the least squares method to force the dictionaries to capture both the texture and geometry of the object.} where the third dimension denotes their $x$, $y$, and $z$ coordinates. We then learned square $192 \times 192$ dictionaries by (approximately) solving \eqref{eq:sai:dl} for different choices of $\mu$ using the efficient block coordinate descent algorithm proposed in \cite{ravishankar2017efficient}.

\begin{figure}[t!]
\centering
\begin{subfigure}[b]{0.14\textwidth}
  \includegraphics[width=\textwidth]{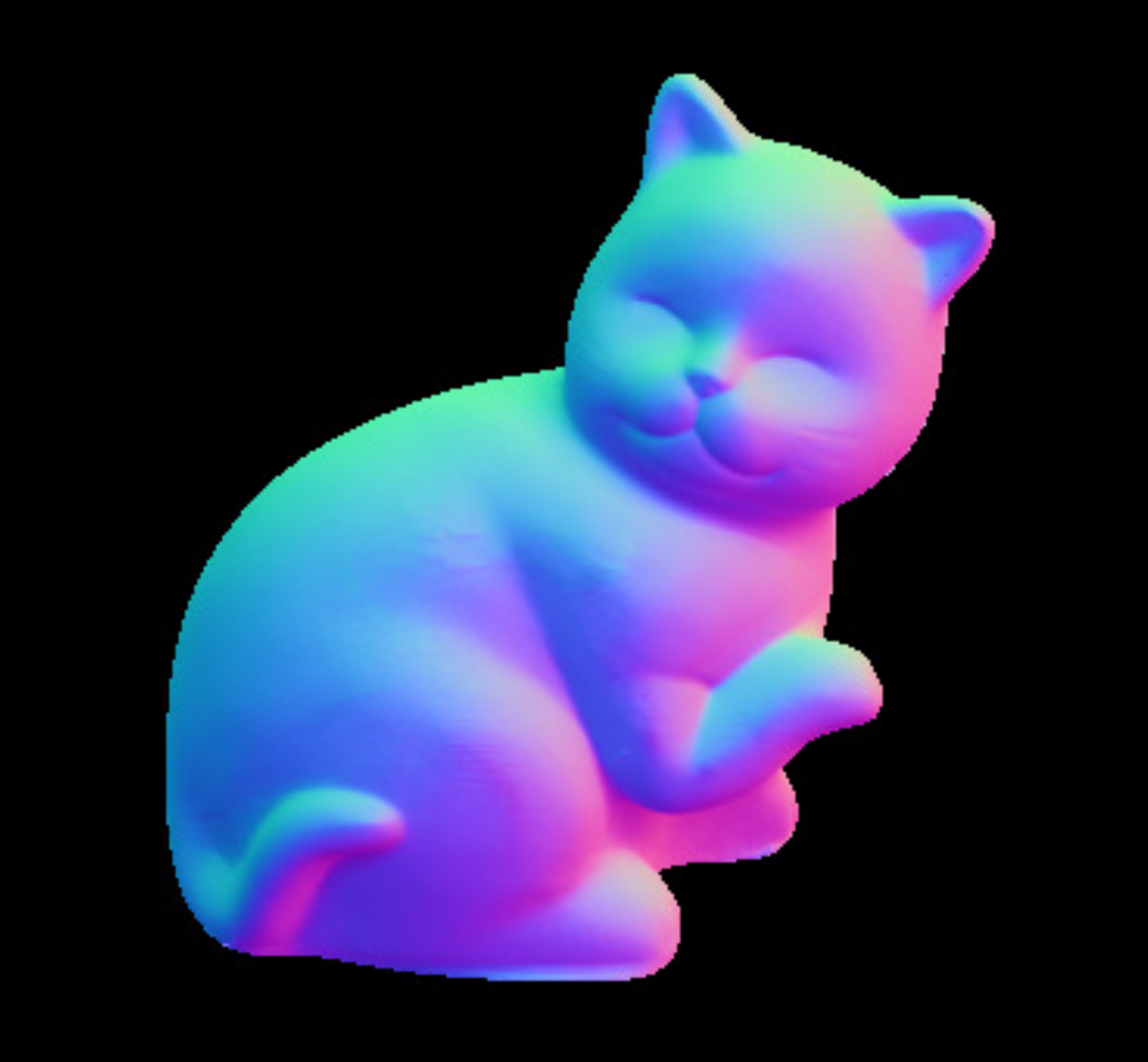}
  \caption{Cat}
\end{subfigure}
\hspace{-2mm}
\begin{subfigure}[b]{0.1635\textwidth}
  \includegraphics[width=\textwidth]{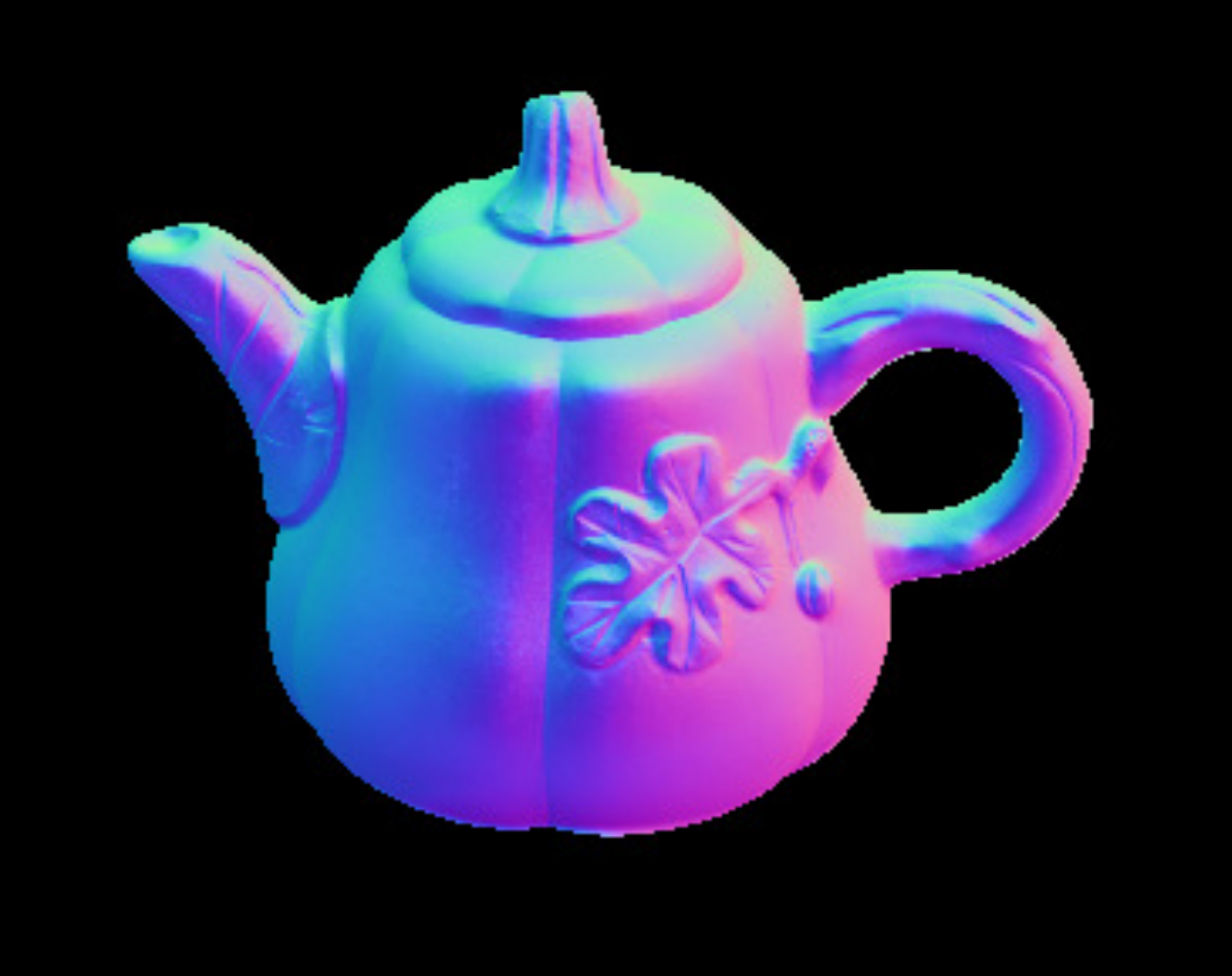}
  \caption{Pot2}
\end{subfigure}
\hspace{-2mm}
\begin{subfigure}[b]{0.1635\textwidth}
  \includegraphics[width=\textwidth]{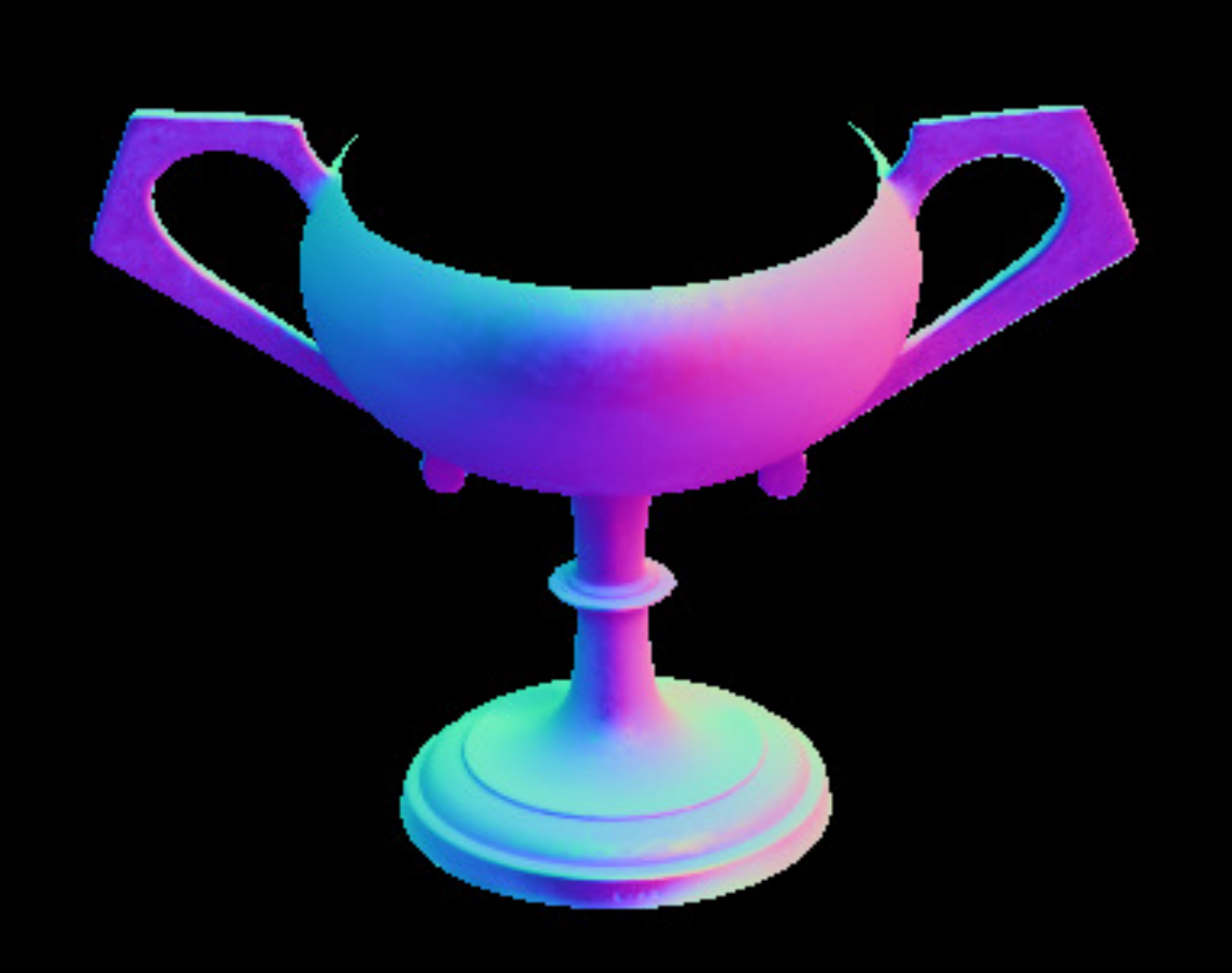}
  \caption{Goblet}
\end{subfigure}
\caption{\revised{Surface normal vectors for three objects from the DiLiGenT dataset \protect{\cite{shi2016}}.}}
\label{fig:dl_dilgent_normals}
\end{figure}

\begin{figure}[t!]
\centering
\includegraphics[width=0.47\textwidth]{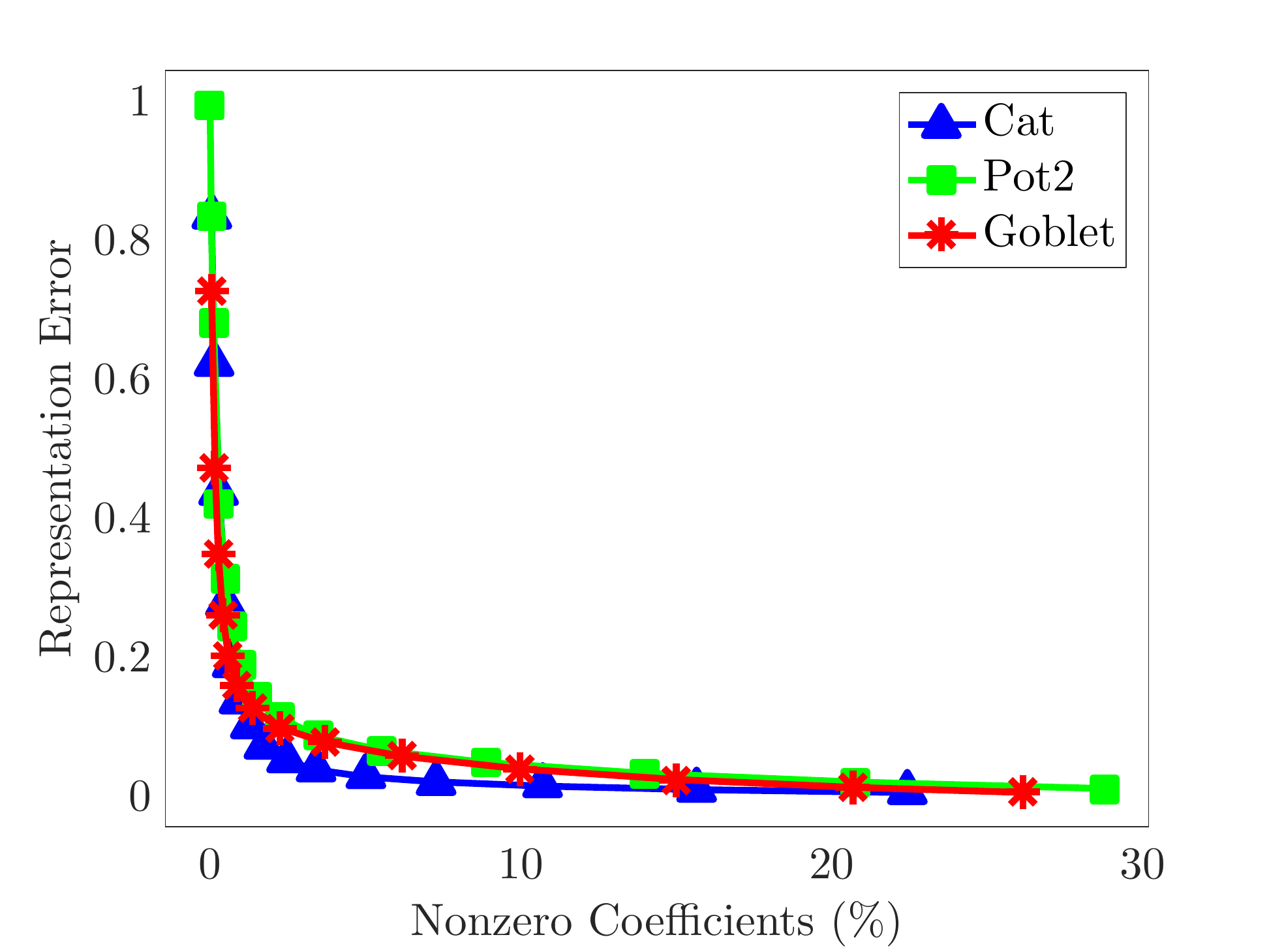}
\caption{\revised{The normalized sparse representation error (NSRE) $\|P - DB\|_{F}/ \|P\|_{F}$ for the $192 \times 192$ dictionaries learned on the $8 \times 8 \times 3$ overlapping spatiotemporal patches of the (unnormalized) normal vectors from Figure~\ref{fig:dl_dilgent_normals}. The plot shows NSRE for various choices of $\mu$ corresponding to different fractions of nonzero coefficients in the sparse codes $B$.}}
\label{fig:dl_rep_error}
\end{figure}

Figure~\ref{fig:dl_rep_error} plots the normalized sparse representation error (NSRE) $\|P - DB\|_{F} / \|P\|_{F}$ versus the sparsity of $B$ as $\mu$ varies. The fact that the NSRE approaches zero while $B$ remains sparse demonstrates that sparse dictionary-based models are sufficiently expressive to capture the intricacies of real surfaces. With this observation in hand, we now present our proposed approaches to dictionary learning-based robust photometric stereo.
}

\subsection{Normal Vectors through Dictionary Learning (DLNV)} \label{subsec:dlnv}
We \revised{first} propose modifying \eqref{eq:ls} by adding an adaptive dictionary learning regularization term applied to the normal vectors. Under this approach, we seek a normal map that agrees with the Lambertian model \eqref{eq:ls} while also having a locally sparse representation with respect to a learned dictionary---resulting in a smoother normal map that is robust to non-idealities in the data. Specifically, we propose solving the optimization problem:
\begin{equation} \label{eq:dlnv}
\begin{array}{rl}
\displaystyle\min_{n,B,D} & \| y - A n \|_2^2 + \lambda \bigg(\displaystyle\sum_{j=1}^w \| P_j n - D b_j \|_2^2 + \mu^2 \| B \|_0 \bigg) \\
\text{s.t.} & \| B \|_{\infty} \leq q, \ \ \| d_i \|_2 = 1 \ \forall i.
\end{array}
\end{equation}
Here, $y = \vec(Y) \in \mathbb{R}^{m_1 m_2 d}$ and $A = L^T \otimes I \in \mathbb{R}^{m_1 m_2 d \times 3 m_1 m_2}$, where $\otimes$ denotes the Kronecker product and $I$ is the $m_1 m_2 \times m_1 m_2$ identity matrix. Furthermore, $n = \vec(N) \in \mathbb{R}^{3 m_1 m_3}$ are the vectorized normal \revised{$\times$ albedo} vectors. \revised{$P_j \in \mathbb{R}^{w_x w_y w_z \times 3 m_1 m_2}$ denotes a $\{ 0, 1 \}$ patch extraction matrix} that extracts vectorized patches from $n$ of dimensions $w_x \times w_y \times w_z$, where $n$ is treated as an $m_1 \times m_2 \times 3$ tensor during extraction. In practice, we extract patches from $N$ using a simple sliding window strategy. Also, $D \in \mathbb{R}^{w_x w_y w_z \times K}$ denotes the learned dictionary whose columns (atoms) $d_i$ can be thought of as vectorized $w_x \times w_y \times w_z$ tensors. \revised{$B \in \mathbb{R}^{K \times w}$ is a sparse coding matrix whose columns $b_j$ define the (usually sparse) linear combinations of dictionary atoms used to represent each patch}. \revised{The parameter $K$ specifies the number of atoms in our dictionary $D$, and $w$ is the number of patches extracted from $n$. Also, $\|.\|_0$ is the familiar \revised{$\ell_0$ ``norm",} and $\lambda \geq 0$ and $\mu \geq 0$ are regularization parameters.}

\revised{We impose the constraint $||B||_{\infty} \triangleq \text{max}_j ||b_j||_{\infty} \leq q$, where $q$ is typically very large, to prevent any instability that could theoretically arise due to \eqref{eq:dlnv} being non-coercive with respect to $B$, but the constraint is inactive in practice \cite{ravishankar2017efficient}. Without loss of generality, we also constrain the dictionary atoms $d_i$ to unit-norm to avoid scaling ambiguity between $D$ and $B$ \cite{kar}.}

Intuitively, the adaptive dictionary learning regularization in \eqref{eq:dlnv} is able to uncover underlying local structure in $N$ that the least squares formulation \eqref{eq:ls} alone cannot deduce from the images. This results in normal vectors that are ``smooth" and free from noise and other non-idealities that may otherwise corrupt them. \revised{Note that, in addition to capturing the geometry of the normal vectors, the dictionary atoms must also capture the texture (albedo) of the surface. This is because, as noted in Section \ref{sec2a}, we absorb the albedo $\rho$ into the normal vectors $n$. Thus, individual elements of $n$ need not have unit norm---their norm corresponds to the estimated albedo at the corresponding point on the surface.}

Henceforth, we refer to this approach as the Dictionary Learning on Normal Vectors (DLNV) method, and we present our algorithm for solving \eqref{eq:dlnv} in Section~\ref{sec:dlalgos}.

\subsection{Non-Lambertian Normal Vectors through Dictionary Learning (PDLNV)} \label{subsec:pdlnv}
\revised{We next} present a method that is based on the Piecewise-linear Least Squares (PLS) model from Section \ref{sec:piecewise_reflectance}. Using \eqref{eq:dr3} as the baseline, our approach is to incorporate a dictionary learning term to increase robustness to corruptions. In particular, we again apply dictionary learning regularization to the normal vectors, thus constraining them to agree with the non-Lambertian model \eqref{eq:dr3} while also admitting a sparse representation in the learned dictionary. Specifically, we propose solving the optimization problem:
\begin{equation} \label{eq:pdlnv}
\begin{array}{rl}
\displaystyle\min_{n,B,D,a} & \displaystyle\sum_{x,y} \bigg(\| C_{xy} a_{xy} - L^T n_{xy} \|_2^2 + \gamma \| 1^T a_{xy} - 1 \|_2^2 \bigg) \ + \\
& \lambda \bigg( \displaystyle\sum_{j=1}^w \| P_j n - D b_j \|_2^2 + \mu^2 \| B \|_0 \bigg) \\[10pt]
\text{s.t.} & \| B \|_{\infty} \leq q, \ \ \| d_i \|_2 = 1 \ \forall i.
\end{array}
\end{equation}
Here, all terms are defined analogously as in Sections~\ref{sec:piecewise_reflectance} and \ref{subsec:dlnv}. Note that we include the constraint $1^T a_{xy} = 1$ from \eqref{eq:dr3} in penalty form, where we typically set parameter $\gamma \geq 0$ to be very large. \revised{As in DLNV, we do not constrain $n_{xy}$ to have unit norm, so the dictionary $D$ captures both the shape and texture of the surface.}

The problem~\eqref{eq:pdlnv} can be thought of as a generalization of DLNV. Indeed, if we set $p = 1$ and $\gamma = \infty$, then \eqref{eq:pdlnv} reduces to \eqref{eq:dlnv}. However, we present both models as distinct methods in this work to highlight the differences between models that rely on the Lambertian assumption versus models that incorporate more complex reflectance models. We investigate the performance of both approaches in detail in Section~\ref{sec:experiments}. Henceforth, we refer to this approach as the Dictionary Learning on Normal Vectors with Piecewise-Linear Reflectance (PDLNV) method, and we present our algorithm for solving \eqref{eq:pdlnv} in Section~\ref{sec:dlalgos}.

\section{Algorithms and Properties} \label{sec:dlalgos}

We propose solving \eqref{eq:dlnv} and \eqref{eq:pdlnv} via block coordinate descent-type algorithms. Specifically, for \eqref{eq:dlnv} we alternate between updating $n$ with $(D,B)$ fixed and updating $(D,B)$ with  $n$ fixed. For \eqref{eq:pdlnv} we use a similar strategy where we alternate between updating $n$, $(D,B)$, and $a$ with all other variables held fixed. For each subproblem, we now derive simple and efficient schemes for minimizing the associated cost.

\subsection{\revised{Updating Dictionary $(D)$ and Sparse Codes $(B)$}}

The $(D,B)$ update is identical for \revised{both} methods. Here we present the update using the notation from \revised{\eqref{eq:dlnv}}. For notational convenience, we define $G \triangleq B^T$ and denote by $P$ the matrix whose $j$th column is $P_j n$. With $n$ fixed, the optimization with respect to $(D,B)$ can be written as
\begin{equation} \label{dbupdate}
\begin{array}{rl}
\displaystyle\min_{G,D} & \|P - DG^T\|_F^2 + \mu^2 \|G\|_0 \\
\text{s.t.} & \|G\|_{\infty} \leq \revised{q}, \ \ \|d_i\|_2 = 1 \ \forall i.
\end{array}
\end{equation}
We (approximately) solve \eqref{dbupdate} by applying a few iterations of block coordinate descent, where we iterate over the columns $g_i$ of $G$ and columns $d_i$ of $D$ sequentially. For each $1 \leq i \leq K$, we minimize \eqref{dbupdate} first with respect to $g_i$ and then with respect to $d_i$, holding all other variables fixed.

We first consider the minimization of \eqref{dbupdate} with respect to $g_i$. Define $E_i \triangleq P - \sum_{k \neq i} d_k g_k^T$, where $E_i$ is computed using the most recent values of the dictionary atoms and coefficients. Then we can write the $g_i$ subproblem as
\begin{equation} \label{gupdate}
\displaystyle\min_{g_i} \ \left \| E_i - d_i g_i^T \right \|_F^2 + \mu^2 \left \| g_i \right \|_0 \ \ \text{s.t.} \ \ ||g_i||_{\infty} \leq q.
\end{equation}
The solution to \eqref{gupdate} is given by \cite{sairajfes}
\begin{equation} \label{gopt}
\hat{g}_i  = \min \left ( | H_{\mu} (E_i^T d_i) |, q1_w \right ) \odot \sign\left ( H_{\mu} (E_i^T d_i ) \right ),
\end{equation}
where $1_w \in \mathbb{R}^{w}$ is a vector of ones, $\min(.,.)$ is applied element-wise to vector arguments, and $\odot$ denotes element-wise multiplication. Furthermore, $H_{\mu}(.)$ denotes the element-wise hard thresholding function, defined as
\begin{equation} \label{softdl}
H_\mu (y) = \begin{cases}
0 & \text{if } |y| < \mu \\
y & \text{if } |y| \geq \mu.
\end{cases}
\end{equation}

Minimizing \eqref{dbupdate} with respect to $d_i$ can be written as
\begin{equation} \label{dupdate}
\displaystyle\min_{d_i} \ \left \| E_i - d_i g_i^T \right \|_F^2 \ \ \text{s.t.} \ \ ||d_i||_{2} = 1.
\end{equation}
The solution to \eqref{dupdate} is given by \cite{sairajfes}
\begin{equation} \label{dopt}
\hat{d}_i = \begin{cases}
\dfrac{E_i g_i}{\left \| E_i g_i \right \|_2}, & \text{if } g_{i}\neq 0 \\ 
u , & \text{if } g_{i} = 0,
\end{cases}
\end{equation}
where \revised{$u \in \mathbb{R}^{w_x w_y w_z}$} is an arbitrary unit-norm vector (e.g., the first column of the \revised{$w_x w_y w_z \times w_x w_y w_z$} identity matrix).

\subsection{\revised{Updating Normal Vectors $(n)$}}
Minimizing \eqref{eq:dlnv} with respect to $n$ yields the problem
\begin{equation} \label{nupdate}
\displaystyle\min_{n} \| y - A n \|_2^2 + \lambda \displaystyle\sum_{j=1}^w \| P_j n - D b_j \|_2^2.
\end{equation}
Although \eqref{nupdate} is a least-squares problem, its normal equation cannot be easily inverted due to the presence of the $A$ matrix. Instead, we perform a few iterations of proximal gradient \cite{parboyd} to (approximately) solve \eqref{nupdate}.\footnote{Proximal gradient is one of many possible iterative schemes for minimizing the quadratic objective \eqref{nupdate}; one could also employ a different algorithm, such as preconditioned conjugate gradient.} The cost function can be written in the form $f(n) + g(n)$ where $f(n) = \| y - A n \|_2^2$ and $g(n) = \lambda \sum_{j=1}^w \| P_j n - D b_j \|_2^2$, so we perform the proximal steps
\begin{equation} \label{nproxupdate}
n^{k+1} = \text{prox}_{\tau g}(n^{k} - \tau \nabla f(n^{k})),
\end{equation}
where
\begin{equation} \label{nprox}
\text{prox}_{\tau g} (y) := \argmin_{x} \ \frac{1}{2} \| y - x \|_2^2 + \tau g(x)
\end{equation}
is the proximal operator of $g$ and $\tau > 0$ is a chosen step size. The updates \eqref{nproxupdate} are guaranteed to converge to a solution of \eqref{nupdate} when $\tau < 1 / \|A\|^2 = 1 / \|L\|^2$, and in fact the cost will monotonically decrease when $\tau \leq 1 / 2\|L\|^2$ is used \cite{parboyd}.

Define $\tilde{n}^{k} \triangleq n^{k} - \tau \nabla f(n^{k}) = n^{k} - 2\tau A^T (An^{k} - y)$. Then, after substituting \eqref{nprox} into \eqref{nproxupdate} and simplifying, one can show that $n^{k+1}$ satisfies the normal equation
\begin{equation} \label{eq:nnormal}
 \bigg(I + 2 \tau \lambda \displaystyle\sum_{j=1}^w P_j^T P_j \bigg) n^{k+1} = \tilde{n}^{k} +  2 \tau \lambda \displaystyle\sum_{j=1}^w P_j^T D b_j.
\end{equation}
The matrix multiplying $n^{k+1}$ in \eqref{eq:nnormal} is diagonal and thus can be efficiently inverted to compute $n^{k+1}$.

In the case of PDLNV, the $n$ update for \eqref{eq:pdlnv} can be solved in an identical manner, where the analogous data matrix $y$ is constructed as
\begin{equation}
y = \mathrm{vec}\left(\begin{bmatrix} C_{11} a_{11}~\ldots~C_{m_1 m_2} a_{m_1 m_2} \end{bmatrix}\right)
\end{equation}
from the most recent values of $a_{xy}$.

\subsection{\revised{Updating Reflectance Function Slopes $(a)$}}
Minimizing \eqref{eq:pdlnv} with respect to $a_{xy}$ yields $m_1 m_2$ problems of the form
\begin{equation} \label{aupdate1}
\min_{a_{xy}} \ \sum_{x,y} \left \| C_{xy} a_{xy} - L^T n_{xy} \right \|_2^2 + 
\gamma \left \| 1^T a_{xy} - 1 \right \|_2^2.
\end{equation}
These are simple least squares problems with $d + 1$ equations and $p$ unknowns that can be solved exactly and in parallel. Indeed, the solution to \eqref{aupdate1} is
\begin{equation} \label{aupdate2}
\hat{a}_{xy} = \begin{bmatrix} C_{xy} \\ \gamma 1^T \end{bmatrix}^\dagger \begin{bmatrix} L^T n_{xy} \\ \gamma \end{bmatrix}.
\end{equation}
The pseudoinverse in \eqref{aupdate2} is a constant that can be pre-computed from the raw images, so $a_{xy}$ can be updated efficiently.

\subsection{Convergence}

The proposed algorithms for solving \eqref{eq:dlnv} and \eqref{eq:pdlnv} alternate between updating $(D,B)$, $n$, and $a$ (PDLNV only) with the other variables held fixed. Except for the $n$ updates of DLNV and PDLNV, all update schemes are either exact block coordinate descent updates or composed of inner iterations of exact block coordinate descent updates, so the objectives in our formulations must be monotonically decreasing (non-increasing) during these updates. Moreover, the proximal gradient step size for the $n$ update can be chosen to guarantee that these iterations also monotonically decrease their objectives. Thus, the cost functions for \revised{both} proposed algorithms are monotonically decreasing and bounded below by zero, so they must converge. Whether the algorithm iterates themselves converge to critical points of the (non-convex) costs is an interesting theoretical question for future work.

\section{Numerical Experiments} \label{sec:experiments}

We now investigate the performance of our proposed dictionary learning-based methods experimentally. To obtain quantitative results, we rely primarily on the recent DiLiGenT benchmark dataset \cite{shi2016}. This dataset contains images of a variety of surfaces of different material and provides the true normal vectors of each object, allowing us to measure the accuracy of the normal vectors produced by each method. We quantify the error in each estimated normal vector by measuring the angular difference between it and the correponding true normal vector.

We evaluate our methods in a variety of settings. For each experiment, we compare the results of our methods to the robust PCA (RPCA) approach of Wu et al. \cite{wu2011}, the sparse regression (SR) method of Ikehata et al. \cite{ikehata2012}, and the constrained bivariate regression (CBR) approach of Ikehata et al. \cite{ikehata2014}. In addition, we compare with the baseline least squares (LS) model defined by \eqref{eq:ls}. \revised{For each existing method, we use the code supplied by the original authors.}

With the exception of LS, each method contains one or more regularization parameters. For each method, we sweep the parameters across a wide range of values and select the optimal parameters for each trial. For existing methods, we include any recommended parameters from the respective papers in our sweep. \revised{Experimentally, on the uncorrupted DiLiGenT dataset, values of $\lambda$ between $10^{-2}$ and $1$ and values of $\mu$ between $10^{-3}$ and $10^{-2}$ are approximately optimal for PDLNV. For DLNV on the uncorrupted DiLiGenT dataset, values of $\lambda$ between $10^{-2}$ and $10^{-1}$ and values of $\mu$ between $1$ and $10$ are approximately optimal. On data with higher levels of added corruption, $\lambda$ values between $1$ and $1000$ and $\mu$ values between $10^{-2}$ and $10^{-1}$ produce approximately optimal results for both methods. For PDLNV, $p = 2$ typically produces the best results. However, for objects with more complex reflectance patterns, such as Harvest, Cow, and Reading, larger values of $p$ tend to produce the best results. In all experiments, we terminate PDLNV after 50 iterations and DLNV after 20 iterations. For the existing methods, we use the termination criteria built into the code supplied by the original authors. In each experiment with additive noise, we run multiple trials with different noise realizations and average the results.}

The majority of the photometric stereo literature has focused primarily on the problem of reconstructing normal vectors from uncorrupted and generally large datasets (many images of each object), such as the DiLiGenT dataset. In cases where additional corruptions were added, the corruptions were typically sparse to better align with the modeling assumptions of each method. In our experiments, we endeavor to fully investigate the robustness of our proposed methods and existing methods to general non-sparse corruptions. Specifically, we corrupt the raw images with Poisson noise, which is a realistic model for noise in real images \cite{hasinoff2014photon}. This model is applicable, for example, when performing photometric stereo in low-light conditions, where noise levels can be significant. \revised{While most available photometric stereo datasets were obtained in controlled environments using high-quality cameras, it is likely that in real-world applications of photometric stereo, one may not be able to require such pristine images \cite{ackermann2015survey}. Thus, among our experimental results in this section, we consider data corrupted by artificial noise to explore the ability of each method to robustly reconstruct normal vectors from non-ideal data.}

 \revised{In experiments where we corrupt the images with noise, the images are corrupted with noise in an additive fashion and the signal-to-noise ratio (SNR) is measured as the power of the signal over the power of the added noise according to
\begin{equation}
\mathrm{SNR} = 10 \log \left ( \frac{\sum_{k = 1}^d \sum_{x = 1}^{m_1} \sum_{y = 1}^{m_2} ( I_{xy}^k)^2}{\sum_{k = 1}^d \sum_{x = 1}^{m_1} \sum_{y = 1}^{m_2} ( \hat{I}_{xy}^k - I_{xy}^k)^2} \right ),
\end{equation}
where $\hat{I}$ denotes the image corrupted by noise and $I$ denotes the true image.}

Many existing photometric stereo algorithms apply a pixel-wise mask as a preprocessing step to remove shadows from the images. Such masks are typically computed by performing a simple thresholding operation on the data and excluding any pixels below a chosen threshold from subsequent computations. While this strategy can improve results in some cases, it does not capture the complexity of shadows present in the image and often results in useful data being rejected. This is of particular importance when working with small or heavily corrupted datasets, where it is important that the reconstruction method has access to as much data as possible to uncover the relevant information. Such robust methods should have the capacity to adapt to shadows in images without the use of a shadow mask. As such, all of the experiments we present here are performed without the use of shadow masks.

\revised{
\subsection{Parameters for Dictionary Learning Based Approaches}
In addition to regularization parameters, which were directly optimized for each method in our experiments, there are multiple model parameters that can be tuned. The dimensions of the dictionary atoms---which correspond to the patch sizes that are extracted from the normal vectors---can be changed, the patch extraction strategy---e.g., non-overlapping patches or overlapping patches with a given spatial stride---can be changed, and one must choose how to initialize the dictionary. Another interesting parameter is the number of atoms (columns) in the dictionary, where we are free to choose between tall (undercomplete), square, or wide (overcomplete) dictionaries. In the case of PDLNV, we must also set the parameter $p$ that controls the number of piecewise segments in the non-Lambertian model; the optimal value for $p$ may depend on the properties of a given surface.

In the following experiments, we use $8 \times 8 \times 3$ patches for both PDLNV and DLNV, where the third dimension corresponds to the $(x,y,z)$ coordinates of each normal vector. In each case, we used a sliding window strategy with a spatial stride of 4 pixels in each direction to extract overlapping patches from the images or normal vectors. We used square dictionaries for our experiments (containing 192 atoms for DLNV and PDLNV), and we initialized each dictionary to the discrete cosine transform (DCT) matrix of appropriate size. For the $n$ updates, we used the step size $\tau = 1/2\|L\|^2$ to guarantee that the updates will monotonically decrease their objectives. A more in-depth investigation of how these parameters affect the results is given in Section \ref{sec:alg_props}.

The images in the DiLiGenT dataset have dimension $614 \times 514$ and we use the full image size in each trial. The images used in Section \ref{sec:ex_non_diligent}---the Hippo and Cat datasets---have dimensions $580 \times 580$ and $500 \times 640$, respectively. 

}

\revised{
\subsection{Comparison Between PDLNV and PLS}
To further motivate our use of dictionary learning and illustrate its strength, we first compare the performance of PDLNV (\ref{eq:pdlnv}) with PLS (\ref{eq:dr3}), the piecewise-linear least squares formulation found in \cite{ikehata2014journal}. Because PLS is a special case of PDLNV (PLS is equivalent to PDLNV with $\lambda = 0$), comparing these two methods allows us to isolate the contributions of dictionary learning---any difference between PLS and PDLNV will be due exclusively to the presence of dictionary learning-based  regularization.

We first compare the performance of each method on the DiLiGenT dataset \cite{shi2016}. We perform reconstructions based on all 96 images in each object, and we do not artificially corrupt the images. Table \ref{tab:full-diligent-pdlnv} compares the performance of PDLNV and PLS on this dataset.

\begin{table}[t!]
\begin{center}
\begin{tabular}{|c|c|c||c|c|}
\cline{1-5}
\multirow{ 3}{*}{Dataset} & \multicolumn{2}{c||}{Mean Angular Error} & \multicolumn{2}{c|}{Median Angular Error} \\
 & \multicolumn{2}{c||}{(degrees)} & \multicolumn{2}{c|}{(degrees)} \\
\cline{2-5}
 & PDLNV & PLS & PDLNV & PLS \\
\cline{1-5}
Ball & \textbf{3.60} & 3.63 & \textbf{1.95} & 2.14 \\
\cline{1-5}
Cat & \textbf{6.40} & 6.44 & \textbf{3.58} & 3.63 \\
\cline{1-5}
Pot1 & \textbf{6.99} & 7.02 & \textbf{3.70} & 3.76 \\
\cline{1-5}
Bear & \textbf{8.51} & 8.76 & \textbf{6.34} & 6.67 \\
\cline{1-5}
Pot2 & \textbf{9.94} & 9.95 & \textbf{7.17} & \textbf{7.17} \\
\cline{1-5}
Buddha & \textbf{13.56} & \textbf{13.56} & 7.84 & \textbf{7.83} \\
\cline{1-5}
Goblet & \textbf{14.58} & \textbf{14.58} & 9.56 & \textbf{9.53} \\
\cline{1-5}
Reading & \textbf{20.18} & 21.25 & \textbf{14.85} & 15.46 \\
\cline{1-5}
Cow & \textbf{13.71} & 16.62 & \textbf{10.11} & 13.69 \\
\cline{1-5}
Harvest & \textbf{20.54} & 30.10 & \textbf{15.28} & 24.33 \\
\hhline{=====}
\textit{Average} & \textbf{\textit{11.80}} & \textit{13.19} & \textbf{\textit{8.04}} & \textit{9.42} \\
\cline{1-5}
\end{tabular}
\end{center}
\caption{\revised{Mean and median angular errors (in degrees) of the estimated normal vectors produced by PDLNV and PLS on full, uncorrupted DiLiGenT datasets. PDLNV performs no worse that PLS on each dataset, and it performs significantly better on the Cow and Harvest objects.}}
\label{tab:full-diligent-pdlnv}
\end{table}

From this table we see that PDLNV either significantly outperforms PLS (such as on the more complex Reading, Cow, and Harvest objects) or PDLNV performs approximately the same as, or marginally better than, PLS (such as on the simpler Ball, Cat, Pot1, Bear, Pot2, Buddha, and Goblet objects). Thus, we conclude that on large, uncorrupted datasets, incorporating dictionary learning into the model does not decrease reconstruction accuracy, and it may significantly improve performance for some datasets.

PDLNV yields a more significant improvement with respect to PLS in performance when investigating smaller datasets and datasets with added corruptions. Indeed, Figure \ref{fig:pdlnv_pot1_images} gives examples of several images contained in the DiLiGenT Pot1 dataset and Figure \ref{fig:pdlnv_pot1_normals} illustrates the normal vectors reconstructed by both PDLNV and PLS on a set of 10 randomly selected images from the DiLiGenT Pot1 dataset. Figure \ref{fig:pdlnv_pot1_normals_zoom} provides a close-up view of the spout of the pot and Figure \ref{fig:pdlnv_pot1_errors} depicts the angular errors of the reconstructed normal vectors. Both PDLNV and PLS have difficulty accurately reconstructing certain regions of the pot---for instance around the base of the handle---and both reconstruct the majority of the spout and body with reasonable accuracy. However, as is clearly visible in Figure \ref{fig:pdlnv_pot1_normals_zoom}, PLS introduces errors in the spout region of the pot while PDLNV almost completely removes these errors. Upon close inspection of Figure \ref{fig:pdlnv_pot1_images}, we see that, depending on the direction of illumination, there is a wide specular lobe centered on the body of the pot. This effect is likely the cause of error in the reconstruction produced by PLS. However, PDLNV is able to mitigate the influence of these non-Lambertian effects and produce a more accurate reconstruction.

\begin{figure}[t!]
\centering
\begin{subfigure}[b]{0.16\textwidth}
  \includegraphics[width=\textwidth]{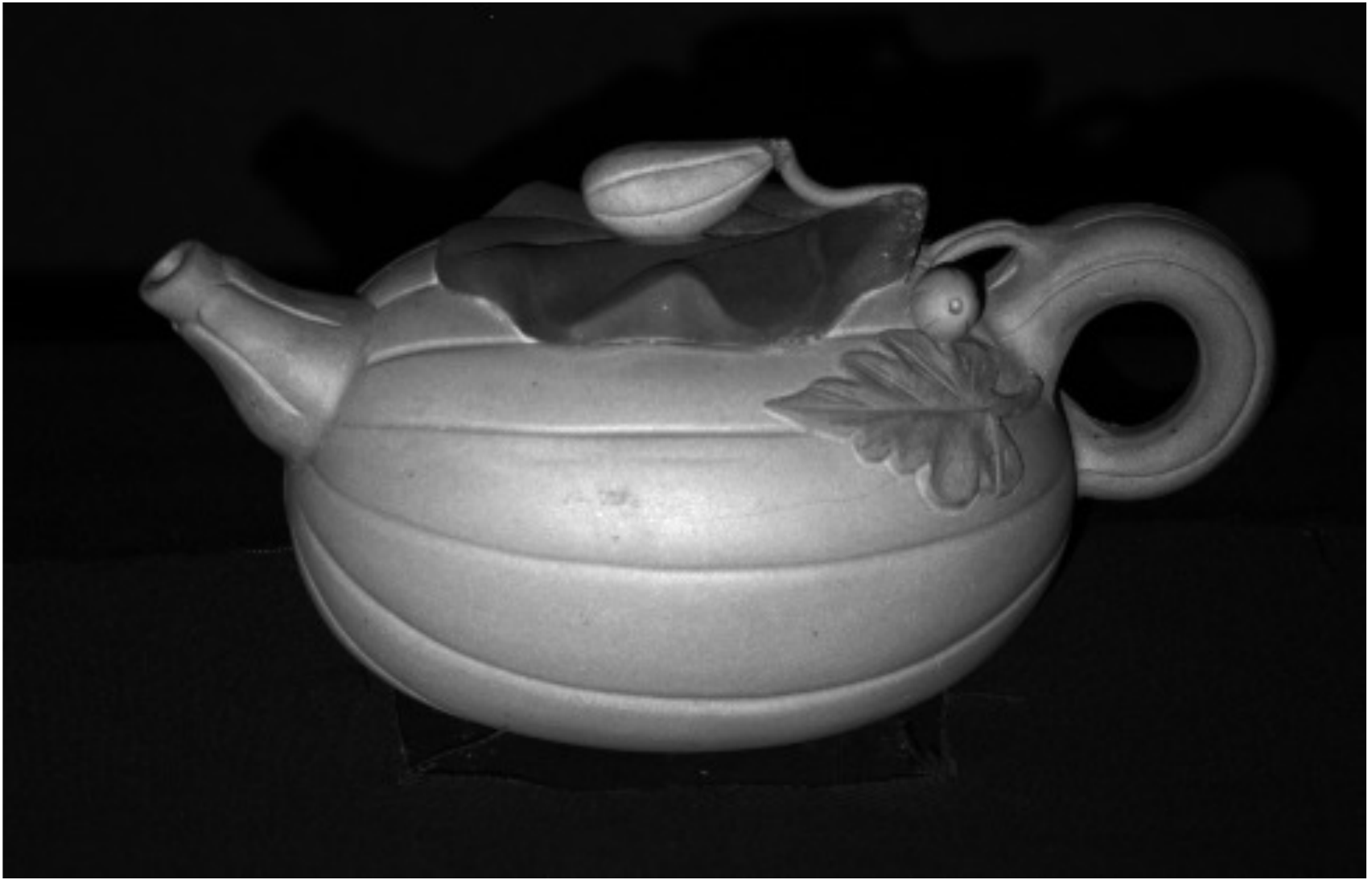}
\end{subfigure}
\hspace{-2mm}
\begin{subfigure}[b]{0.16\textwidth}
  \includegraphics[width=\textwidth]{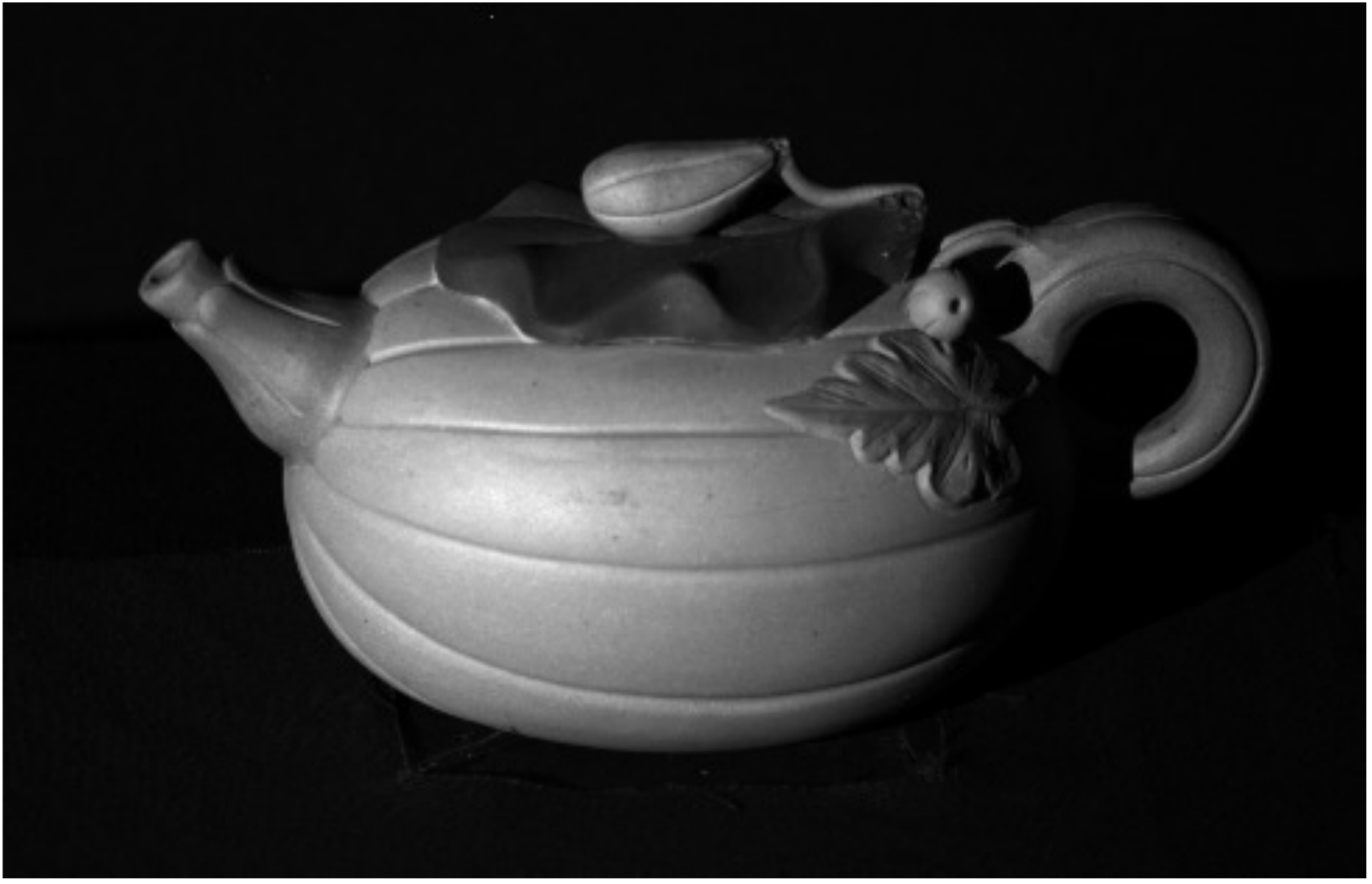}
\end{subfigure}
\hspace{-2mm}
\begin{subfigure}[b]{0.16\textwidth}
  \includegraphics[width=\textwidth]{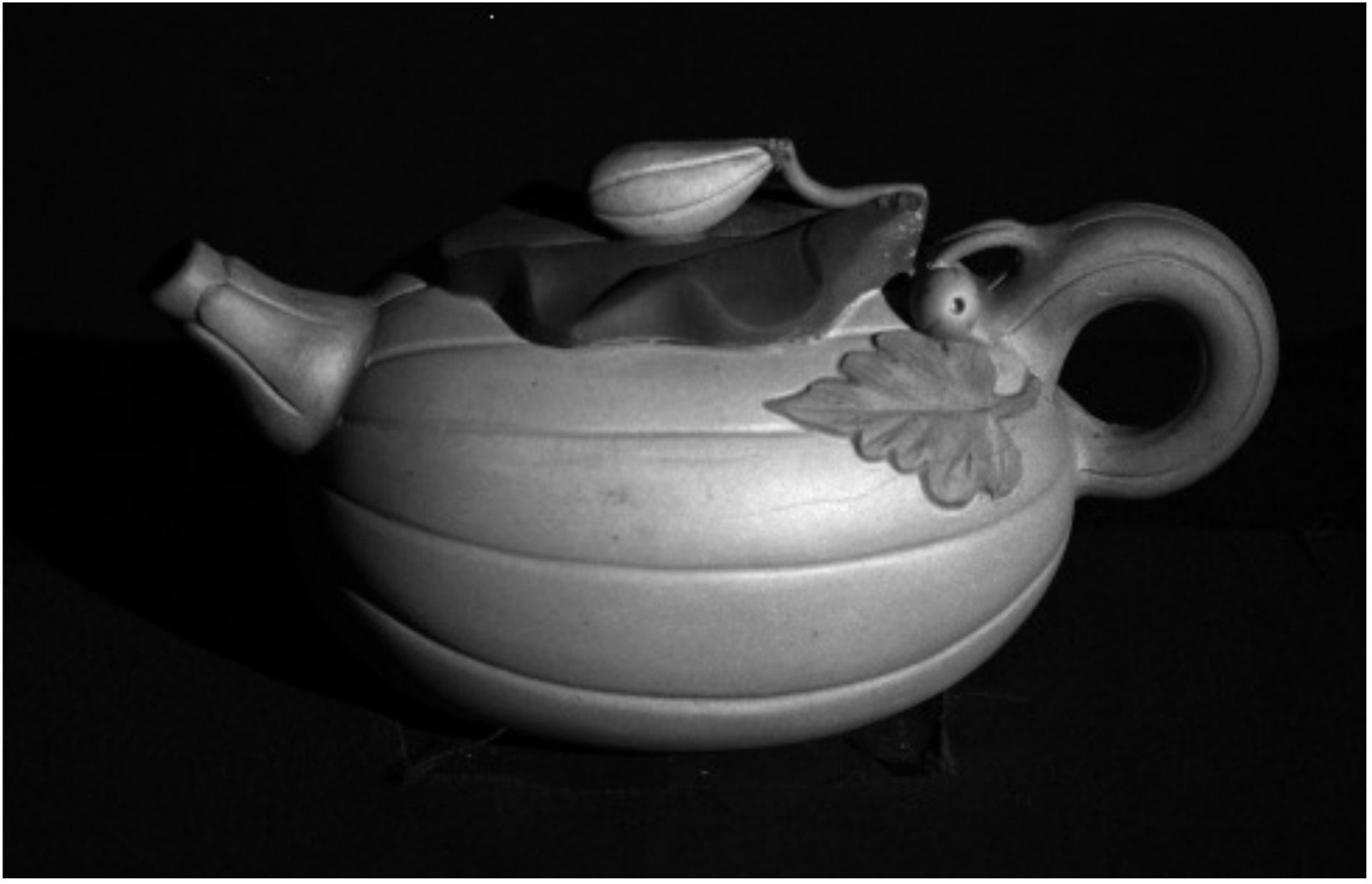}
\end{subfigure}
\caption{\revised{Example images from the DiLiGenT Pot1 dataset used to reconstruct normal vectors.}}
\label{fig:pdlnv_pot1_images}
\end{figure}

\begin{figure}[t!]
\centering
\begin{subfigure}[b]{0.16\textwidth}
  \includegraphics[width=\textwidth]{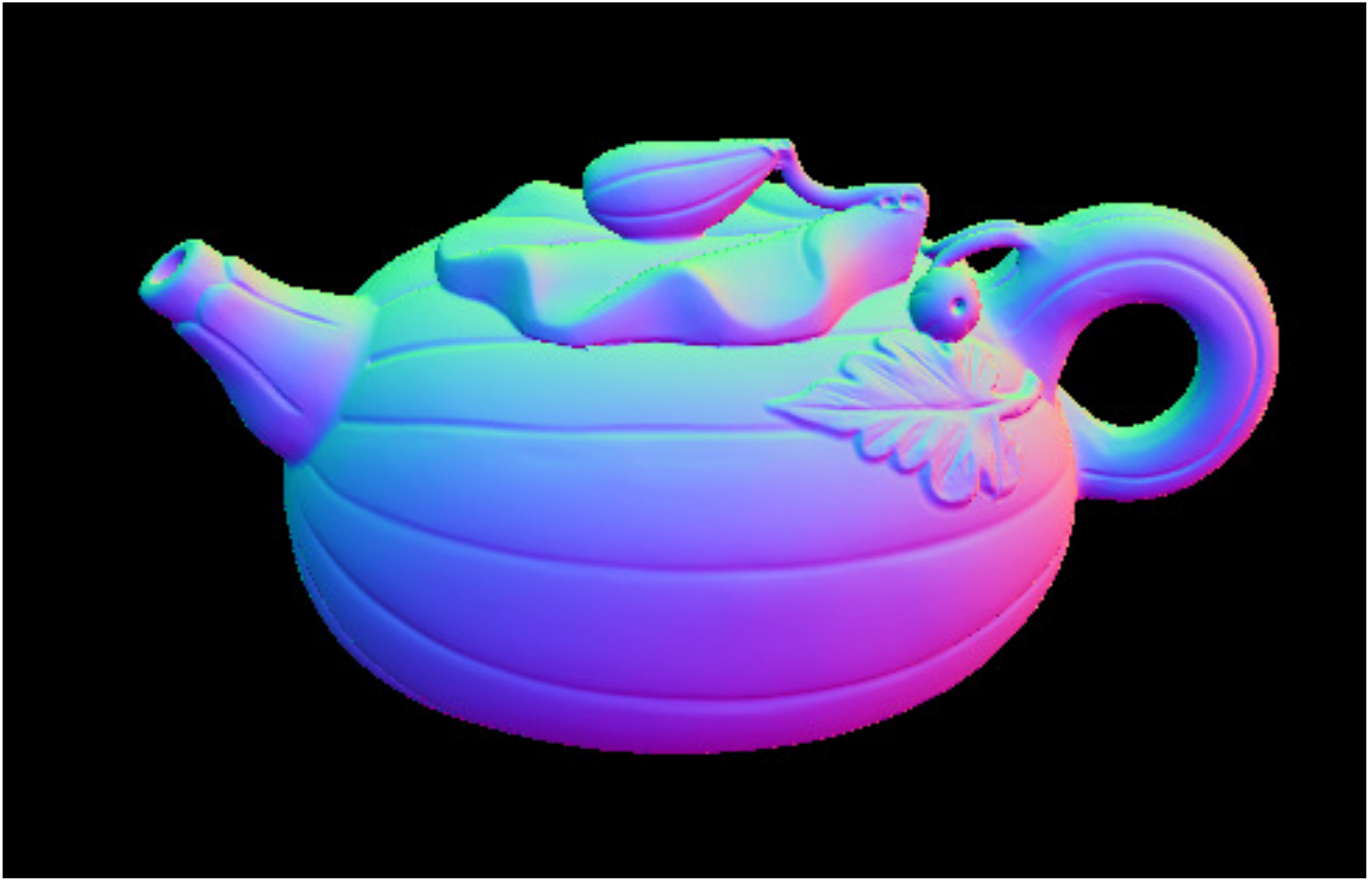}
  \caption{Truth}
\end{subfigure}
\hspace{-2mm}
\begin{subfigure}[b]{0.16\textwidth}
  \includegraphics[width=\textwidth]{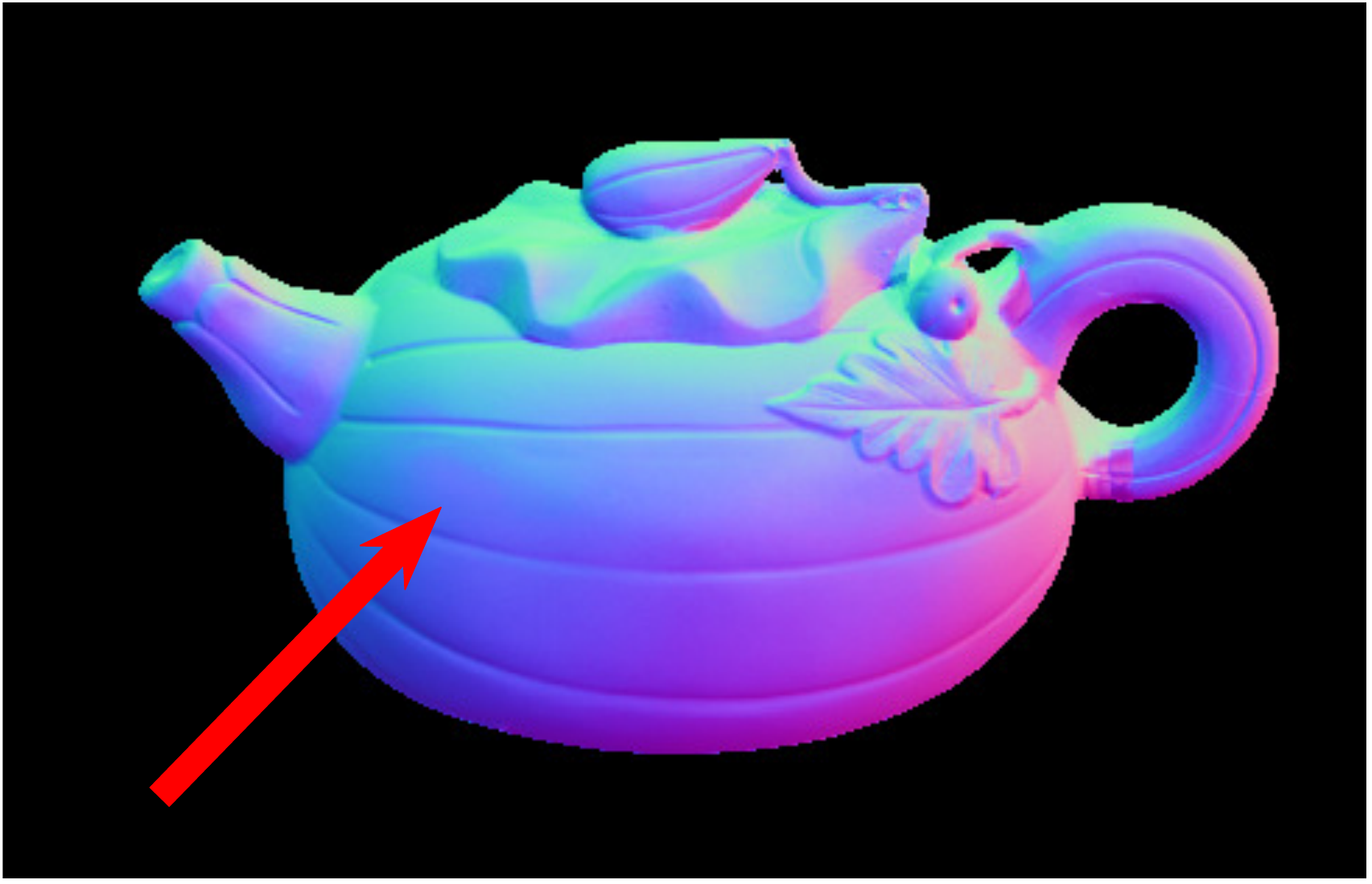}
  \caption{\textbf{PDLNV}}
\end{subfigure}
\hspace{-2mm}
\begin{subfigure}[b]{0.16\textwidth}
  \includegraphics[width=\textwidth]{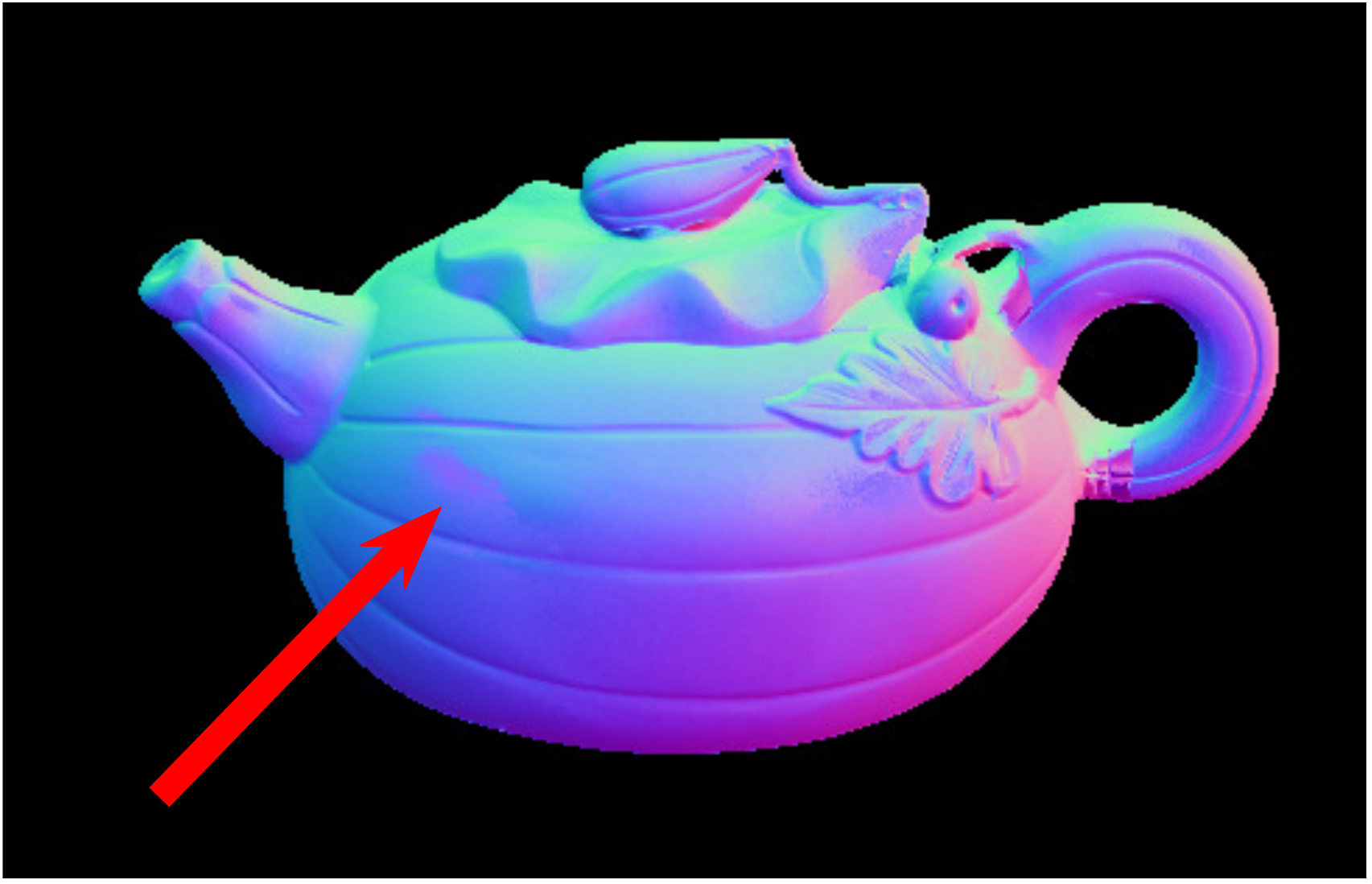}
  \caption{PLS}
\end{subfigure}
\caption{\revised{Normal vector reconstructions for the uncorrupted DiLiGenT Pot1 dataset with 10 images.}}
\label{fig:pdlnv_pot1_normals}
\end{figure}

\begin{figure}[t!]
\centering
\begin{subfigure}[b]{0.16\textwidth}
  \includegraphics[width=\textwidth]{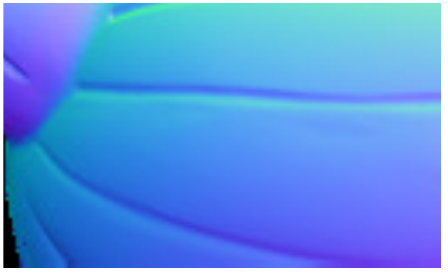}
  \caption{Truth}
\end{subfigure}
\hspace{-2mm}
\begin{subfigure}[b]{0.16\textwidth}
  \includegraphics[width=\textwidth]{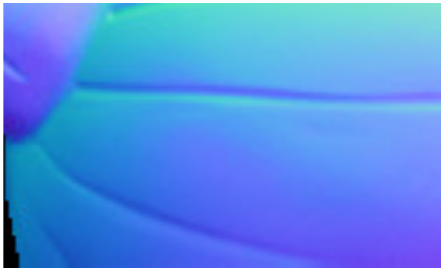}
  \caption{\textbf{PDLNV}}
\end{subfigure}
\hspace{-2mm}
\begin{subfigure}[b]{0.16\textwidth}
  \includegraphics[width=\textwidth]{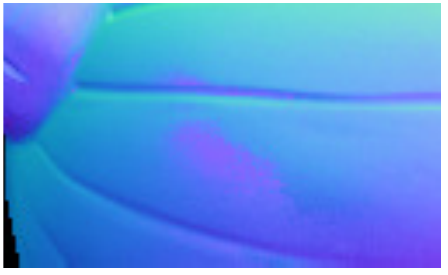}
  \caption{PLS}
\end{subfigure}
\caption{\revised{Close-up view of the spout region from Figure~\ref{fig:pdlnv_pot1_normals}. These images illustrate the ability of dictionary learning to remove corruptions introduced by the piecewise-linear least squares model.}}
\label{fig:pdlnv_pot1_normals_zoom}
\end{figure}

\begin{figure}[t!]
\centering
\begin{subfigure}[b]{0.22\textwidth}
  \includegraphics[width=\textwidth]{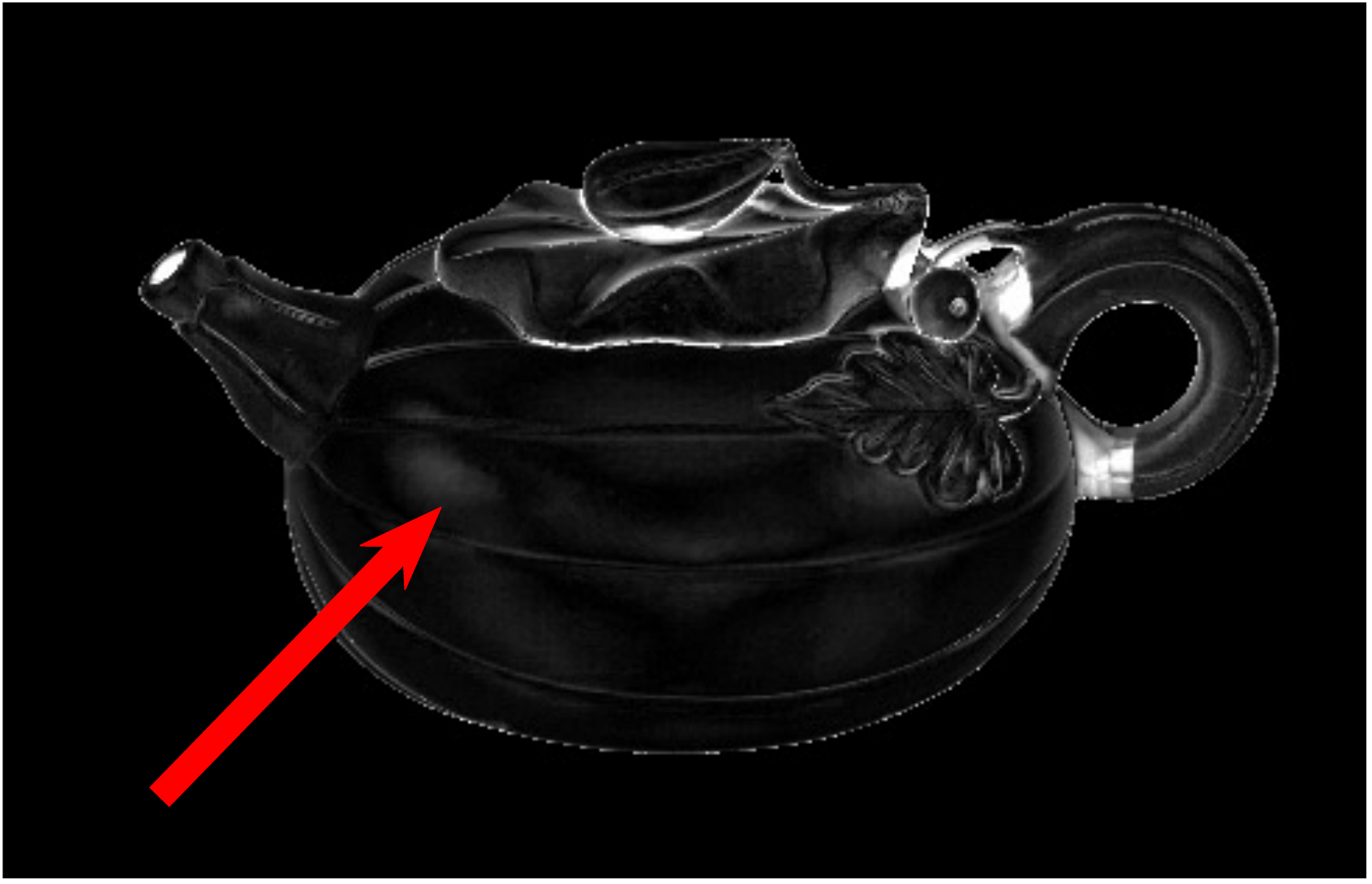}
  \caption{\textbf{PDLNV (7.96)}}
\end{subfigure}
\hspace{-2mm}
\begin{subfigure}[b]{0.22\textwidth}
  \includegraphics[width=\textwidth]{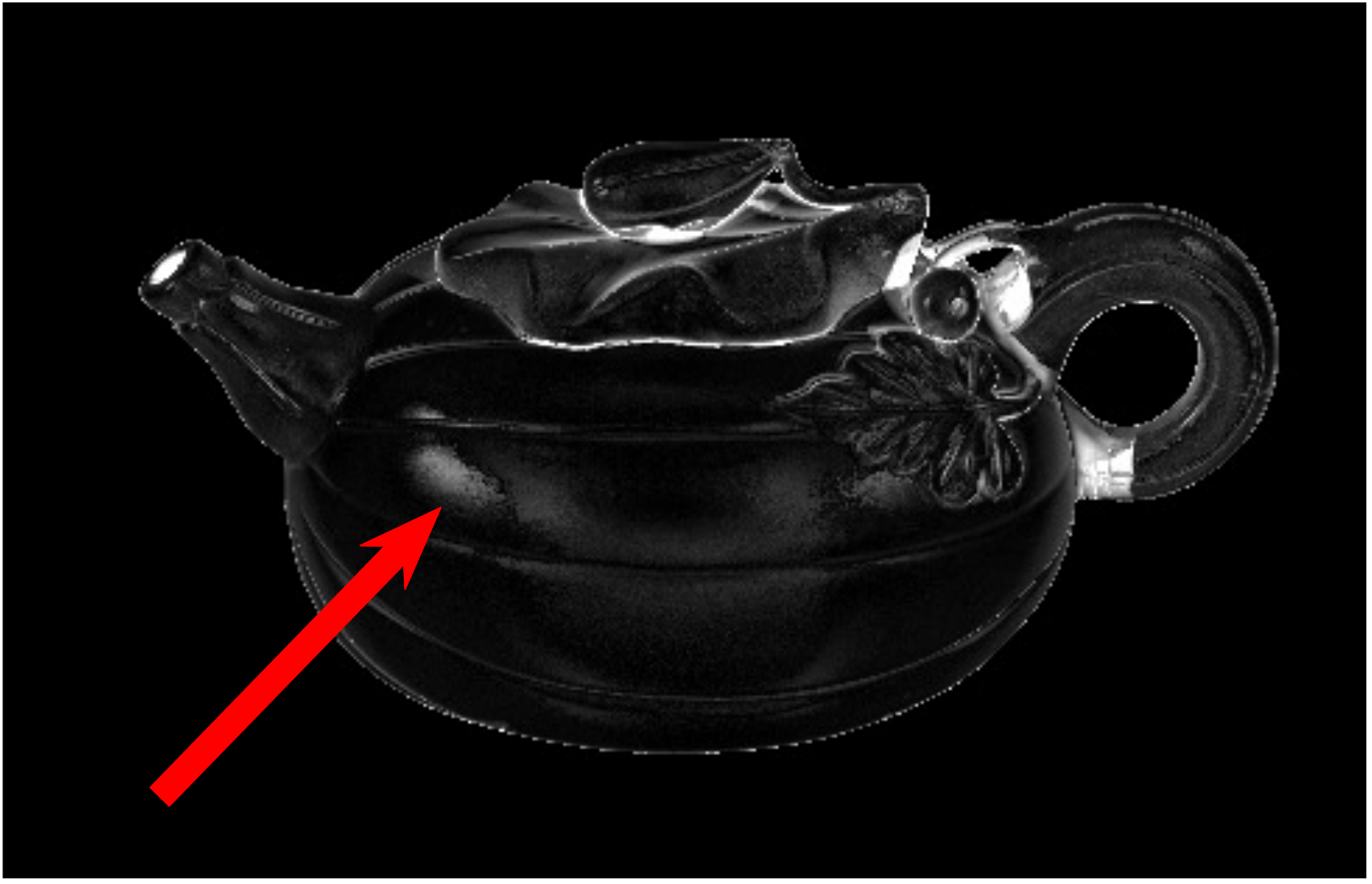}
  \caption{PLS (8.28)}
\end{subfigure}
\hspace{-2mm}
\begin{subfigure}[b]{0.037\textwidth}
\includegraphics[width=\textwidth]{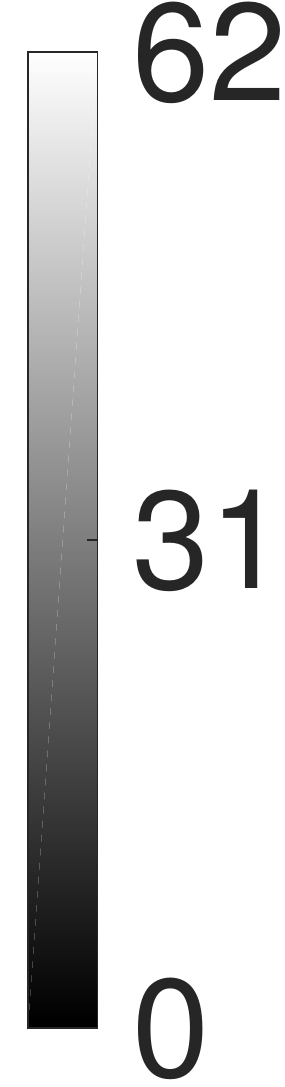}
\caption*{}
\end{subfigure}
\caption{\revised{Normal vector error maps for the uncorrupted DiLiGenT Pot1 dataset with 10 images. Mean angular error (in degrees) for each reconstruction is shown in parentheses.}}
\label{fig:pdlnv_pot1_errors}
\end{figure}

\begin{figure}[t!]
\centering
\includegraphics[width=0.45\textwidth]{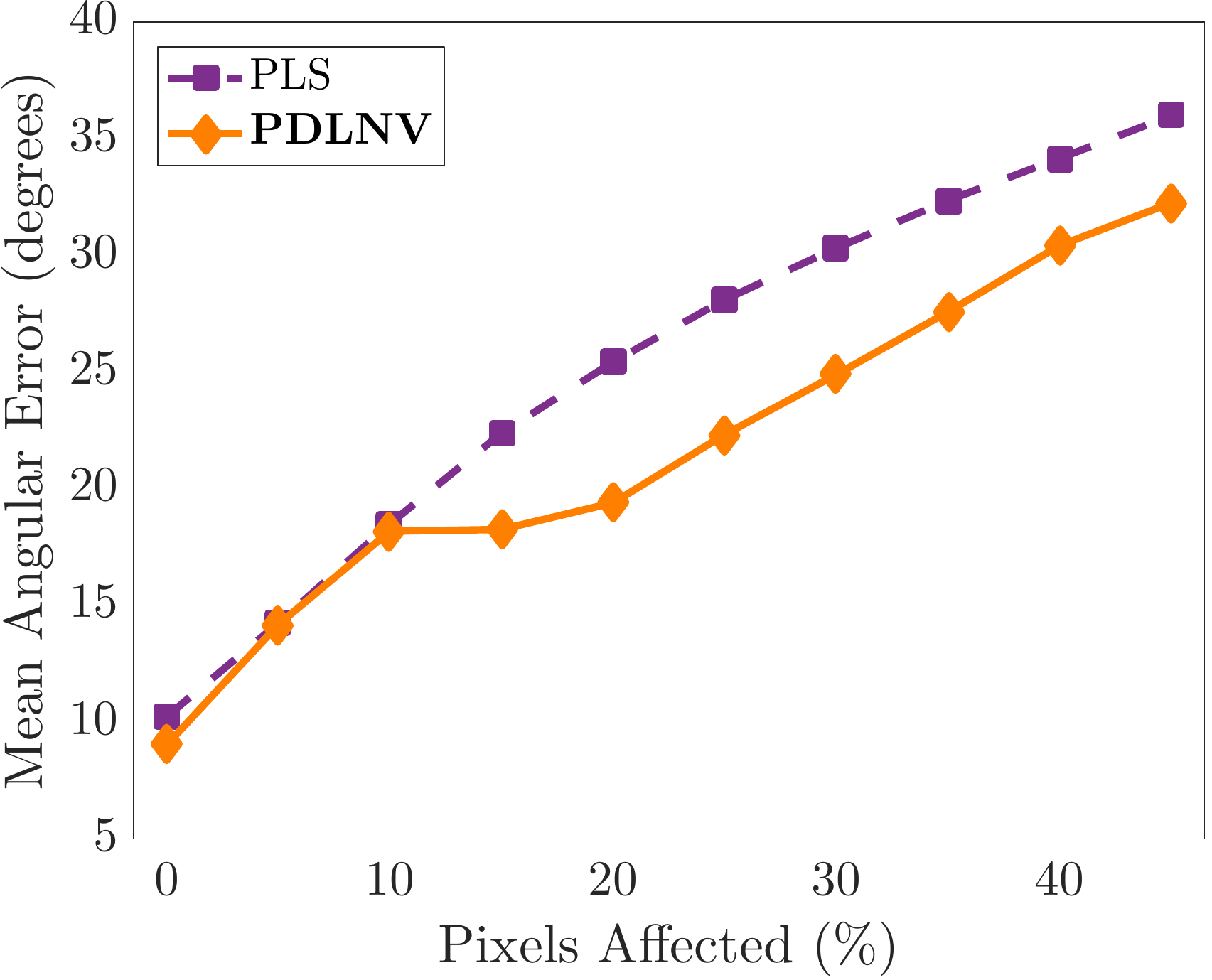}
\caption{\revised{Comparison of the reconstruction errors obtained by PDLNV and PLS on the DiLiGenT Bear dataset with 20 images and salt and pepper noise added. This figure illustrates that dictionary learning significantly improves robustness to added corruptions.}}
\label{fig:pdlnv_bear_sp}
\vspace{-5mm}
\end{figure}

\begin{figure}[t!]
\centering
\includegraphics[width=0.45\textwidth]{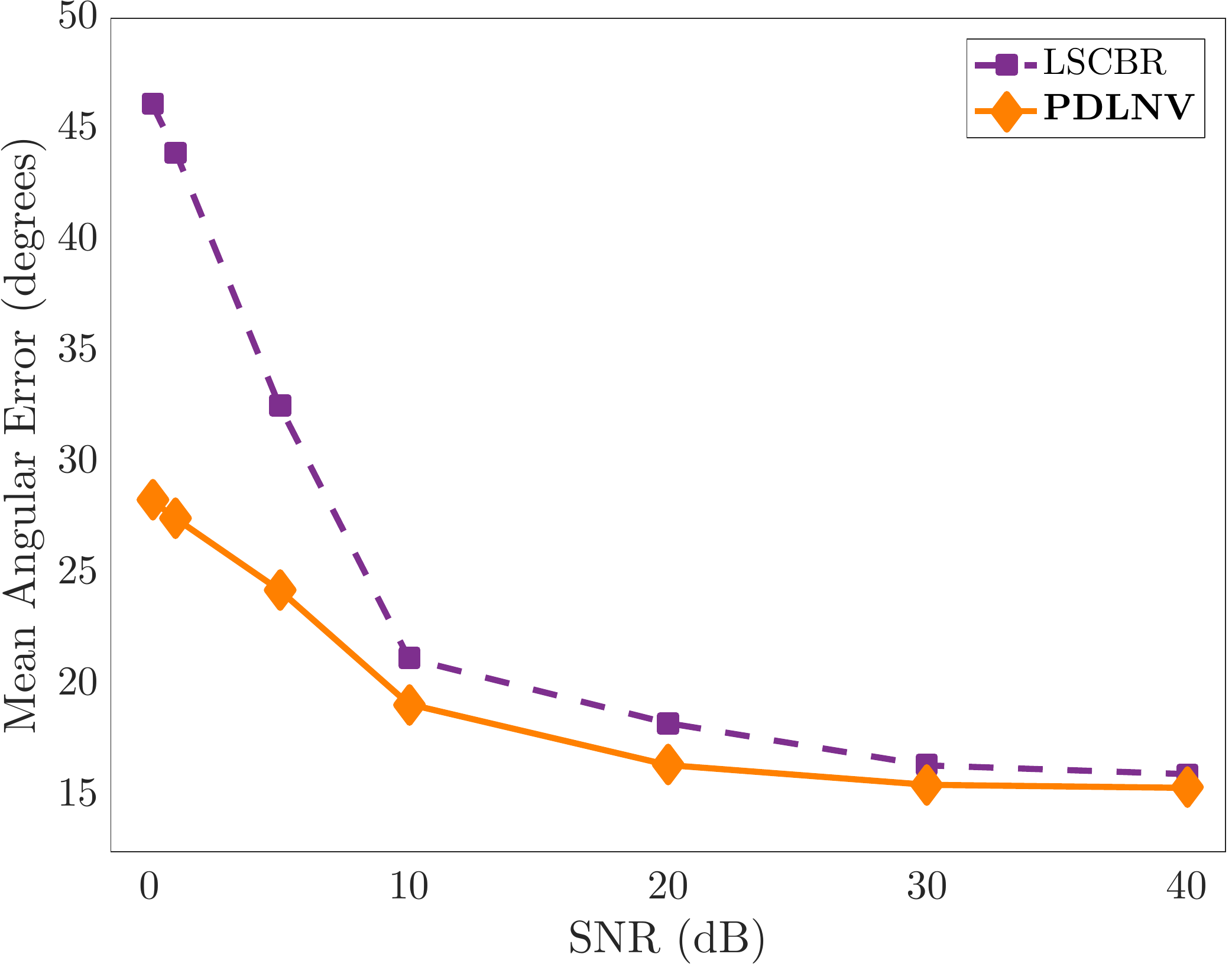}
\caption{\revised{Comparison of reconstruction error obtained by PDLNV and PLS on DiLiGenT Cow dataset with 20 images and Poisson noise added. This figure illustrates that dictionary learning significantly improves robustness to added corruptions.}}
\label{fig:pdlnv_cow_poisson}
\vspace{-5mm}
\end{figure}

Next, we compare the performance of PDLNV and PLS on data corrupted by synthetic noise. We first add salt and pepper noise to the input images, randomly setting a given percentage of the pixels to 0 or 1. The results of this experiment are shown in Figure \ref{fig:pdlnv_bear_sp}. For small amounts of added corruption (10\% of the pixels affected or less), PDLNV and PLS perform comparably. However, for larger amounts of corruption, PDLNV outperforms PLS significantly, by roughly 5 degrees.

Figure \ref{fig:pdlnv_cow_poisson} illustrates the results of a similar experiment with Poisson noise added instead of salt and pepper noise. Here PDLNV again significantly improves on PLS, especially in the low SNR regime.

\begin{table*}[t!]
\begin{center}
\begin{tabular}{|c|c|c|c|c|c|c||c|c|c|c|c|c|}
\cline{1-13}
\multirow{ 2}{*}{Dataset} & \multicolumn{6}{c||}{Mean Angular Error (degrees)} & \multicolumn{6}{c|}{Median Angular Error (degrees)} \\
\cline{2-13}
 & PDLNV & DLNV & CBR & SR & RPCA & LS & PDLNV & DLNV & CBR & SR & RPCA & LS \\
\cline{1-13}
Ball & 3.60  & 3.82 & 6.78 & \textbf{2.08} & 3.20 & 4.10 & 1.95 & \textbf{1.85} & 2.06 & 2.02 & 2.02 & 2.41\\
\cline{1-13}
Cat & \textbf{6.40} & 8.10 & 8.05 & 6.73 & 7.96 & 8.41 & \textbf{3.58} & 6.15 & 3.88 & 5.75 & 6.03 & 6.52 \\
\cline{1-13}
Pot1 & \textbf{6.99} & 8.67 & 8.57 & 7.24 & 8.81 & 8.89 &  \textbf{3.70} & 6.39 & 4.15 & 5.29 & 6.61 & 6.65 \\
\cline{1-13}
Bear & 8.51 & 8.32 & 9.77 & \textbf{6.01} & 7.89 & 8.39 & 6.34 & 6.20 & 7.07 & \textbf{4.30} & 6.01 & 6.18 \\
\cline{1-13}
Pot2 & \textbf{9.94} & 13.88 & 10.56 & 11.98 & 11.94 & 14.65 & \textbf{7.17} & 12.79 & \textbf{6.95} & 8.32 & 10.07 & 11.60 \\
\cline{1-13}
Buddha & 13.56 & 14.72 & 14.90 & \textbf{11.11} & 13.88 & 14.92 & 7.84 & 10.33 & 8.85 & \textbf{7.74} & 9.25 & 10.54 \\
\cline{1-13}
Goblet & \textbf{14.58} & 17.69 & 15.10 & 15.53 & 15.14 & 18.50 & 9.56 & 16.04 & \textbf{9.35} & 12.23 & 11.35 & 15.70 \\
\cline{1-13}
Reading & 20.18 & 19.58 & 19.39 & \textbf{12.56} & 17.42 & 19.80 & 14.85 & 12.51 & 13.99 & \textbf{7.18} & 11.64 & 12.50 \\
\cline{1-13}
Cow & 13.71 & 17.58 & 15.68 & 22.42 & \textbf{11.96} & 25.60 & 10.11 & 12.30 & 13.95 & 21.32 & \textbf{9.82} & 26.32 \\
\cline{1-13}
Harvest & \textbf{20.54} & 27.07 & 26.93 & 26.80 & 25.50 & 30.62 & \textbf{15.28} & 23.34 & 22.21 & 19.00 & 20.32 & 25.33 \\
\hhline{=============}
\textit{Average} & \textbf{\textit{11.80}} & \textit{13.94} & \textit{13.57} & \textit{12.25} & \textit{12.37} & \textit{15.39} & \textbf{\textit{8.04}} & \textit{10.79} & \textit{9.25} & \textit{9.32} & \textit{9.31} & \textit{12.38} \\
\cline{1-13}
\end{tabular}
\end{center}
\caption{Mean and median angular errors (in degrees) of the estimated normal vectors for the full, uncorrupted DiLiGenT datasets. \revised{PDLNV achieves state-of-the-art results on five of the ten DiLiGenT objects}.}
\label{tab:full_DiLiGenT}
\end{table*}

\begin{figure*}[t!]
\centering
\begin{subfigure}[b]{0.14\textwidth}
  \includegraphics[width=\textwidth]{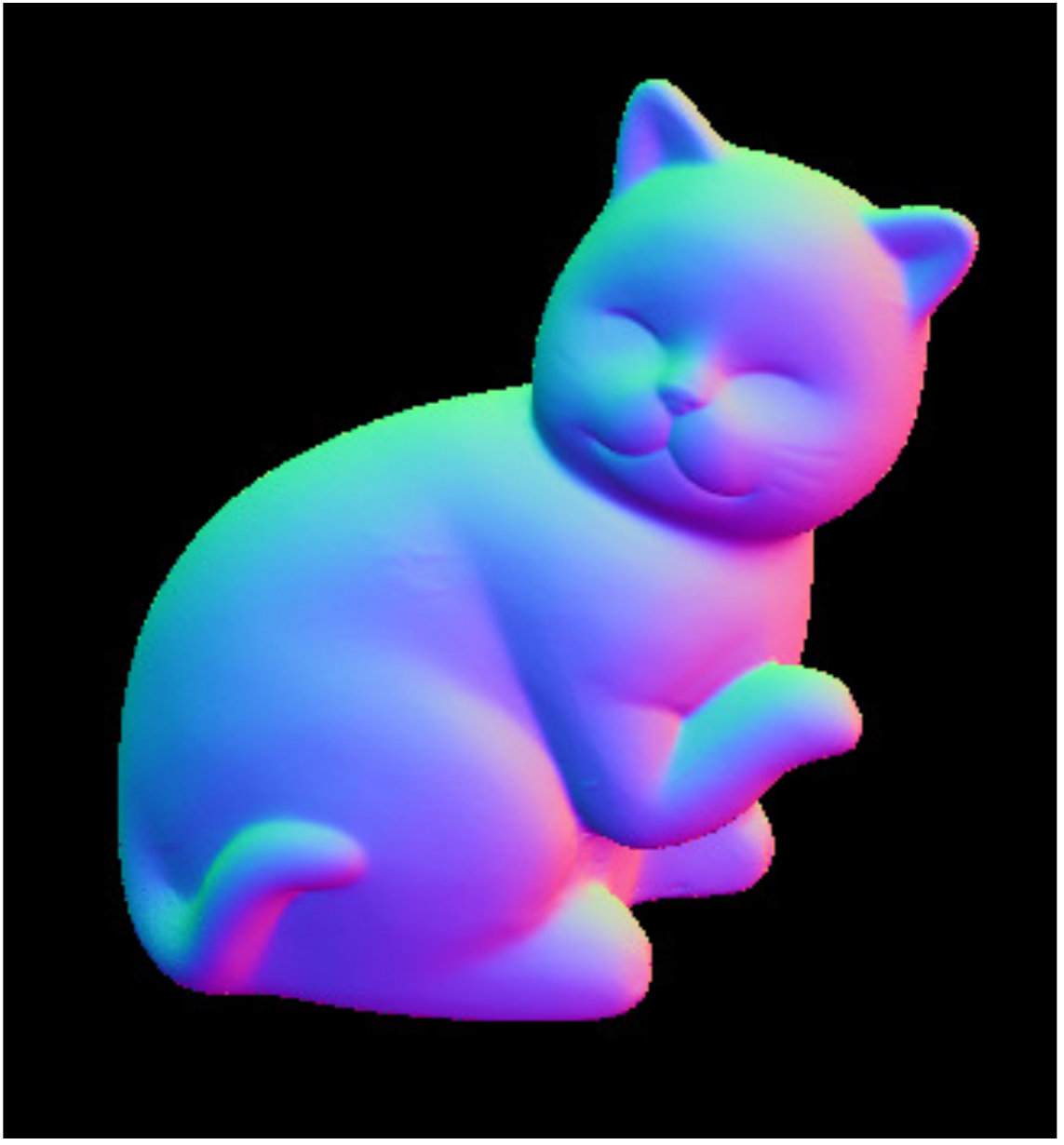}
  \caption{Truth}
\end{subfigure}
\hspace{-2mm}
\begin{subfigure}[b]{0.14\textwidth}
  \includegraphics[width=\textwidth]{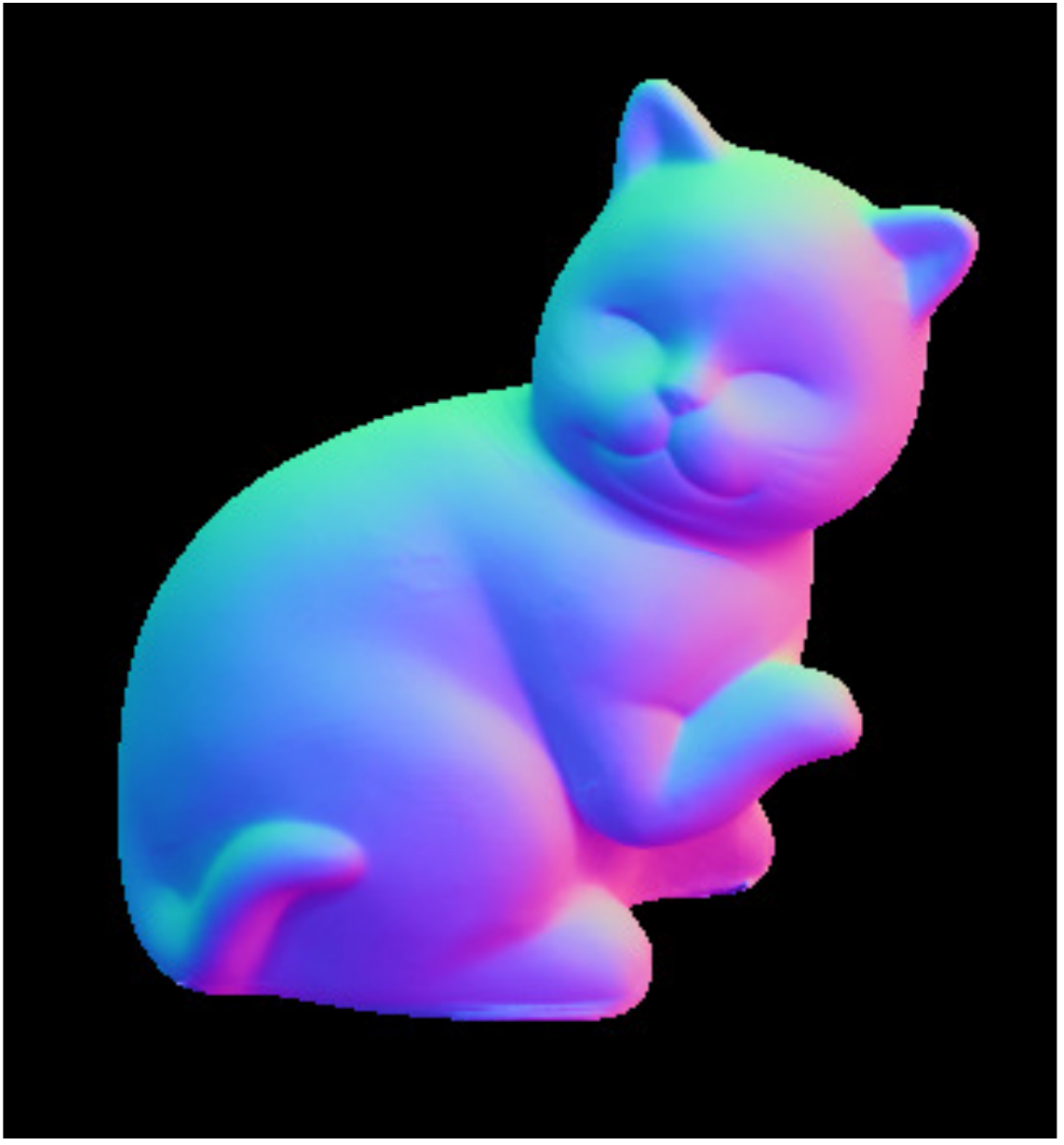}
  \caption{\textbf{PDLNV}}
\end{subfigure}
\hspace{-2mm}
\begin{subfigure}[b]{0.14\textwidth}
  \includegraphics[width=\textwidth]{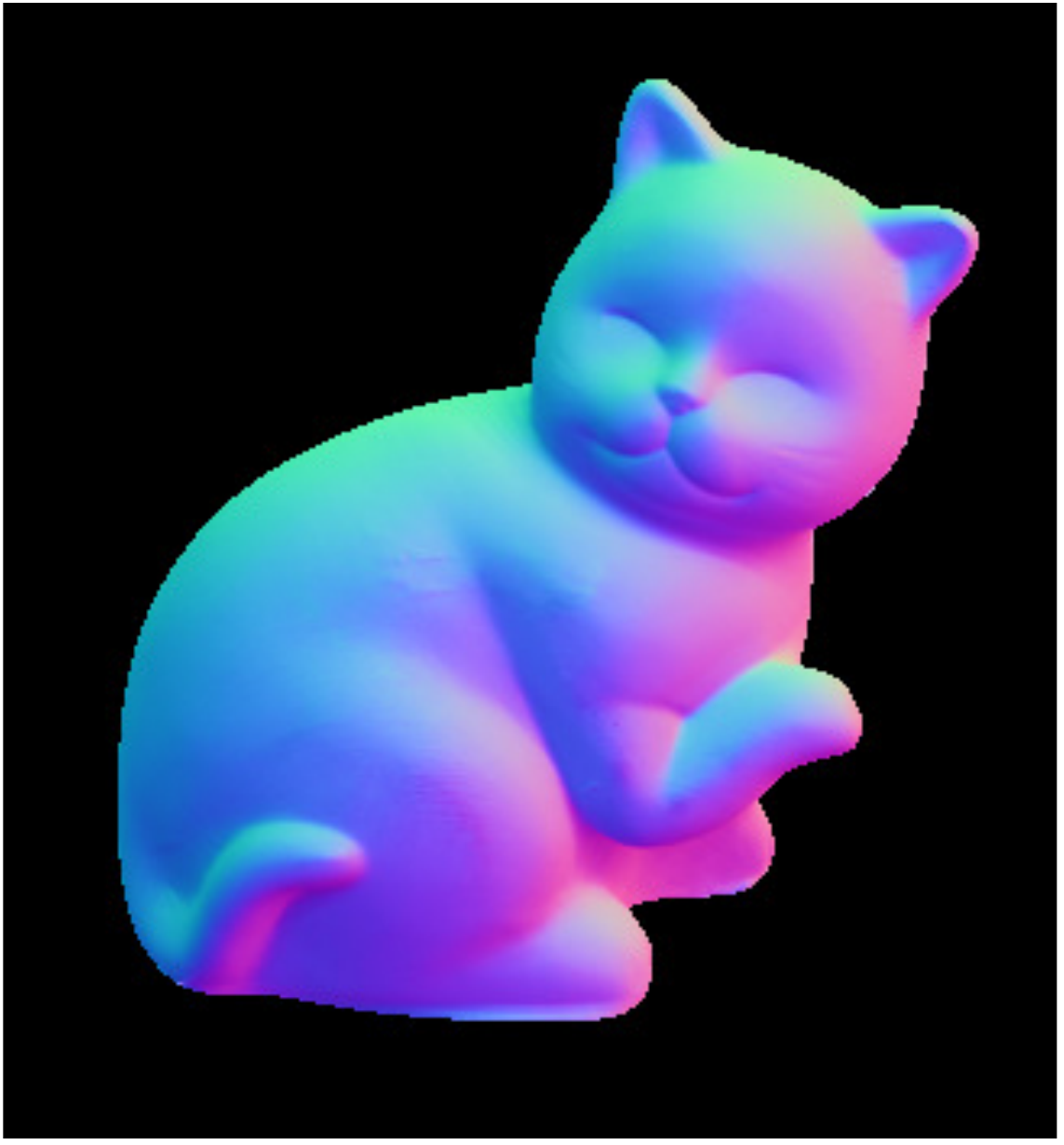}
  \caption{\textbf{DLNV}}
\end{subfigure}
\hspace{-2mm}
\begin{subfigure}[b]{0.14\textwidth}
  \includegraphics[width=\textwidth]{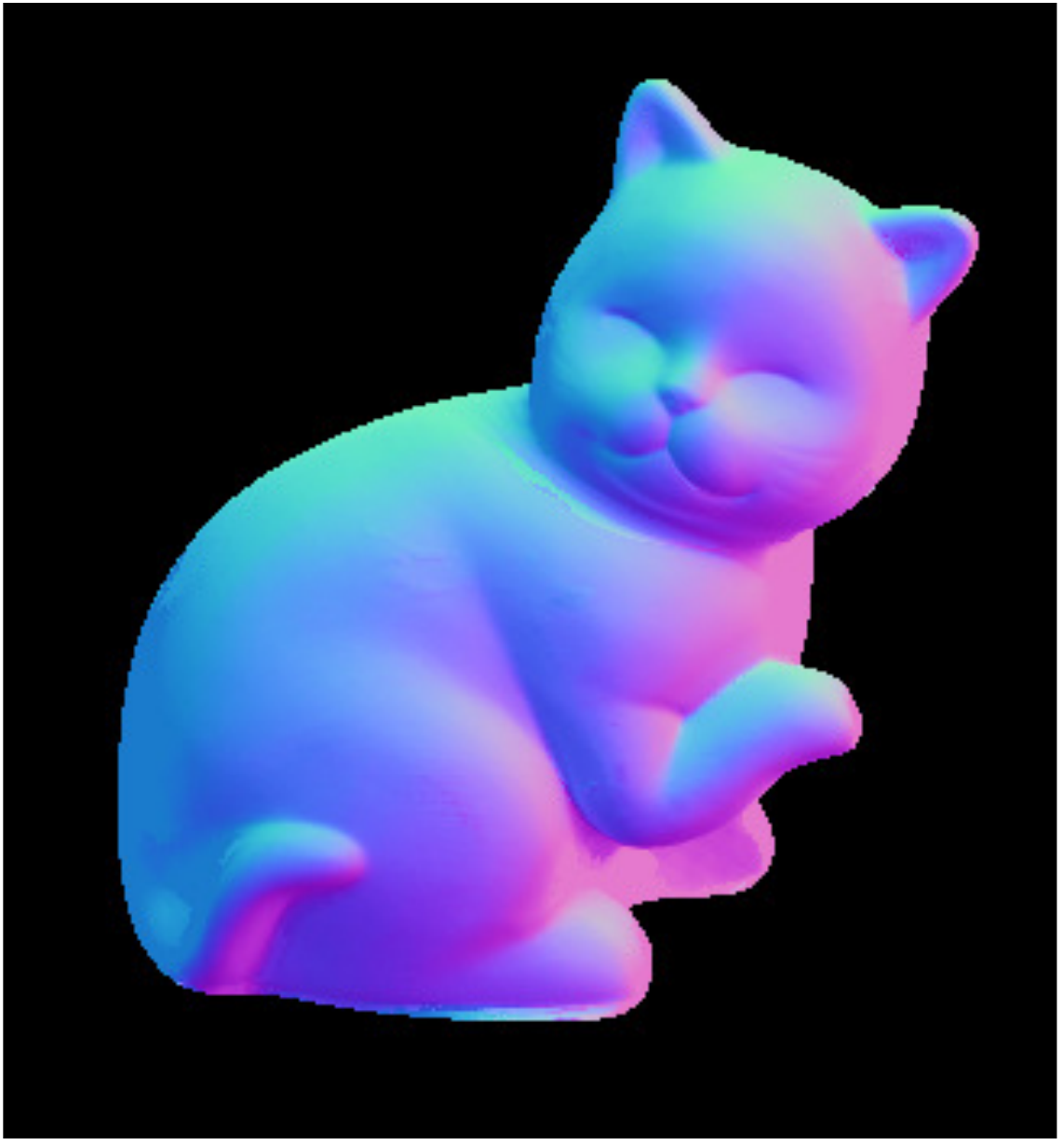}
  \caption{CBR}
\end{subfigure}
\hspace{-2mm}
\begin{subfigure}[b]{0.14\textwidth}
  \includegraphics[width=\textwidth]{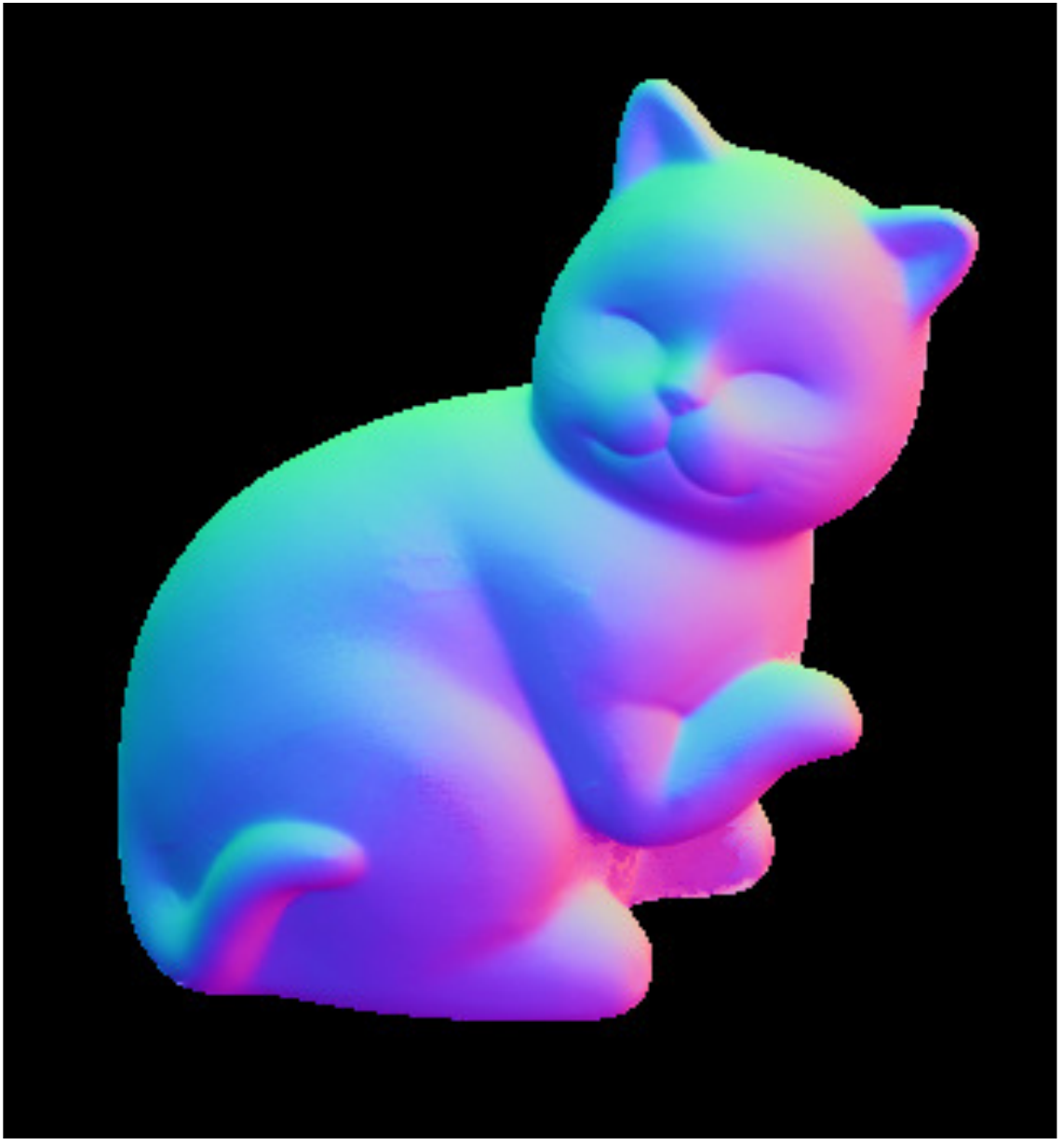}
  \caption{SR}
\end{subfigure}
\hspace{-2mm}
\begin{subfigure}[b]{0.14\textwidth}
  \includegraphics[width=\textwidth]{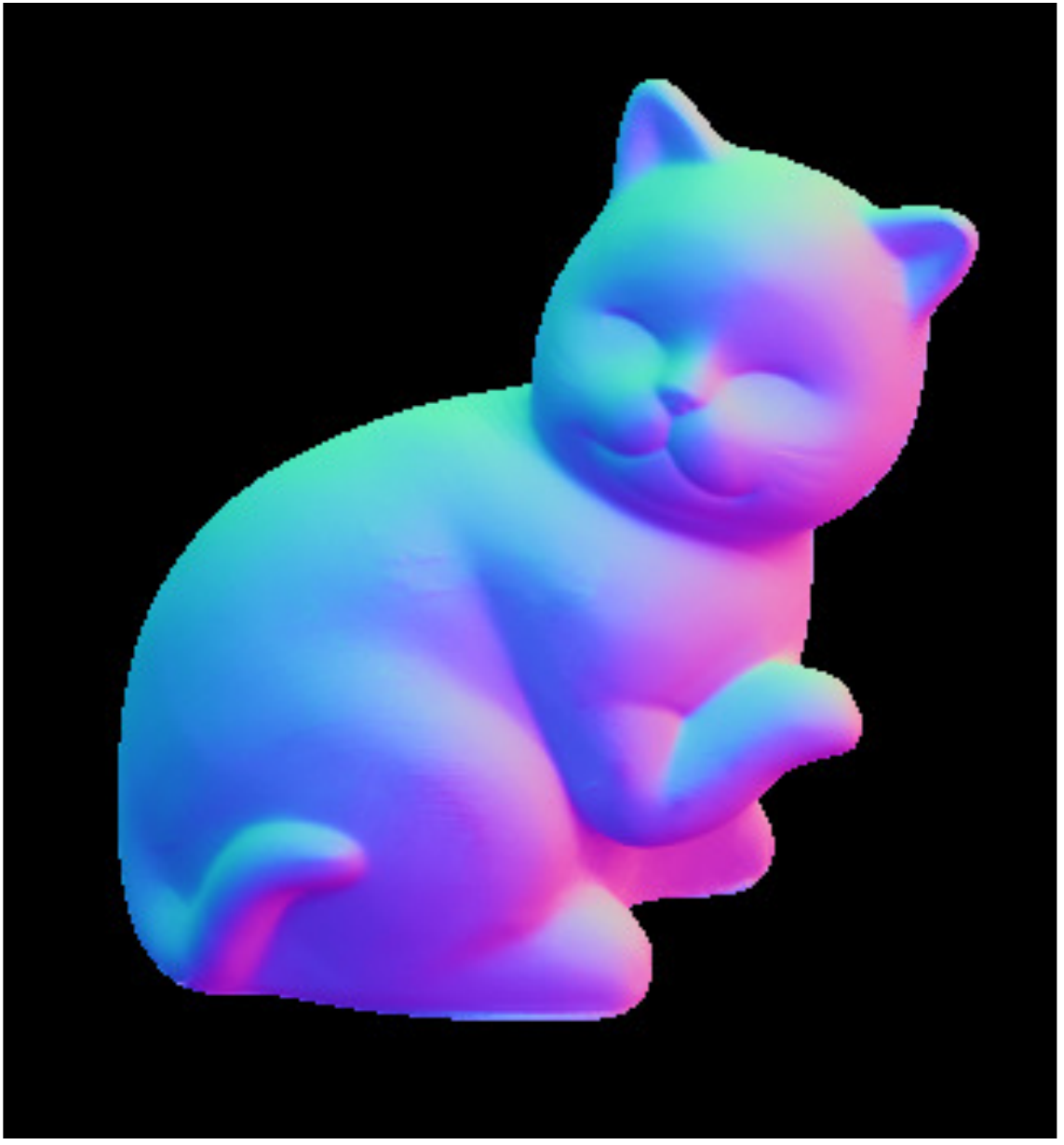}
  \caption{RPCA}
\end{subfigure}
\hspace{-2mm}
\begin{subfigure}[b]{0.14\textwidth}
  \includegraphics[width=\textwidth]{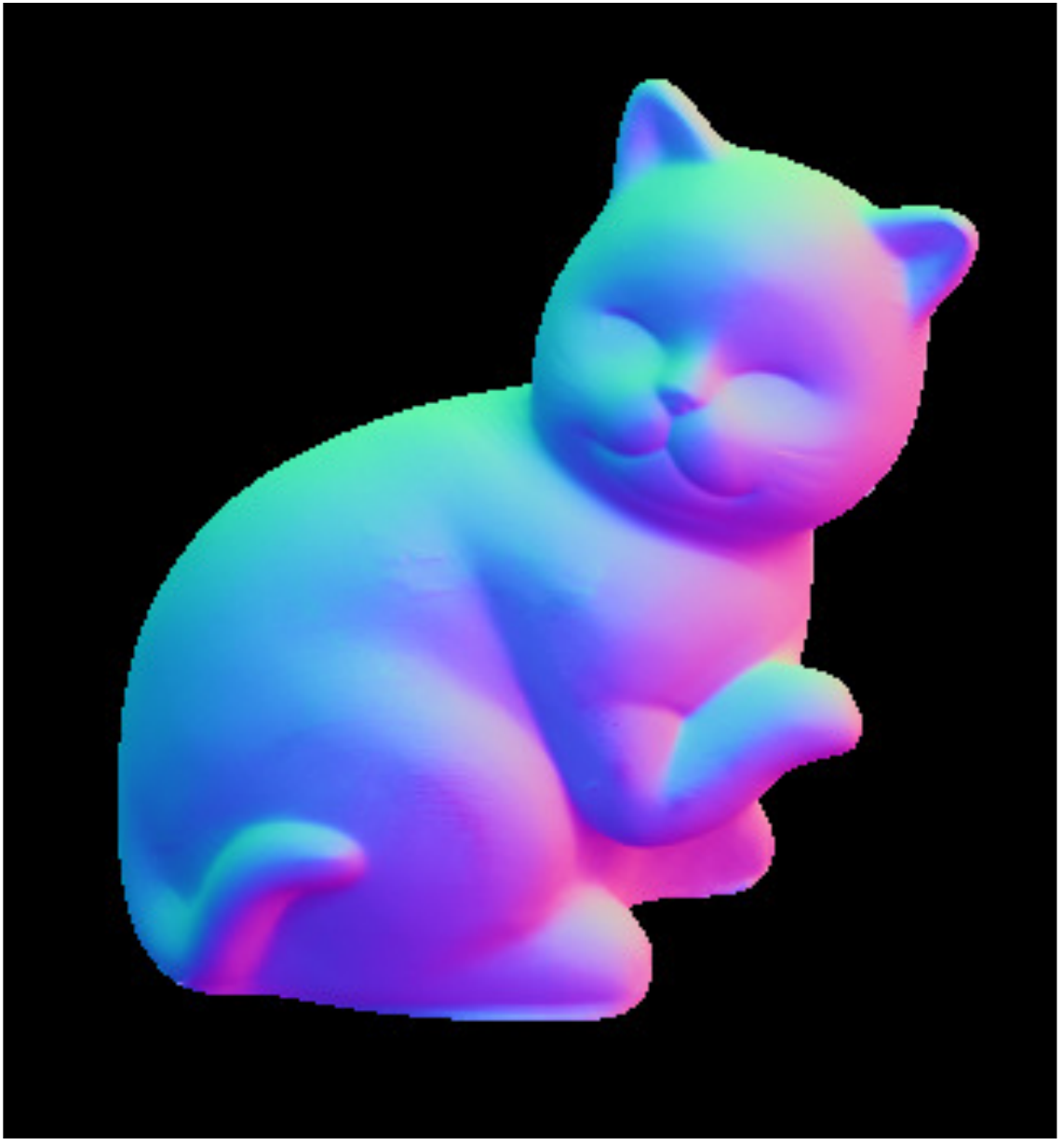}
  \caption{LS}
\end{subfigure}
\caption{\revised{Normal vector reconstructions for the full, uncorrupted DiLiGenT Cat dataset.}}
\label{fig:cat_normals}
\end{figure*}

\begin{figure*}[t!]
\centering
\begin{subfigure}[b]{0.15\textwidth}
  \includegraphics[width=\textwidth]{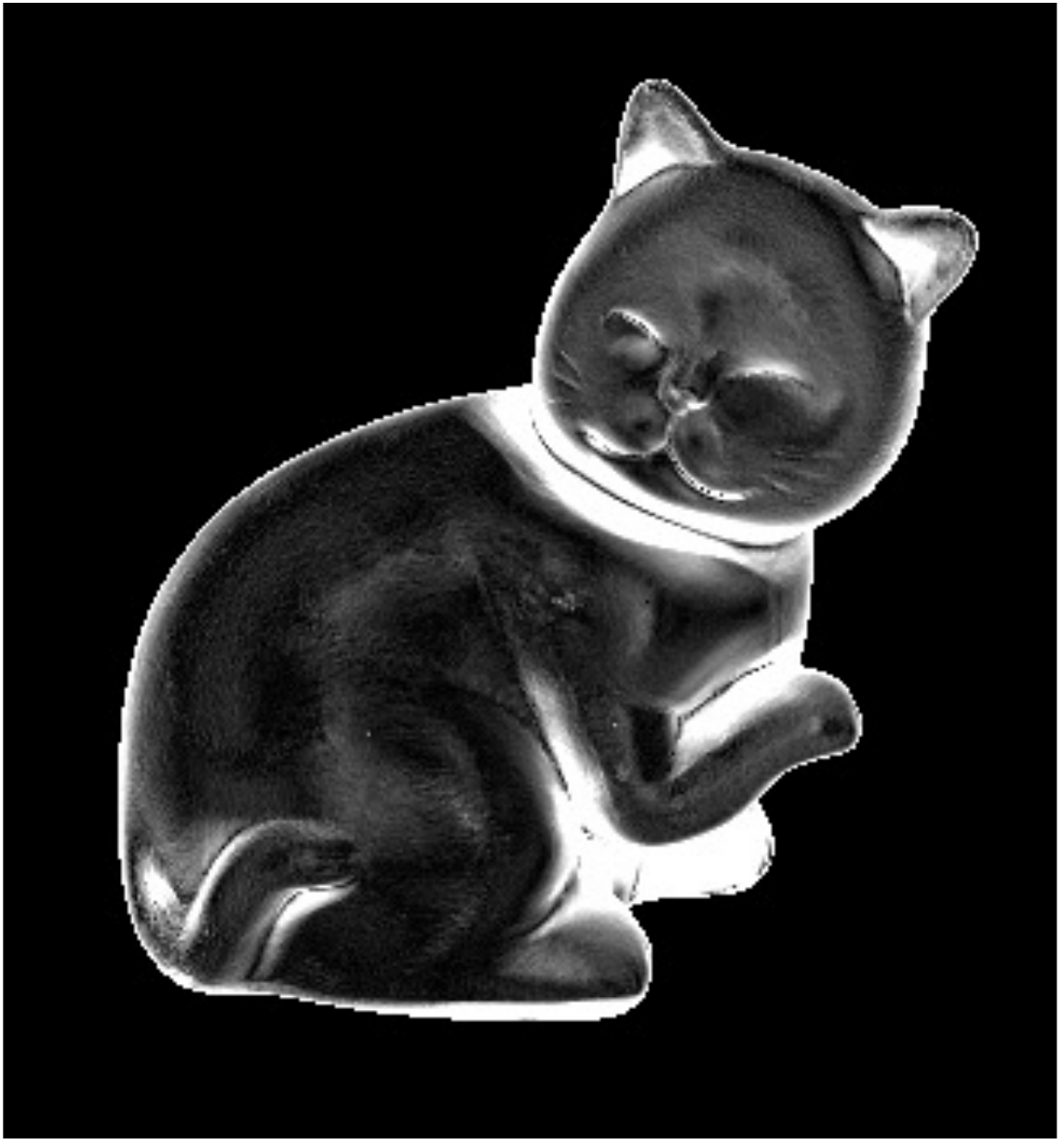}
  \caption{\textbf{PDLNV} (6.40)}
\end{subfigure}
\hspace{-2mm}
\begin{subfigure}[b]{0.15\textwidth}
  \includegraphics[width=\textwidth]{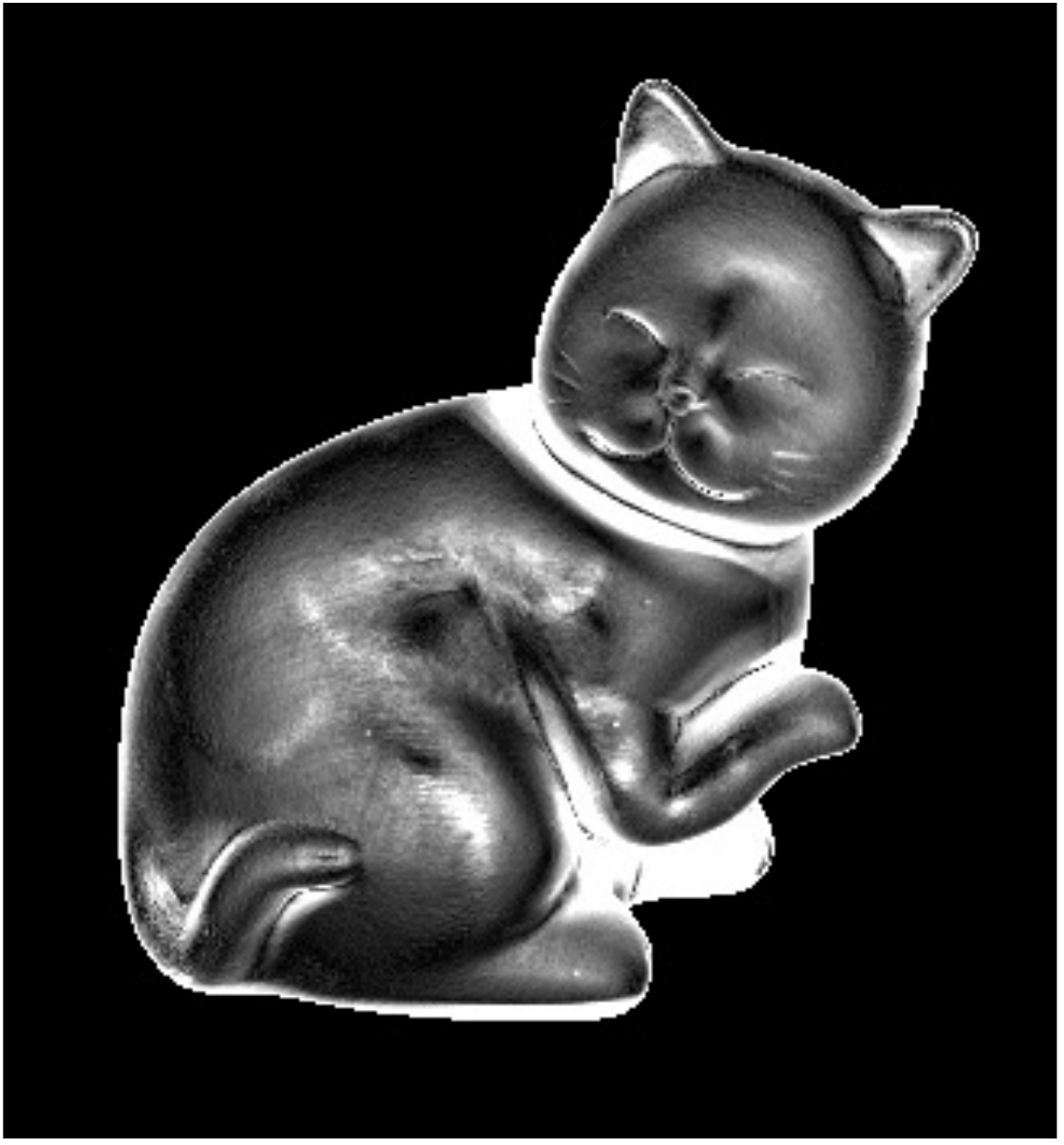}
  \caption{\textbf{DLNV} (8.10)}
\end{subfigure}
\hspace{-2mm}
\begin{subfigure}[b]{0.15\textwidth}
  \includegraphics[width=\textwidth]{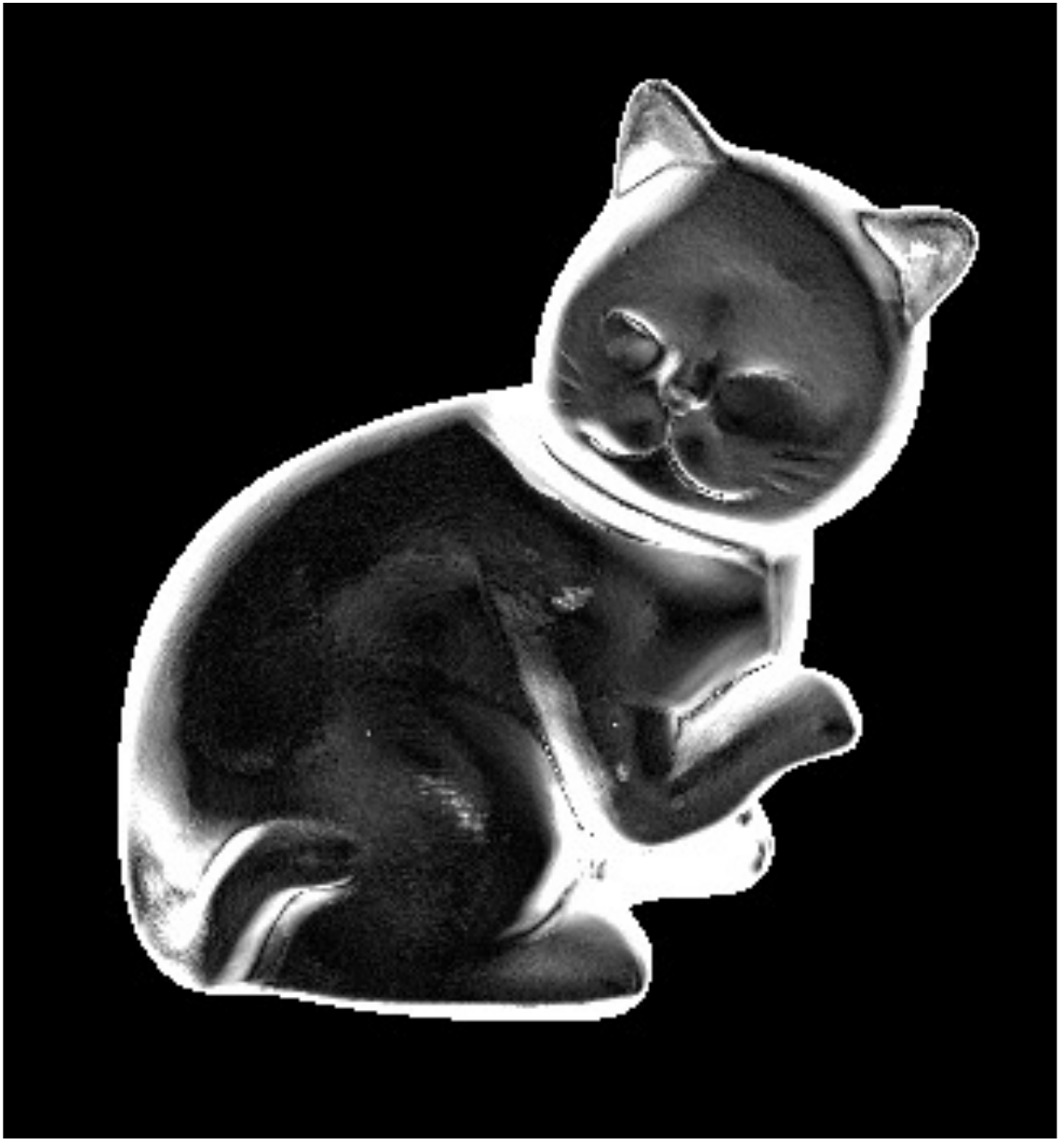}
  \caption{CBR (8.05)}
\end{subfigure}
\hspace{-2mm}
\begin{subfigure}[b]{0.15\textwidth}
  \includegraphics[width=\textwidth]{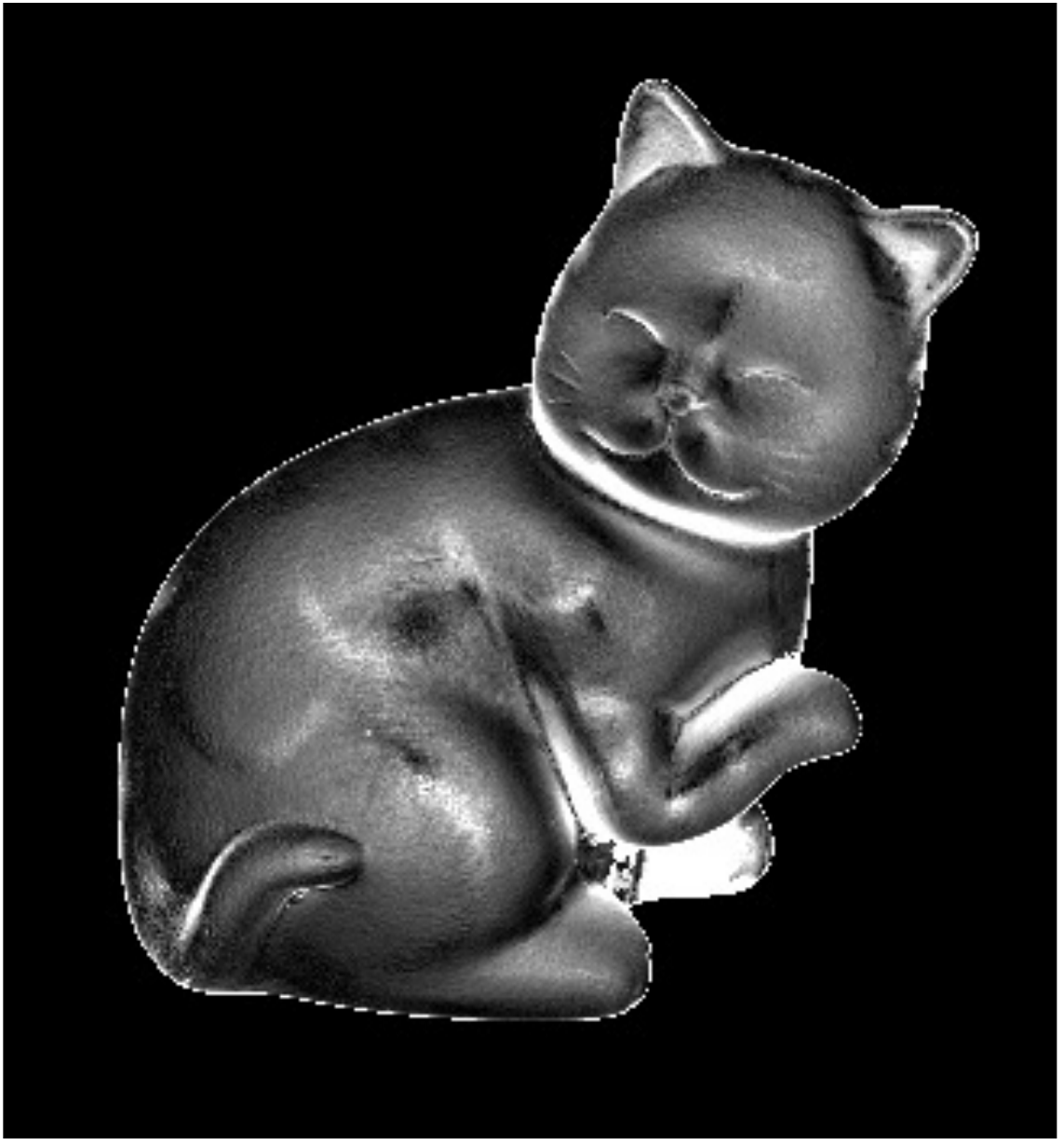}
  \caption{SR (6.73)}
\end{subfigure}
\hspace{-2mm}
\begin{subfigure}[b]{0.15\textwidth}
  \includegraphics[width=\textwidth]{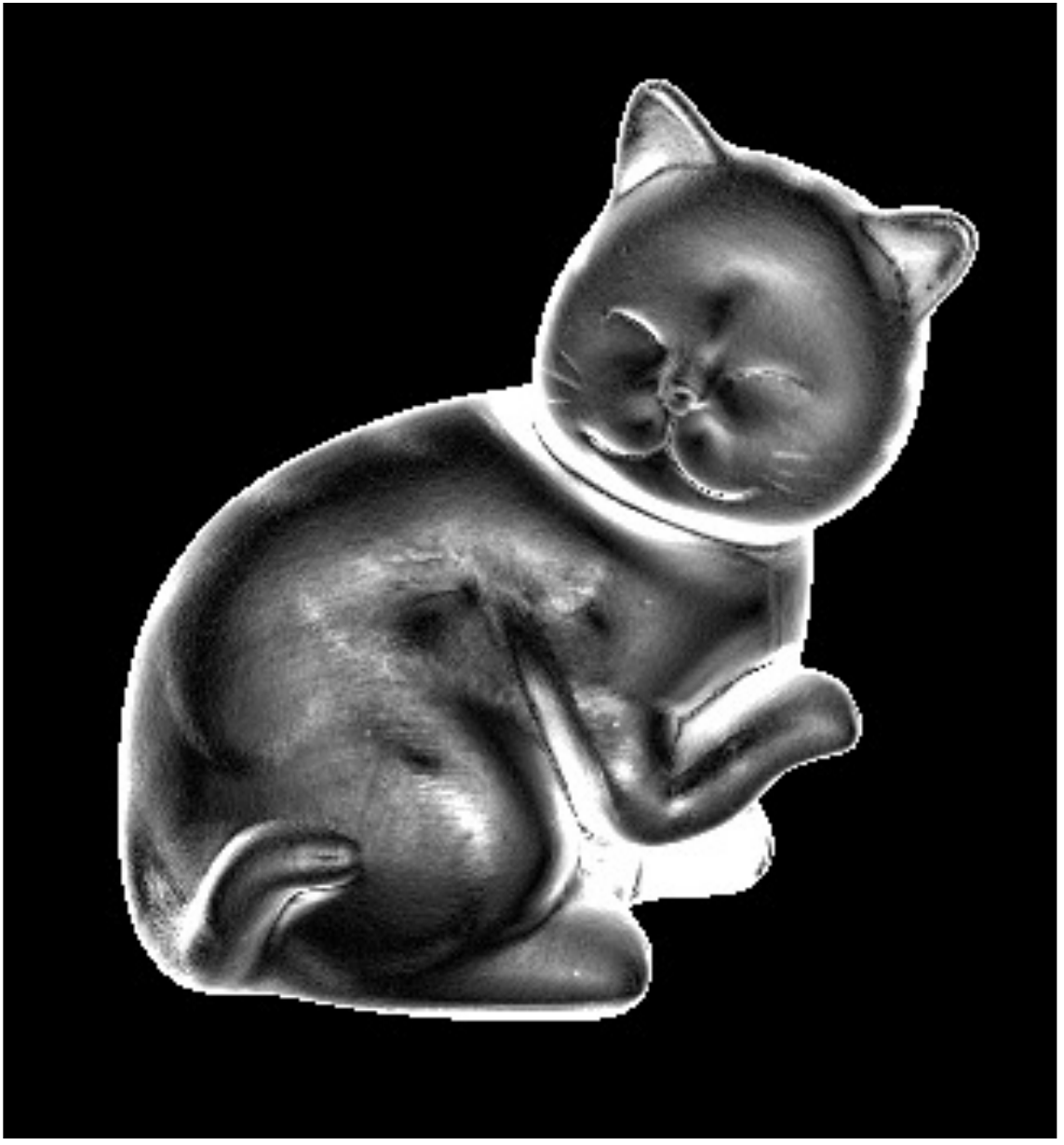}
  \caption{RPCA (7.96)}
\end{subfigure}
\hspace{-2mm}
\begin{subfigure}[b]{0.15\textwidth}
  \includegraphics[width=\textwidth]{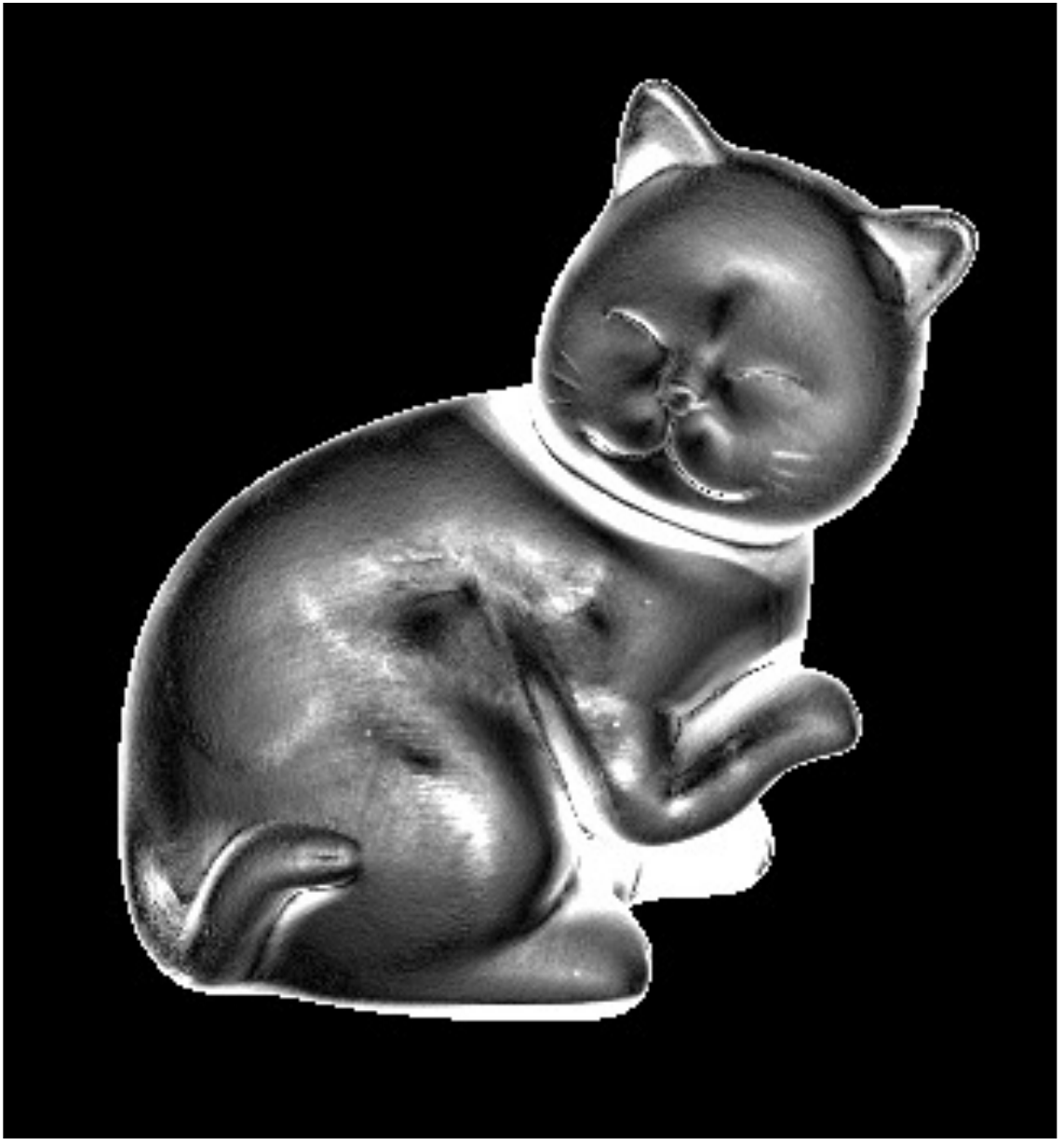}
  \caption{LS (8.41)}
\end{subfigure}
\hspace{-2mm}
\begin{subfigure}[b]{0.04\textwidth}
\includegraphics[width=\textwidth]{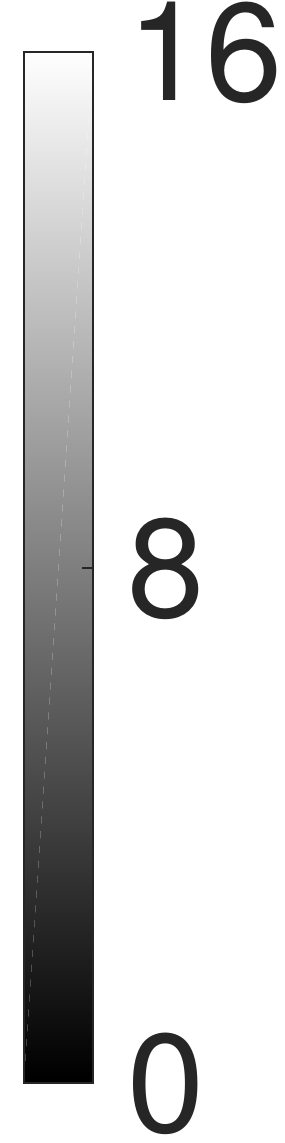}
\caption*{}
\end{subfigure}
\caption{\revised{Normal vector error maps for the full, uncorrupted DiLiGenT Cat dataset. Mean angular errors (in degrees) for each reconstruction are shown in parentheses. See Table \ref{tab:full_DiLiGenT} for the corresponding median reconstruction errors. PDLNV achieves smaller error on the body of the cat and the edge of the back and head compared to the existing methods.}}
\label{fig:cat_errors}
\end{figure*}

As these results illustrate, incorporating dictionary learning into the PLS model can significantly improve reconstruction accuracy---on both corrupted and uncorrupted datasets. Note that PDLNV is one of many possible applications of dictionary learning to photometric stereo. In principle, it could be fruitful to apply dictionary learning-based regularization to each of the other methods considered in our experiments, which may yield improvements similar to those observed here with respect to PLS. While such experiments are beyond the scope of this work, the results presented in this section demonstrate the viability of adaptive dictionary learning as a regularizer for photometric stereo problems.
}

\subsection{Evaluation on Uncorrupted DiLiGenT Dataset}

\revised{We next compare the performance of PDLNV to other existing state-of-the-art approaches on} the DiLiGenT dataset \cite{shi2016}. For each object, we use all 96 images present in the dataset and do not add any additional corruptions to the images.

\begin{figure*}[t!]
\centering
\begin{minipage}{0.45\textwidth}
\includegraphics[width=\columnwidth]{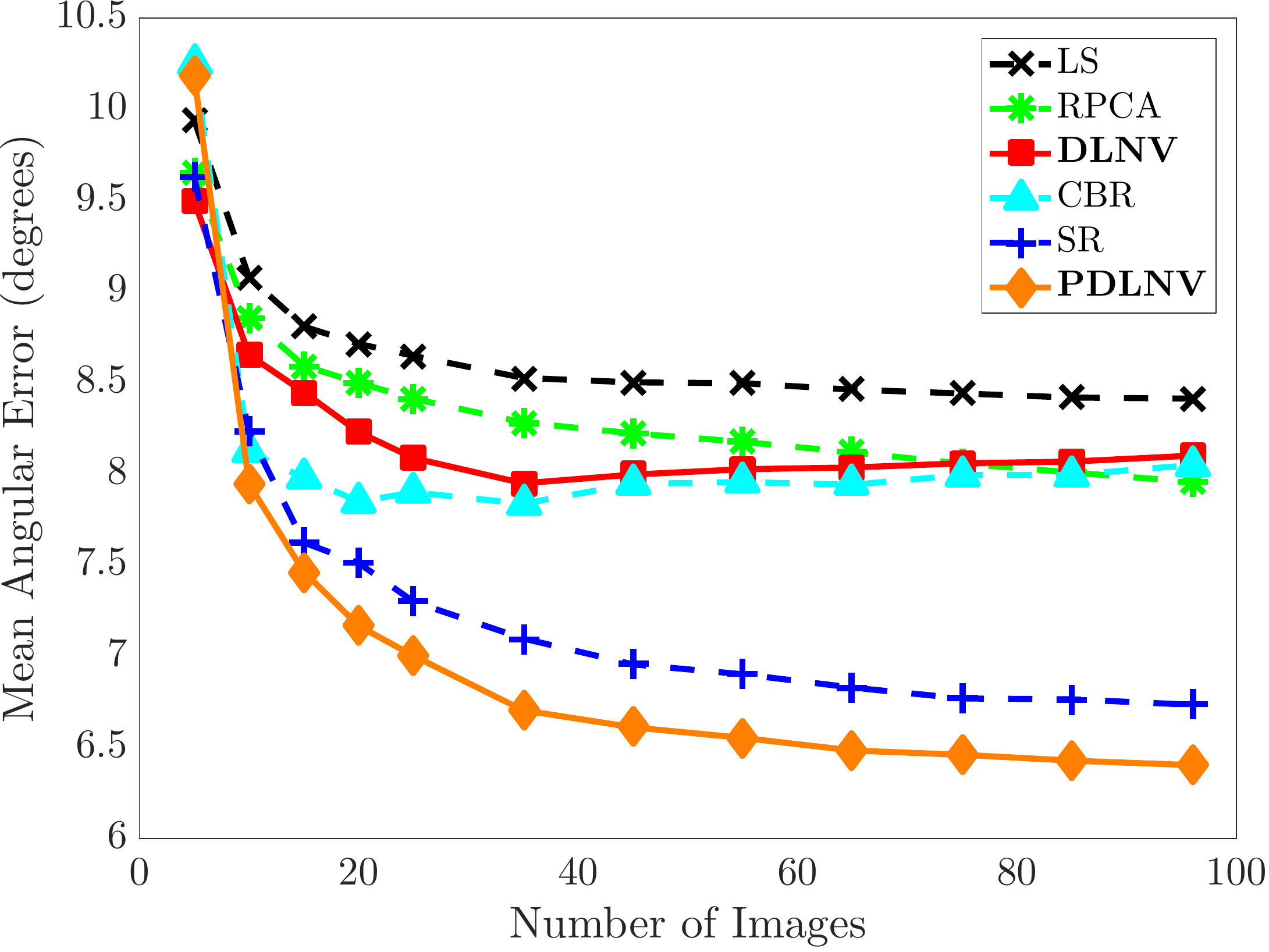}
\caption{Mean angular errors of the estimated normal vectors for the DiLiGenT Cat dataset as a function of number of images used during reconstruction.}
\label{fig:sweep_nim_na_inf_0_Dcats}
\end{minipage}
\hspace{4mm}
\begin{minipage}{0.45\textwidth}
\includegraphics[width=\columnwidth]{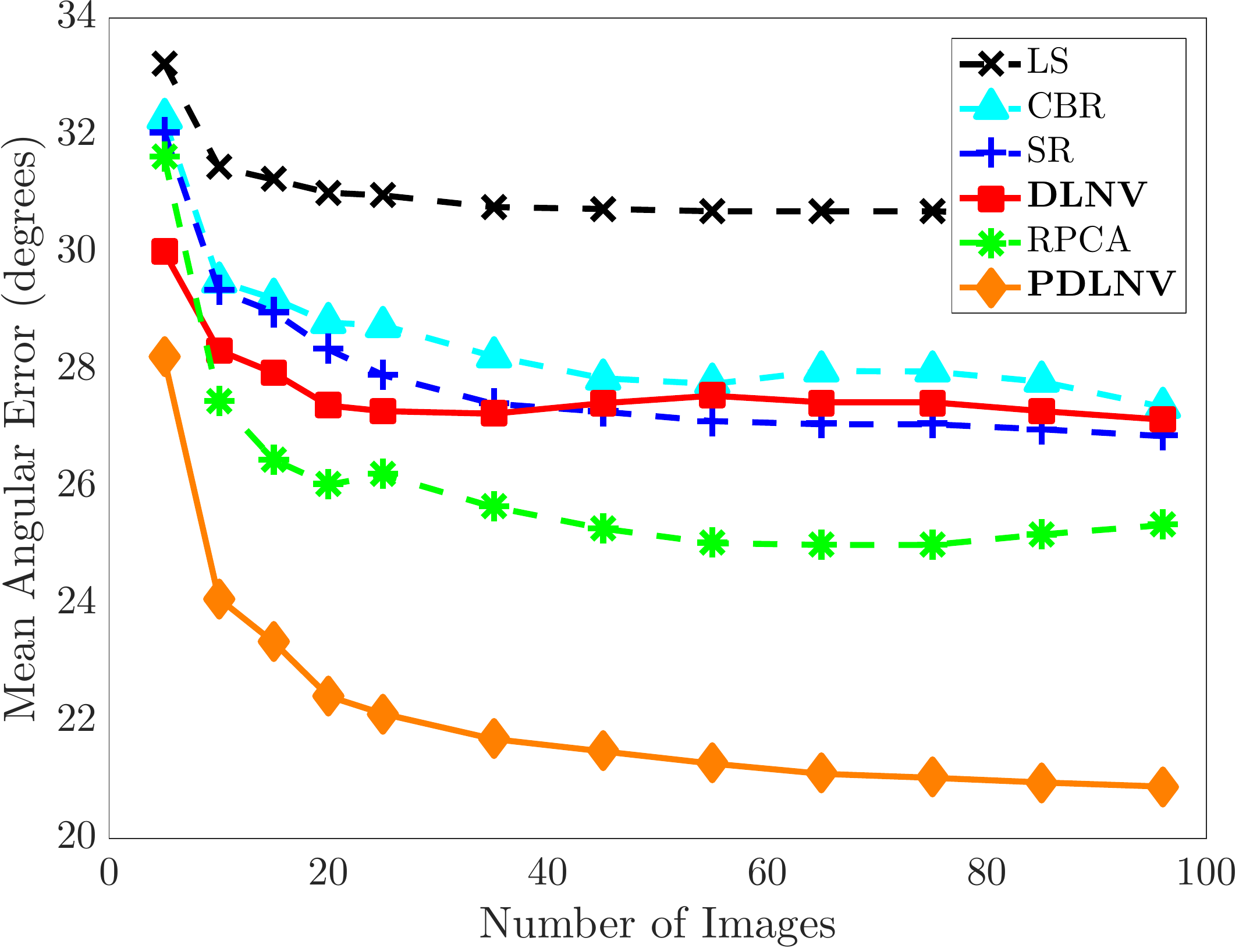}
\caption{Mean angular errors of the estimated normal vectors for the DiLiGenT Harvest dataset as a function of number of images used during reconstruction.}
\label{fig:sweep_nim_na_inf_0_Dharvests}
\end{minipage}
\end{figure*}

Table~\ref{tab:full_DiLiGenT} presents the mean and median angular errors of the reconstructed normal vectors for each method on each dataset. \revised{With respect to mean angular error, PDLNV outperforms all existing approaches on 5 of the 10 objects}. In cases where our methods do not outperform existing approaches, with the exception of the Reading dataset, we achieve comparable angular errors to the best performer. As we will demonstrate, the primary strength of our proposed methods is constructing normal vectors from images that are less pristine than the DiLiGenT datasets. However, Table~\ref{tab:full_DiLiGenT} shows that PDLNV is still able to perform better than or comparable to methods specifically designed to operate on large, clean datasets. \revised{Figures \ref{fig:cat_normals} and \ref{fig:cat_errors} illustrate the normal vector reconstructions and corresponding error maps for the full DiLiGenT Cat dataset. While it is difficult to discern a large difference between the normal maps, the error maps clearly illustrate the improvement PDLNV achieves with respect to the existing methods. Certain regions of the cat, such as the lower hand, pose difficulties for each approach. However, as demonstrated by Figure \ref{fig:cat_errors}, PDLNV is able to accurately reconstruct the body of the cat while also accurately reconstructing the normal vectors around the edges, a result that no other method achieves. We observe that this finding is generally true: in the case of uncorrupted datasets, dictionary learning is able to improve the performance on relatively smooth regions of the surface and regions affected by wide specular lobes. At the same time, it typically does not compromise the fine details at the edges or at sharp corners. Note that, to better illustrate the differences between the various approaches in this example, we have applied error clipping to Figure \ref{fig:cat_errors}.}

\revised{\begin{table}[t!]
\begin{center}
\begin{tabular}{|c|c|c|c|c|c|}
\cline{1-6}
& PDLNV & DLNV & CBR & SR & RPCA \\ 
\cline{1-6}
Ball & 521.4 & 156.5 & \textbf{15.8} & 872.2 & 1076.3 \\
\cline{1-6}
Cat & 600.4 & 146.7 & \textbf{34.6} & 2009.6 & 2506.2 \\
\cline{1-6}
Pot1 & 702.6 & 149.9 & \textbf{45.1} & 2508.6 & 1421.6 \\
\cline{1-6}
Bear & 640.3 & 225.5 & \textbf{37.1} & 3335.0 & 4249.1 \\
\cline{1-6}
Pot2 & 624.4 & 154.8 & \textbf{31.1} & 1639.8 & 5328.6 \\
\cline{1-6}
Buddha & 654.0 & 147.0 & \textbf{34.9} & 2007.1 & 4878.1 \\
\cline{1-6}
Goblet & 602.2 & 156.8 & \textbf{34.3} & 1283.4 & 6524.6 \\
\cline{1-6}
Reading & 589.2 & 149.7 & \textbf{22.5} & 1338.5 & 2707.2 \\
\cline{1-6}
Cow & 609.8 & 164.9 & \textbf{27.7} & 1310.5 & 6179.8 \\
\cline{1-6}
Harvest & 609.9 & 156.6 & \textbf{100.9} & 2473.0 & 5449.9 \\
\cline{1-6}
\end{tabular}
\end{center}
\caption{\revised{Runtimes (in seconds) of each method on the full DiLiGenT datasets.}}
\label{tab:runtimes}
\end{table}}

\revised{Table~\ref{tab:runtimes} displays the runtimes for each of the methods on the full DiLiGenT dataset. As this table illustrates, our methods achieve competitive runtimes and are significantly faster than several state-of-the-art approaches. We also note that, as our methods are iterative, one is free to decrease runtime at the expense of a small decrease in accuracy by terminating the algorithm after fewer iterations.}

In practice, it may be infeasible to collect 96 images of an object under varying lighting conditions. As such, it is important to develop methods that can accurately estimate normal vectors from smaller datasets. Figures~\ref{fig:sweep_nim_na_inf_0_Dcats} and \ref{fig:sweep_nim_na_inf_0_Dharvests} illustrate the angular errors of the normal vectors estimated by each method on the uncorrupted DiLiGenT Cat and Harvest datasets as a function of the number of images used. In these experiments, we randomly selected images from among the original 96 images and averaged the results across 10 trials. From both figures, it is clear that PDLNV outperforms all other methods for most dataset sizes.

\begin{figure}[t!]
\centering
\includegraphics[width=0.45\textwidth]{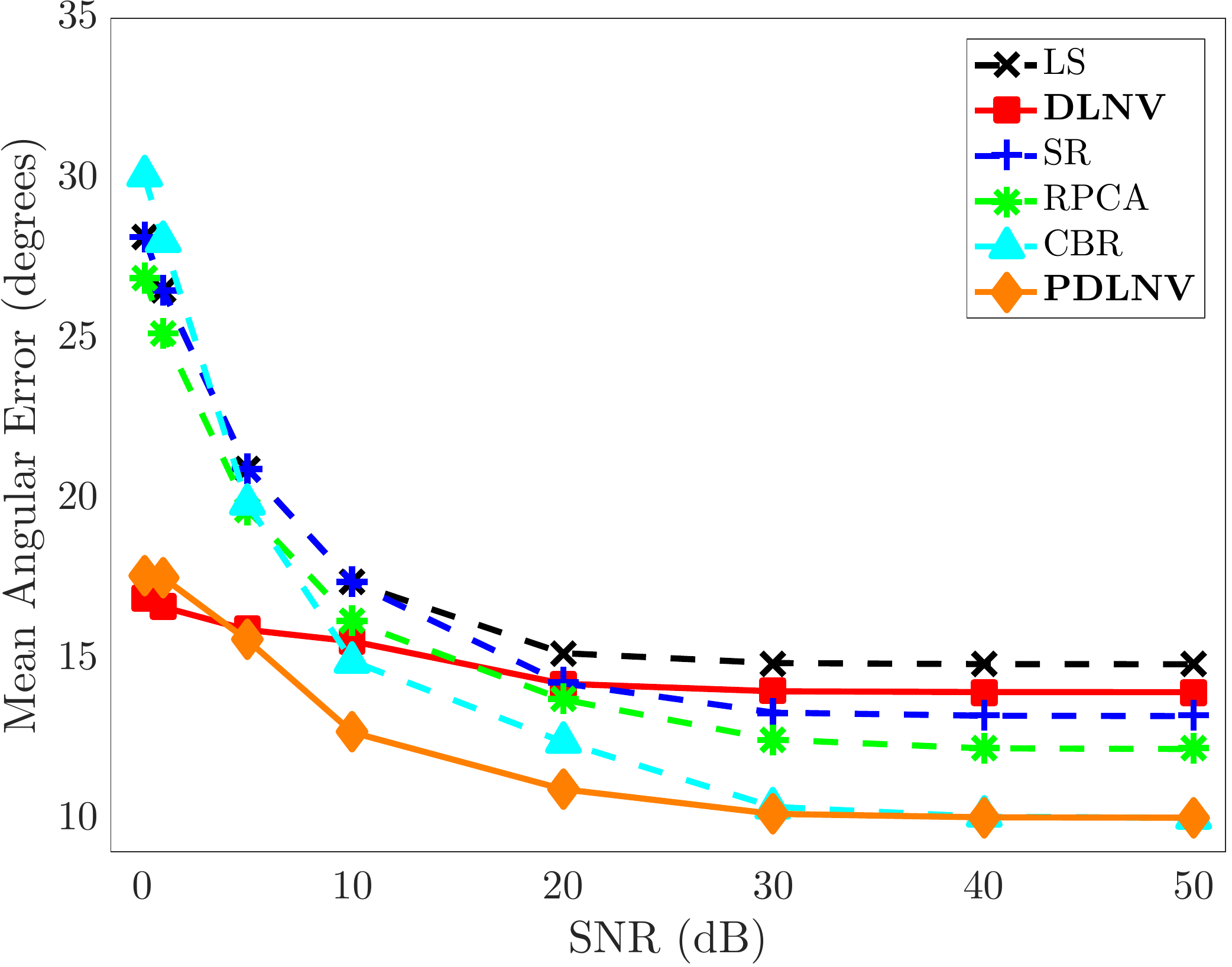}
\caption{\revised{Mean angular errors of the estimated normal vectors for the DiLiGenT Pot2 dataset with 20 images versus SNR}.}
\label{fig:sweep_snr_20_na_0_Dpot2s_all}
\end{figure}

\begin{figure*}[t!]
\centering
\begin{subfigure}[b]{0.14\textwidth}
  \includegraphics[width=\textwidth]{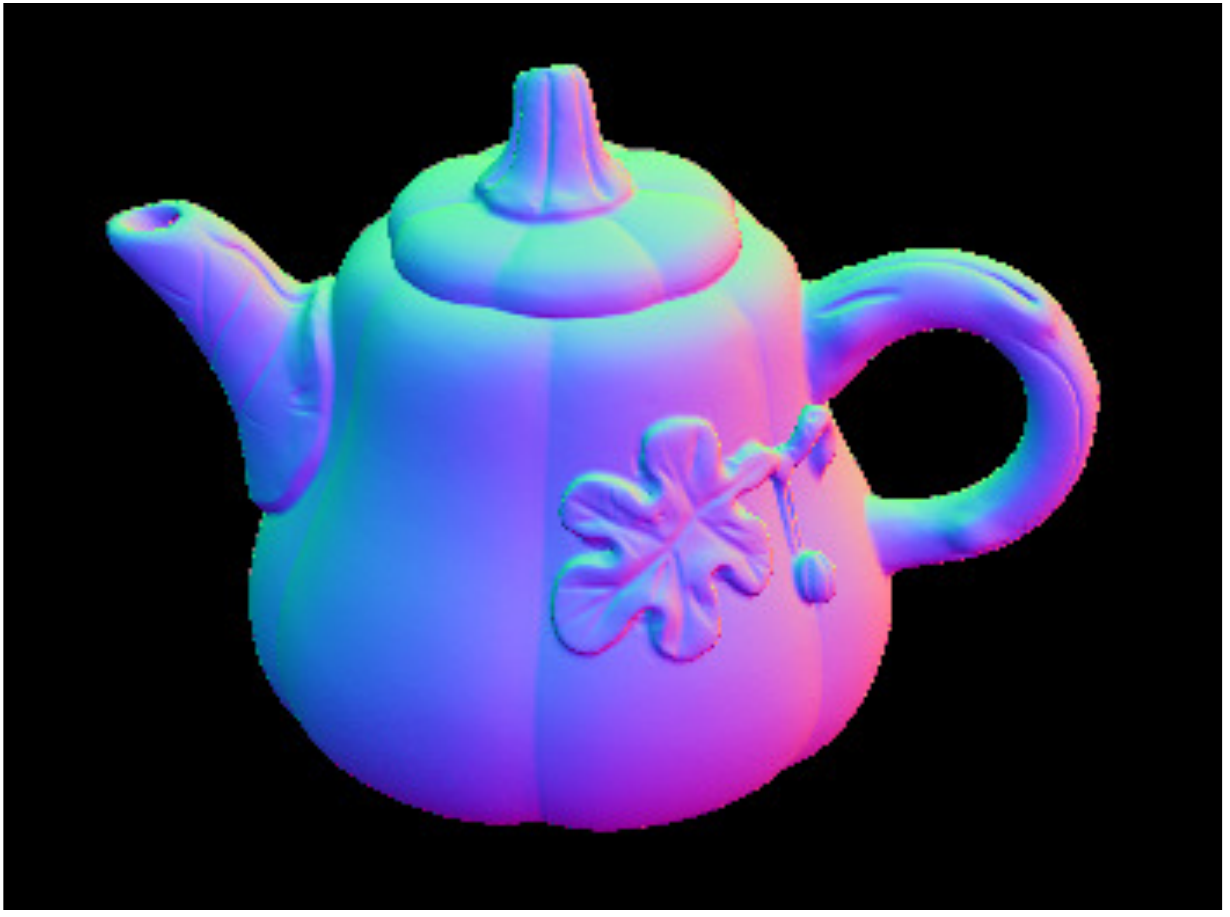}
  \caption{Truth}
\end{subfigure}
\hspace{-2mm}
\begin{subfigure}[b]{0.14\textwidth}
  \includegraphics[width=\textwidth]{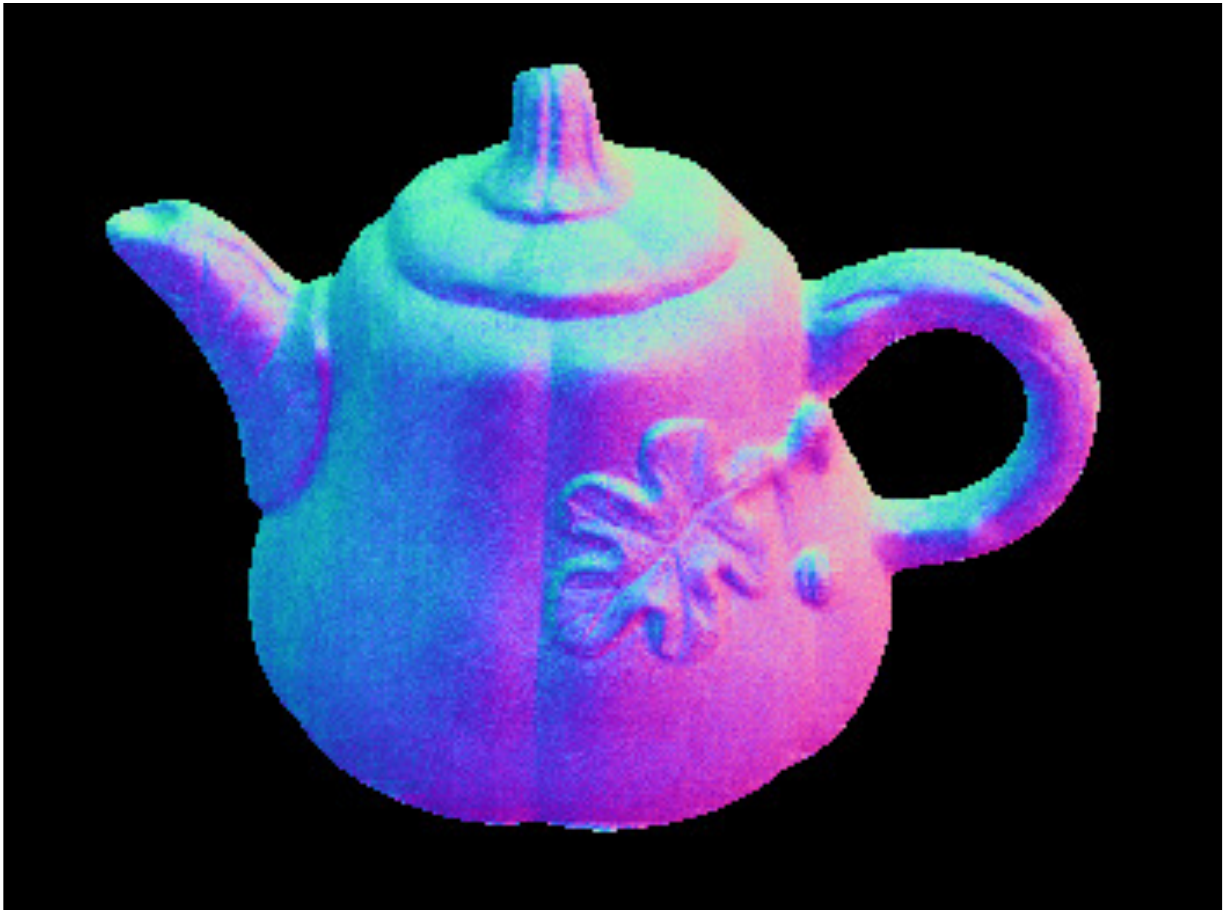}
  \caption{\textbf{PDLNV}}
\end{subfigure}
\hspace{-2mm}
\begin{subfigure}[b]{0.14\textwidth}
  \includegraphics[width=\textwidth]{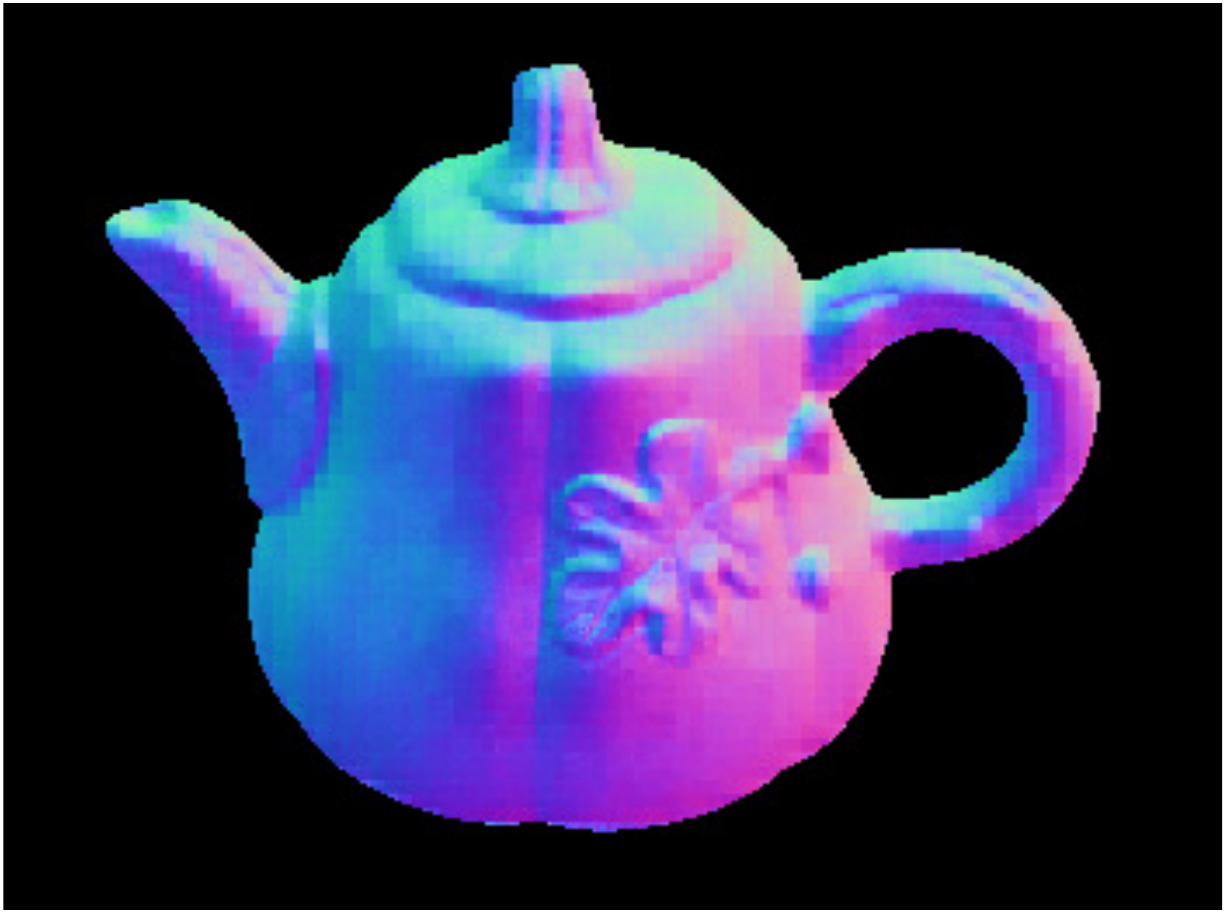}
  \caption{\textbf{DLNV}}
\end{subfigure}
\hspace{-2mm}
\begin{subfigure}[b]{0.14\textwidth}
  \includegraphics[width=\textwidth]{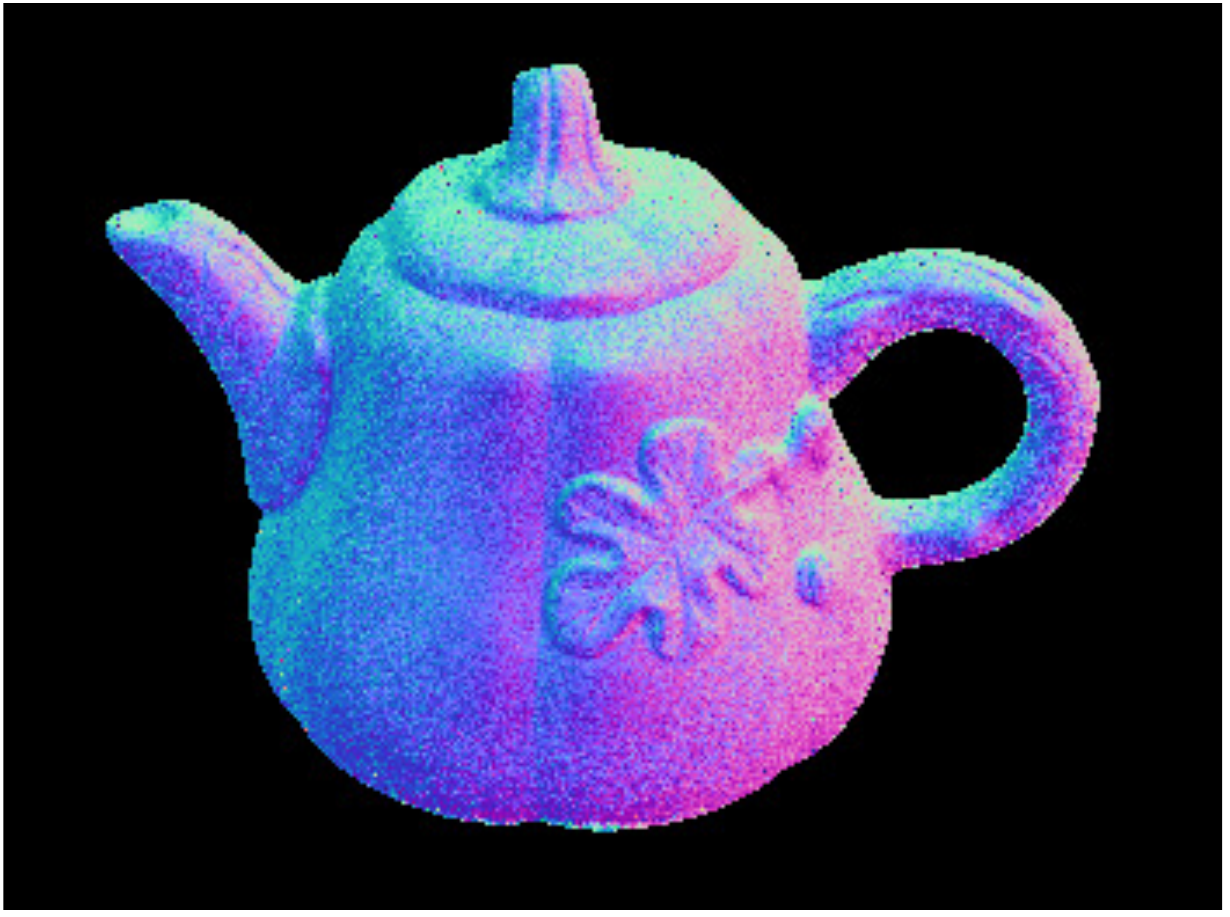}
  \caption{CBR}
\end{subfigure}
\hspace{-2mm}
\begin{subfigure}[b]{0.14\textwidth}
  \includegraphics[width=\textwidth]{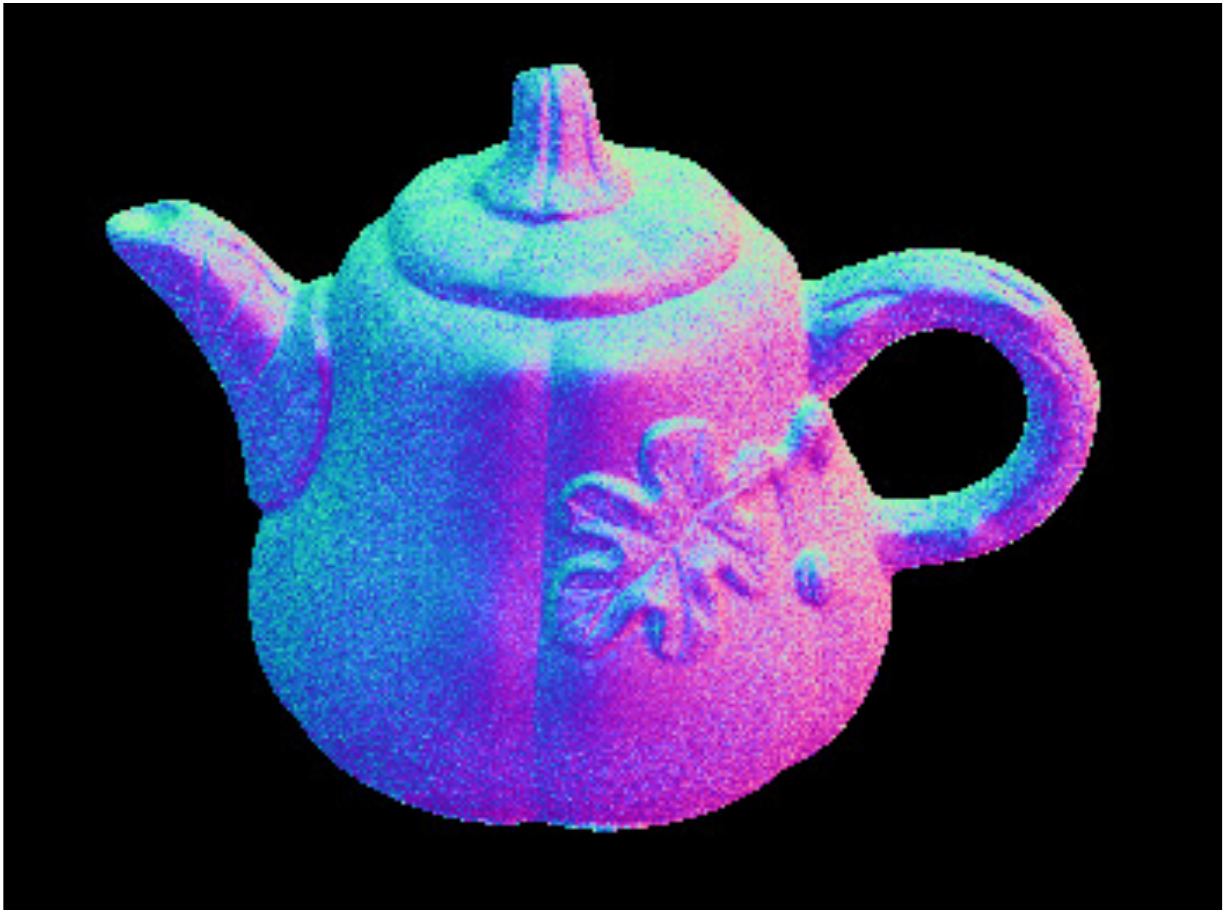}
  \caption{SR}
\end{subfigure}
\hspace{-2mm}
\begin{subfigure}[b]{0.14\textwidth}
  \includegraphics[width=\textwidth]{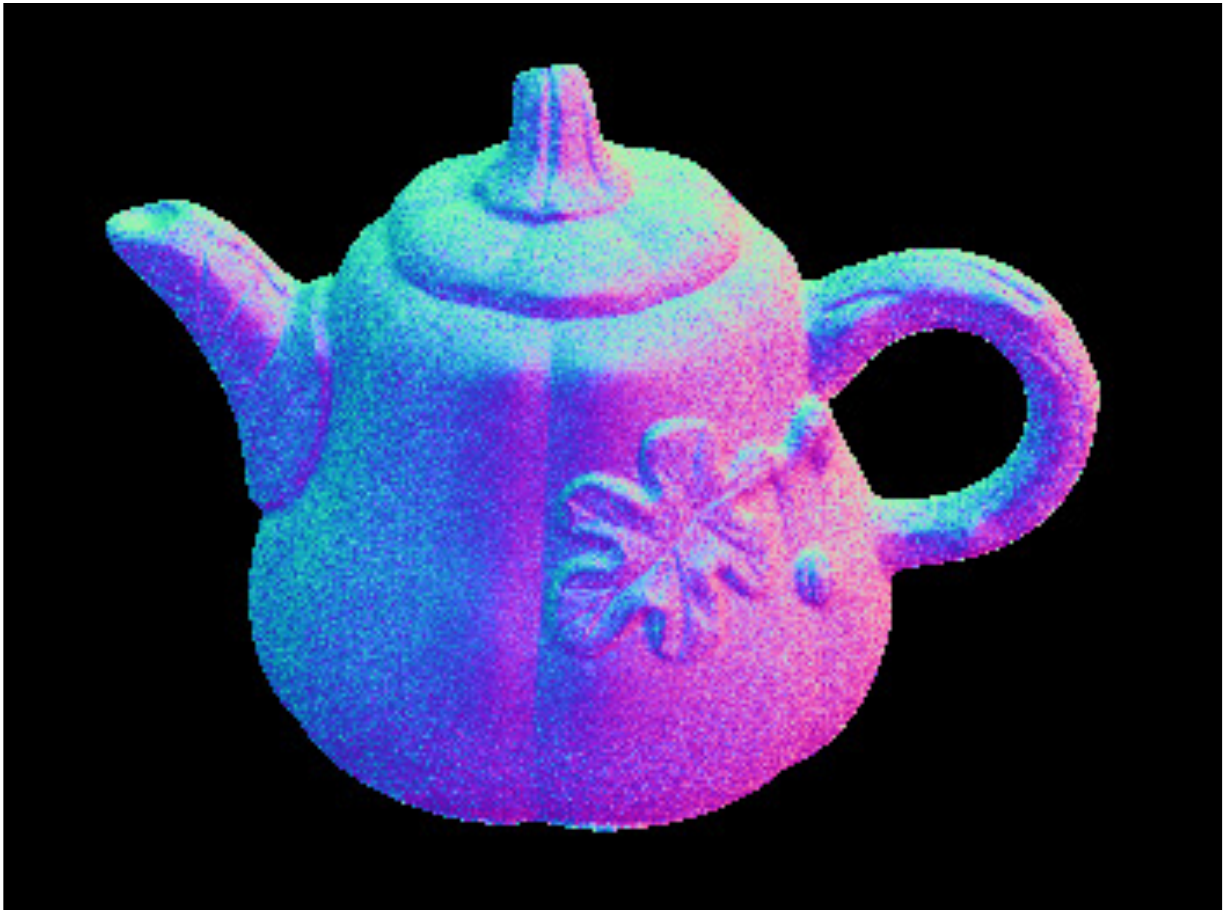}
  \caption{RPCA}
\end{subfigure}
\hspace{-2mm}
\begin{subfigure}[b]{0.14\textwidth}
  \includegraphics[width=\textwidth]{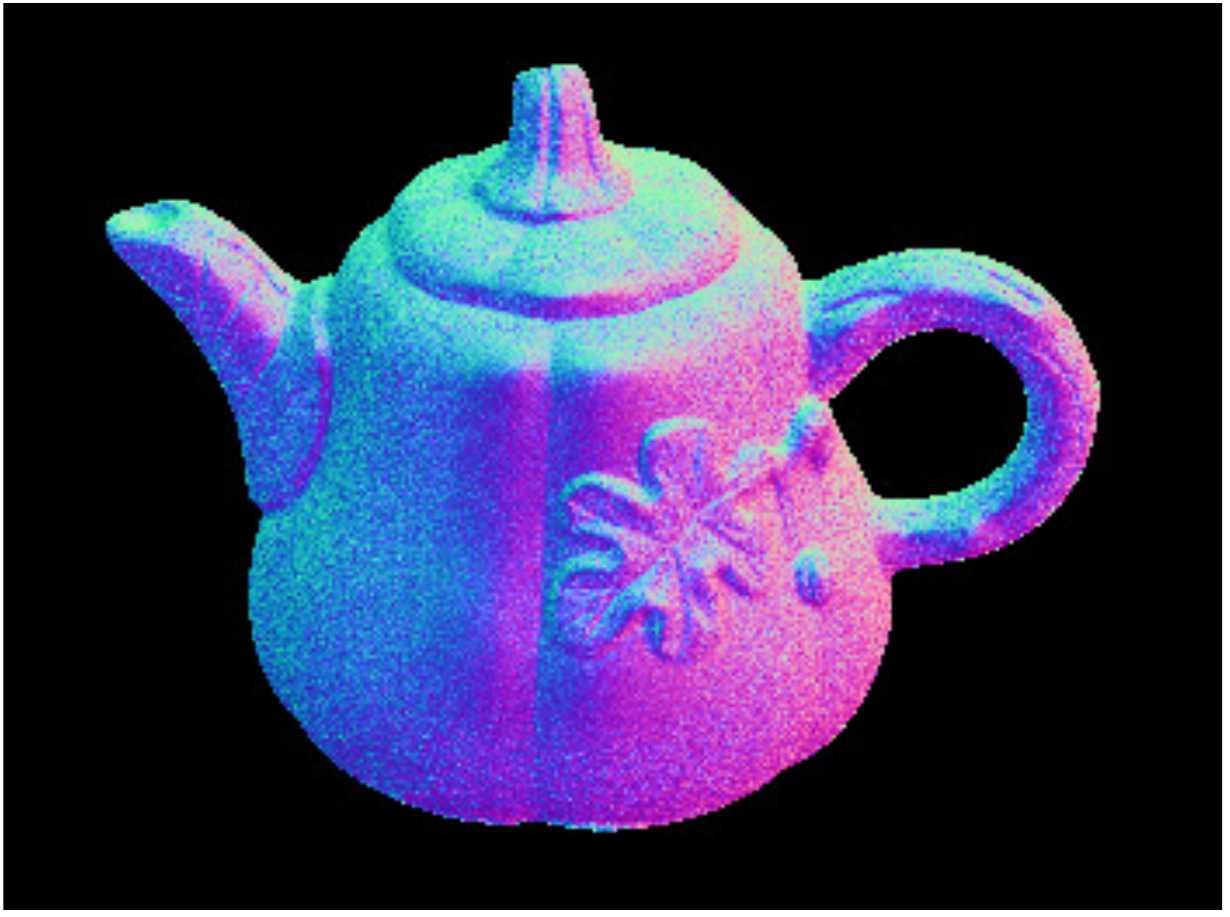}
  \caption{LS}
\end{subfigure}
\caption{Normal vector reconstructions for the DiLiGenT Pot2 dataset with 20 images and \revised{Poisson noise added with resulting SNR 5 dB.}}
\label{fig:pot2_normals}
\end{figure*}

\begin{figure*}[t!]
\centering
\begin{subfigure}[b]{0.14\textwidth}
  \includegraphics[width=\textwidth]{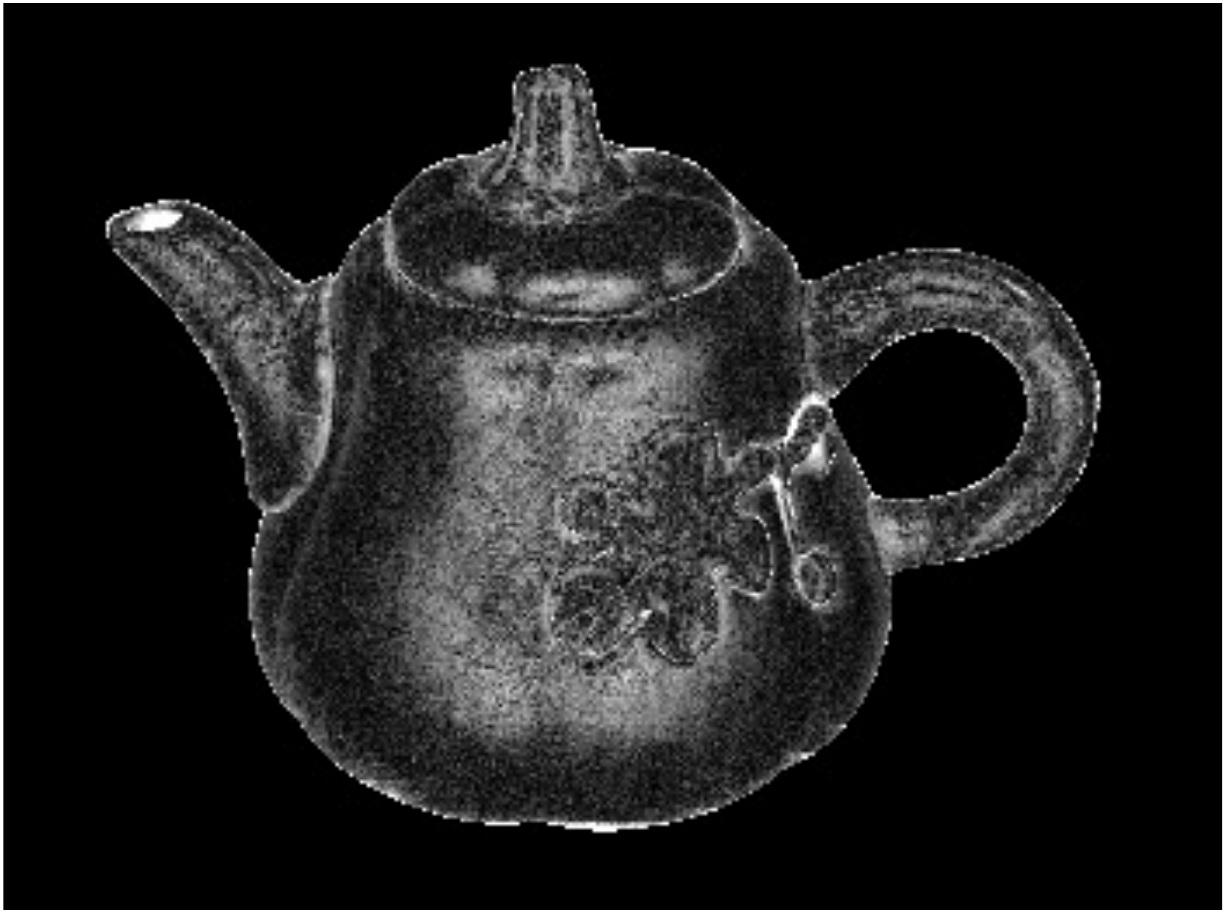}
  \caption{\textbf{PDLNV} (15.72)}
\end{subfigure}
\hspace{-2mm}
\begin{subfigure}[b]{0.14\textwidth}
  \includegraphics[width=\textwidth]{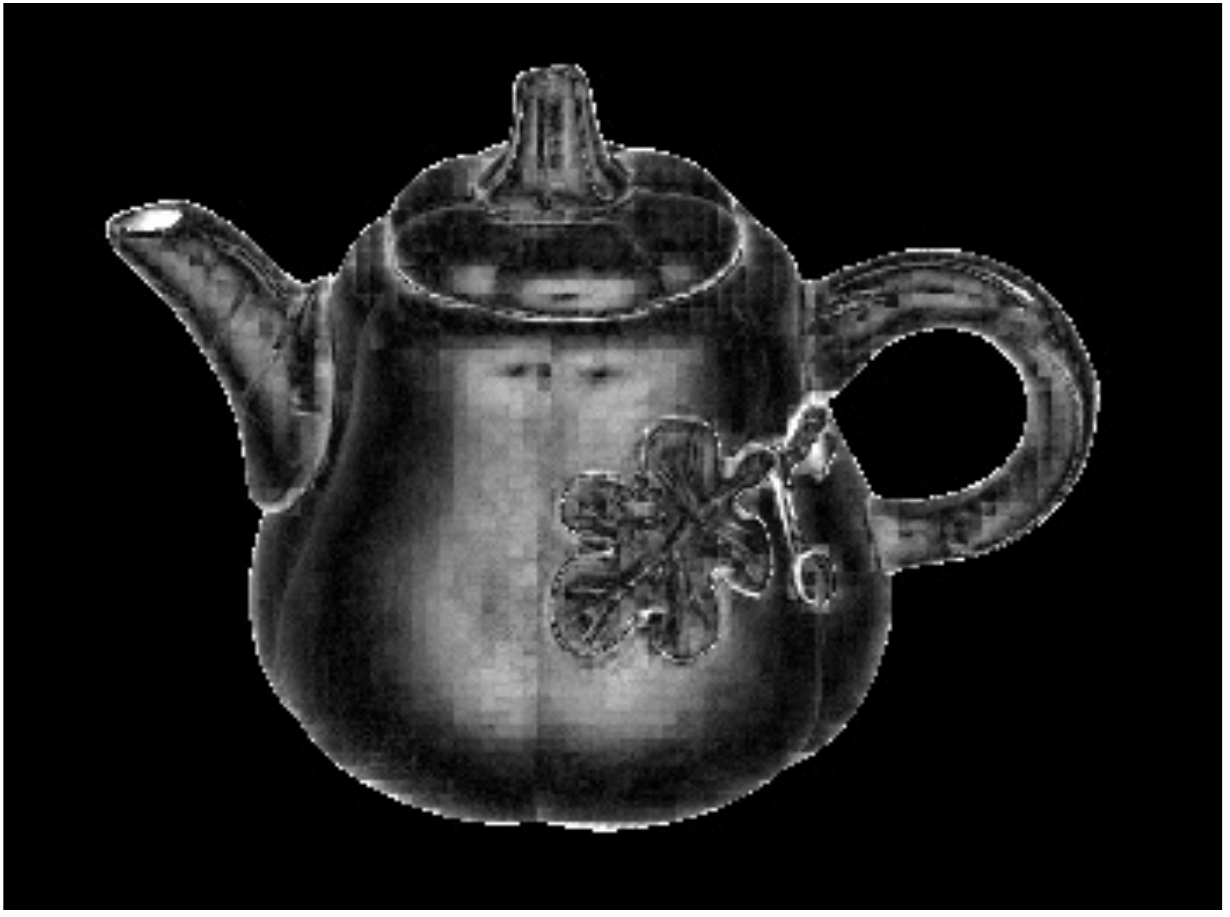}
  \caption{\textbf{DLNV} (16.10)}
\end{subfigure}
\hspace{-2mm}
\begin{subfigure}[b]{0.14\textwidth}
  \includegraphics[width=\textwidth]{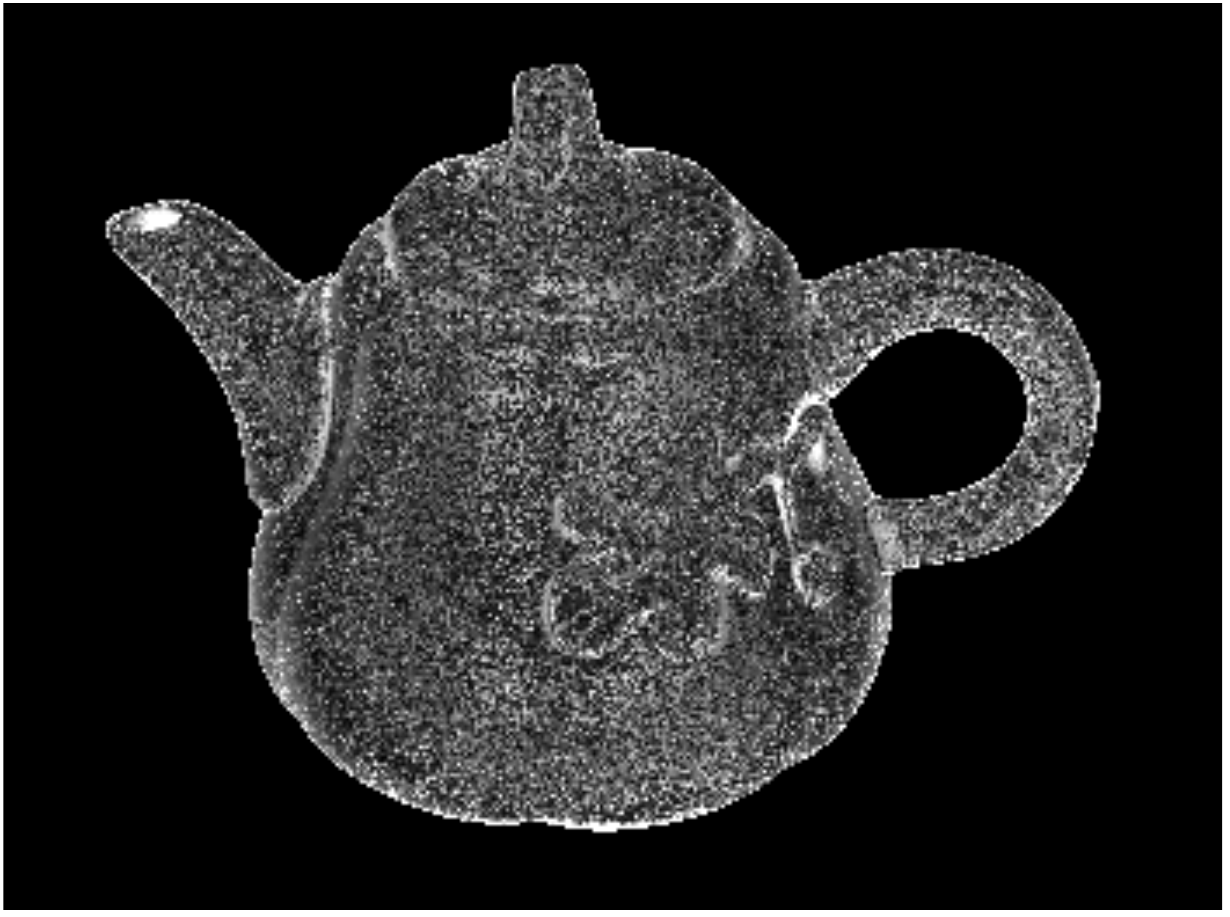}
  \caption{CBR (19.70)}
\end{subfigure}
\hspace{-2mm}
\begin{subfigure}[b]{0.14\textwidth}
  \includegraphics[width=\textwidth]{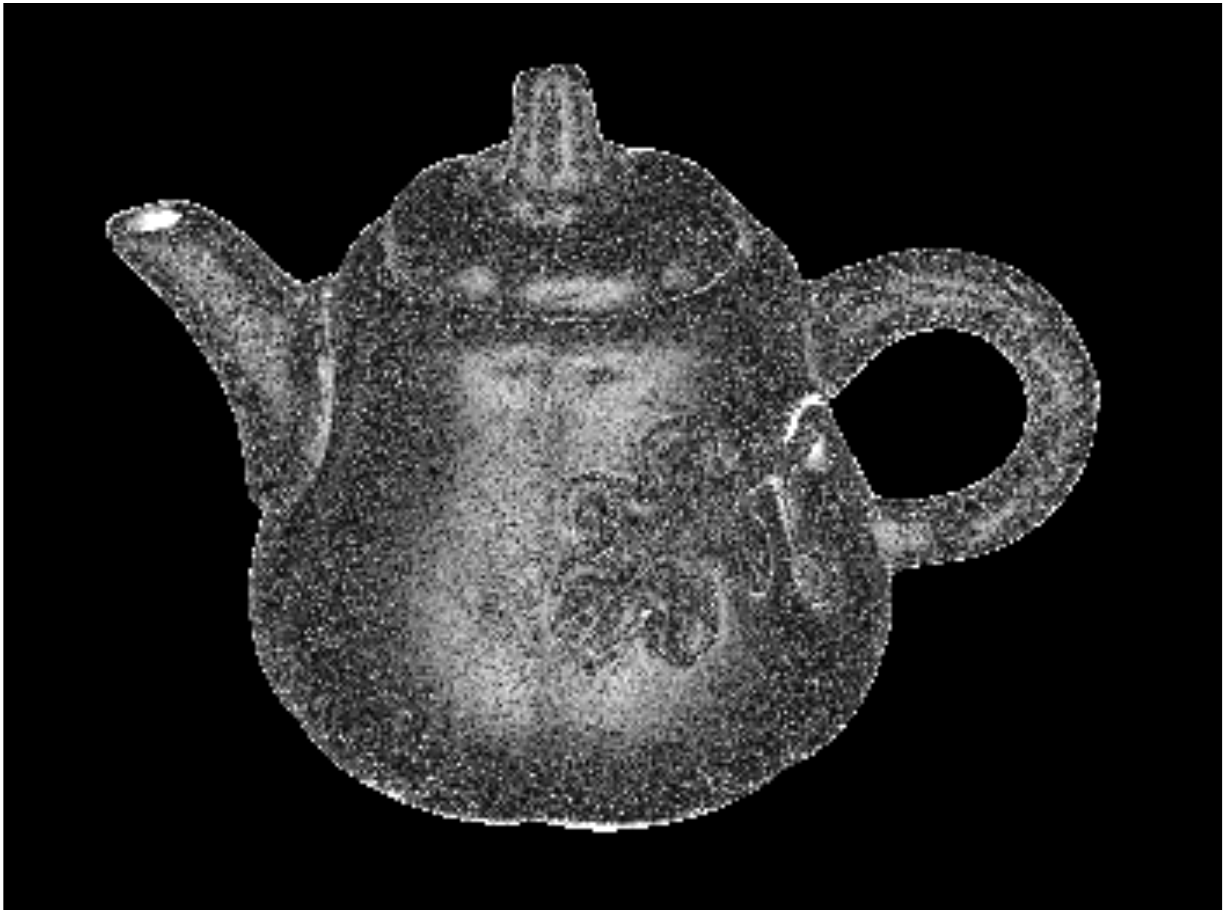}
  \caption{SR (21.20)}
\end{subfigure}
\hspace{-2mm}
\begin{subfigure}[b]{0.14\textwidth}
  \includegraphics[width=\textwidth]{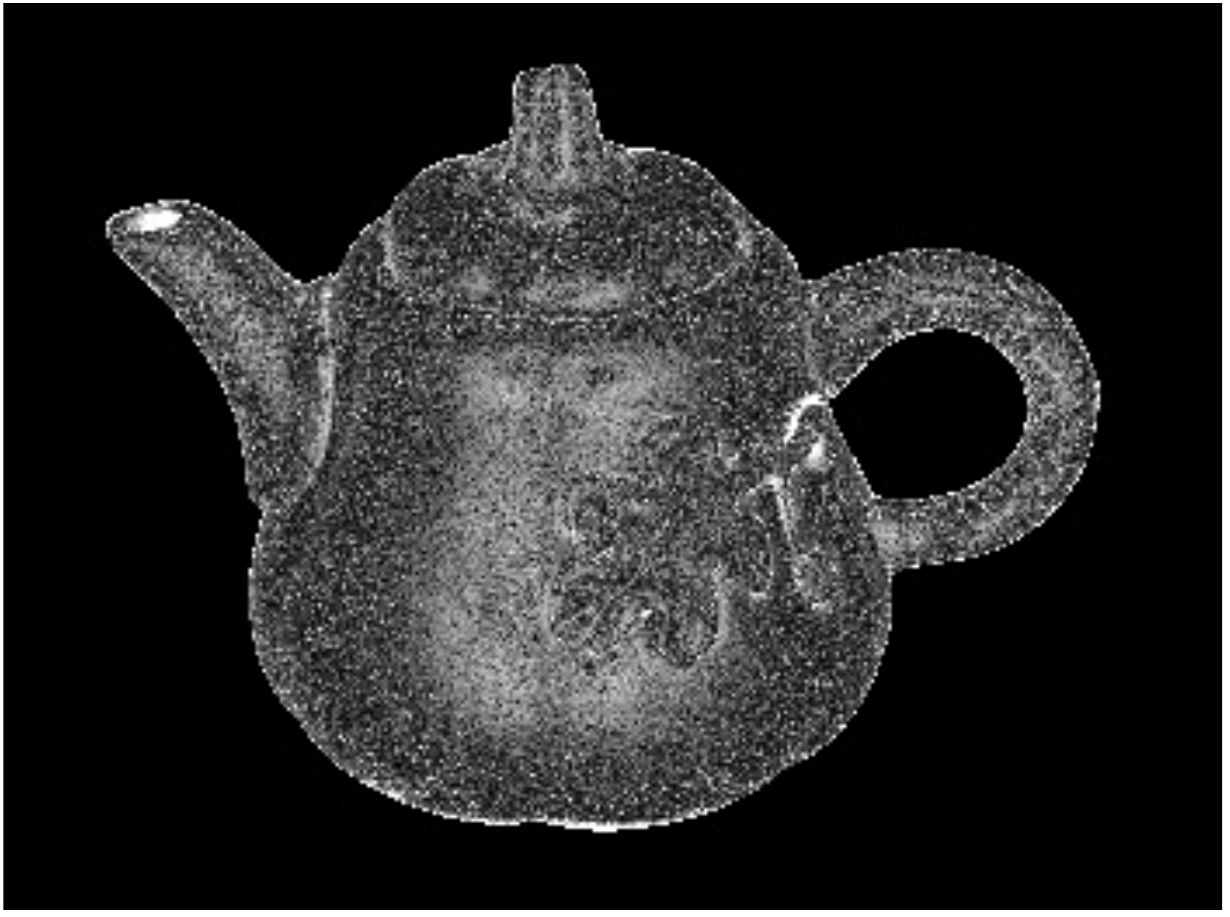}
  \caption{RPCA (19.95)}
\end{subfigure}
\hspace{-2mm}
\begin{subfigure}[b]{0.14\textwidth}
  \includegraphics[width=\textwidth]{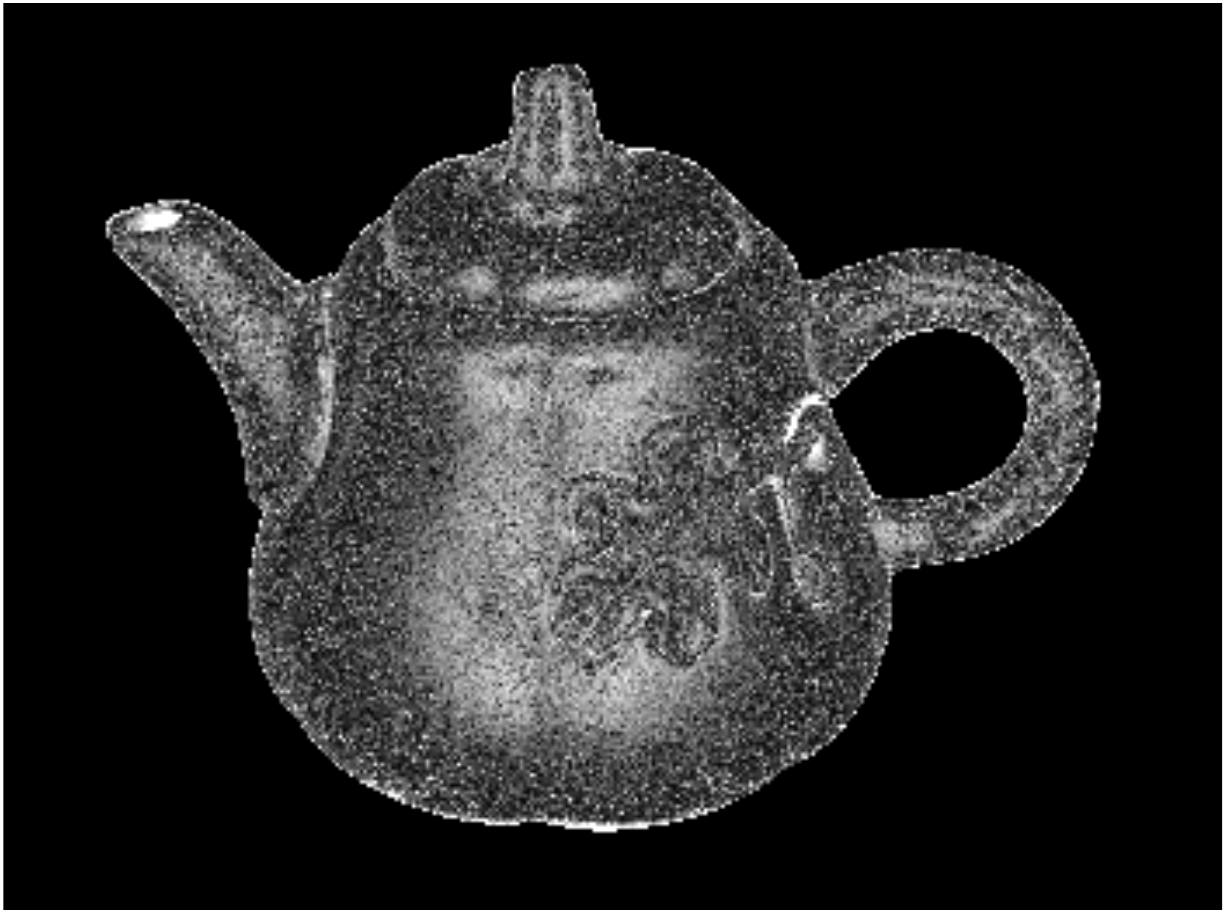}
  \caption{LS (21.20)}
\end{subfigure}
\hspace{-2mm}
\begin{subfigure}[b]{0.035\textwidth}
\includegraphics[width=\textwidth]{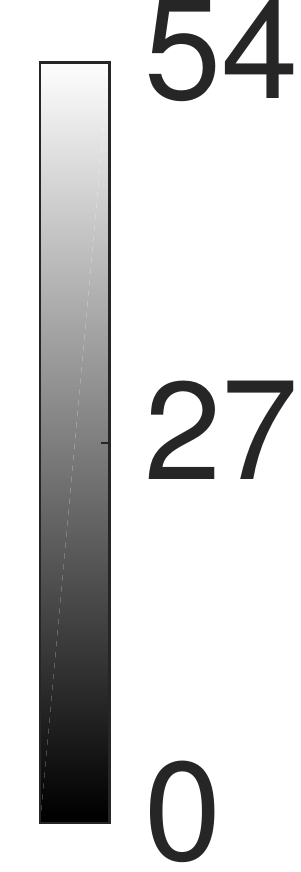}
\caption*{}
\end{subfigure}
\caption{Normal vector error maps for the DiLiGenT Pot2 dataset with 20 images and \revised{Poisson noise added with resulting SNR 5dB. Mean angular errors (in degrees) for each reconstruction are shown in parentheses.}}
\label{fig:pot2_errors}
\end{figure*}

\begin{figure}[t!]
\centering
\includegraphics[width=0.45\textwidth]{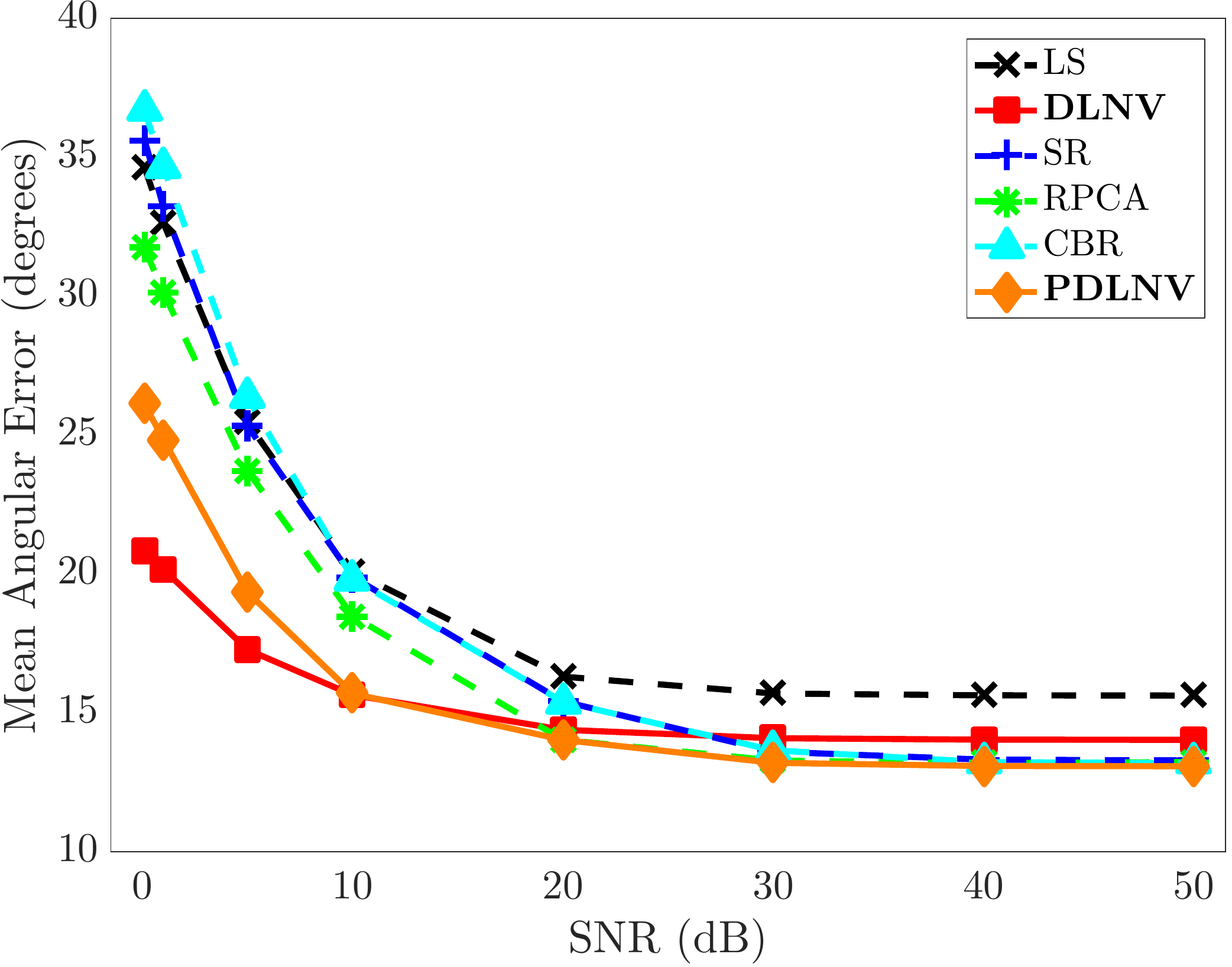}
\caption{\revised{Mean angular error averaged across all DiLiGenT objects with 20 images versus SNR.}}
\label{fig:sweep_snr_20_na_0_aggregate}
\end{figure}

\subsection{Evaluation on Corrupted DiLiGenT Dataset}

We next compare the performance of our proposed methods to existing methods on images corrupted with Poisson noise. Specifically, we subsample the DiLiGenT Pot2 dataset to 20 images and then corrupt these images with Poisson noise of a given signal-to-noise-ratio (SNR).

Figure~\ref{fig:sweep_snr_20_na_0_Dpot2s_all} plots the angular errors of the estimated normal vectors for each algorithm as a function of SNR. It is clear that our proposed dictionary learning-based approaches are significantly more robust to high levels of non-sparse corruptions than existing methods. In particular, for SNR values below 10 dB, our methods outperform the existing methods by up to 10 degrees. Furthermore, the angular errors produced by our dictionary learning-based methods vary significantly less than existing approaches, indicating that the normal vector reconstructions are much more stable and robust to these corruptions.

Figures~\ref{fig:pot2_normals} and \ref{fig:pot2_errors} show the normal vector reconstructions and the corresponding error maps produced by each method on the Pot2 dataset with 20 images at a noise level of 5 dB. As these figures illustrate, the dictionary learning-based reconstructions are significantly more accurate and robust to noise than the existing methods.

\revised{Figure \ref{fig:sweep_snr_20_na_0_aggregate} plots the reconstruction error averaged across all objects from the DiLiGenT dataset (subsampled to 20 images each) with Poisson noise added. That is, each point on the curves corresponds to the average error obtained by a given method and SNR averaged across all objects. As this figure illustrates, over the entire DiLiGenT dataset, PDLNV and DLNV once again significantly outperform existing approaches in the low SNR regime.}

\subsection{Evaluation on non-DiLiGenT Datasets}\label{sec:ex_non_diligent}

\begin{figure}[t!]
\centering
\includegraphics[width=0.45\textwidth]{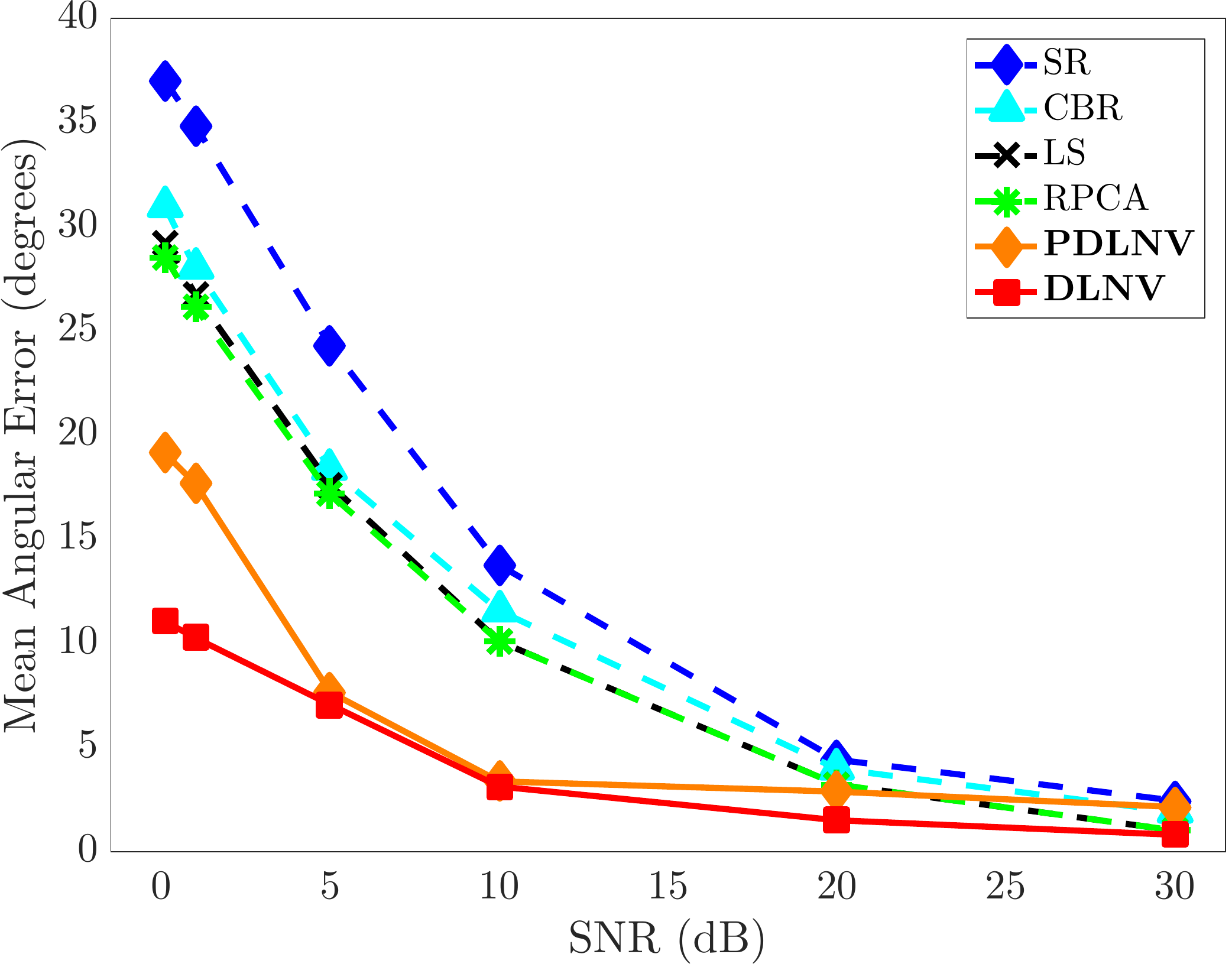}
\caption{Mean angular errors of estimated normal vectors for the Hippo dataset with 20 images versus SNR.}
\label{fig:sweep_snr_20_na_0_hippo_all}
\end{figure}

\begin{figure*}[t!]
\centering
\begin{subfigure}[b]{0.14\textwidth}
  \includegraphics[width=\textwidth]{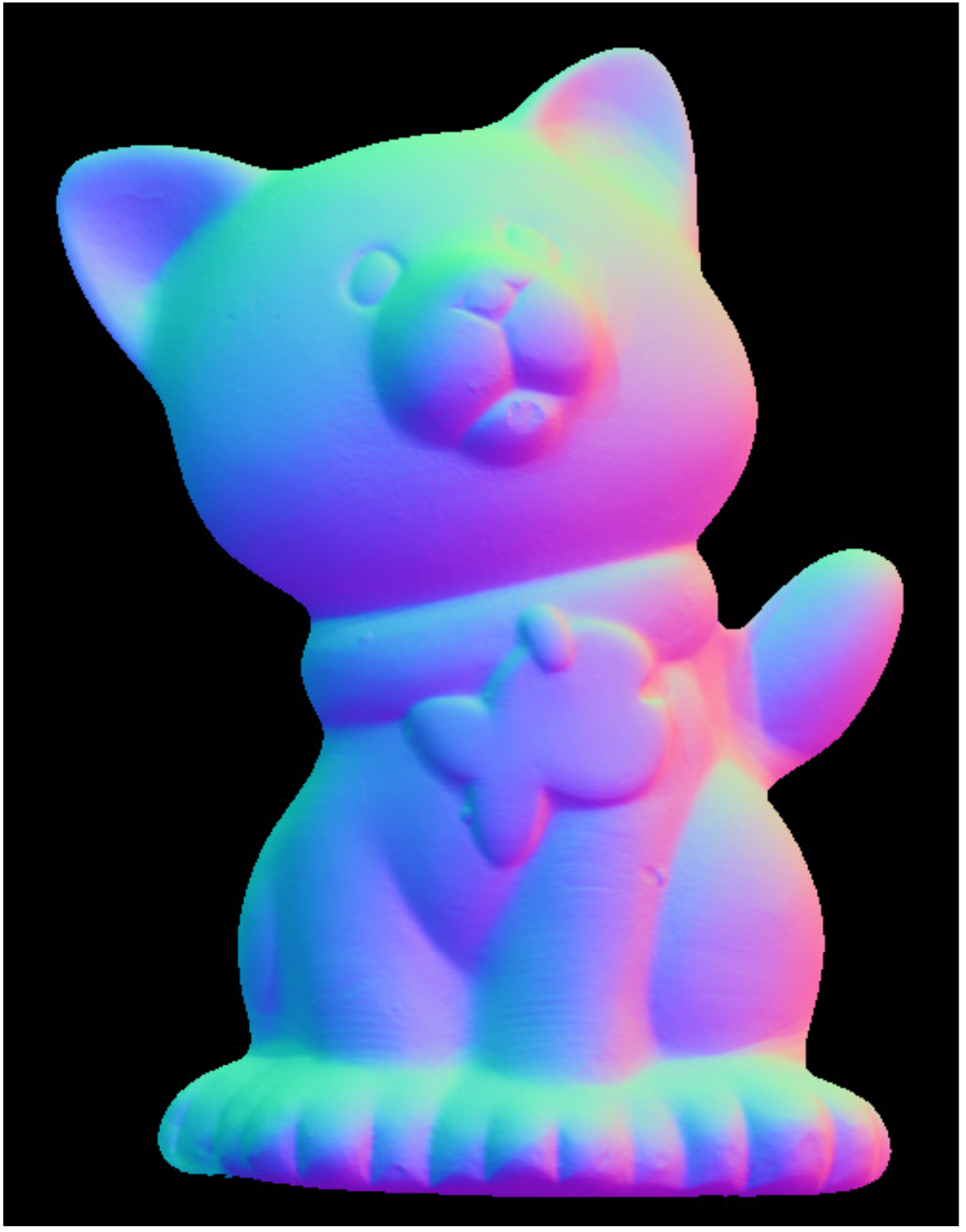}
  \caption{Reference}
\end{subfigure}
\hspace{-2mm} 
\begin{subfigure}[b]{0.14\textwidth}
  \includegraphics[width=\textwidth]{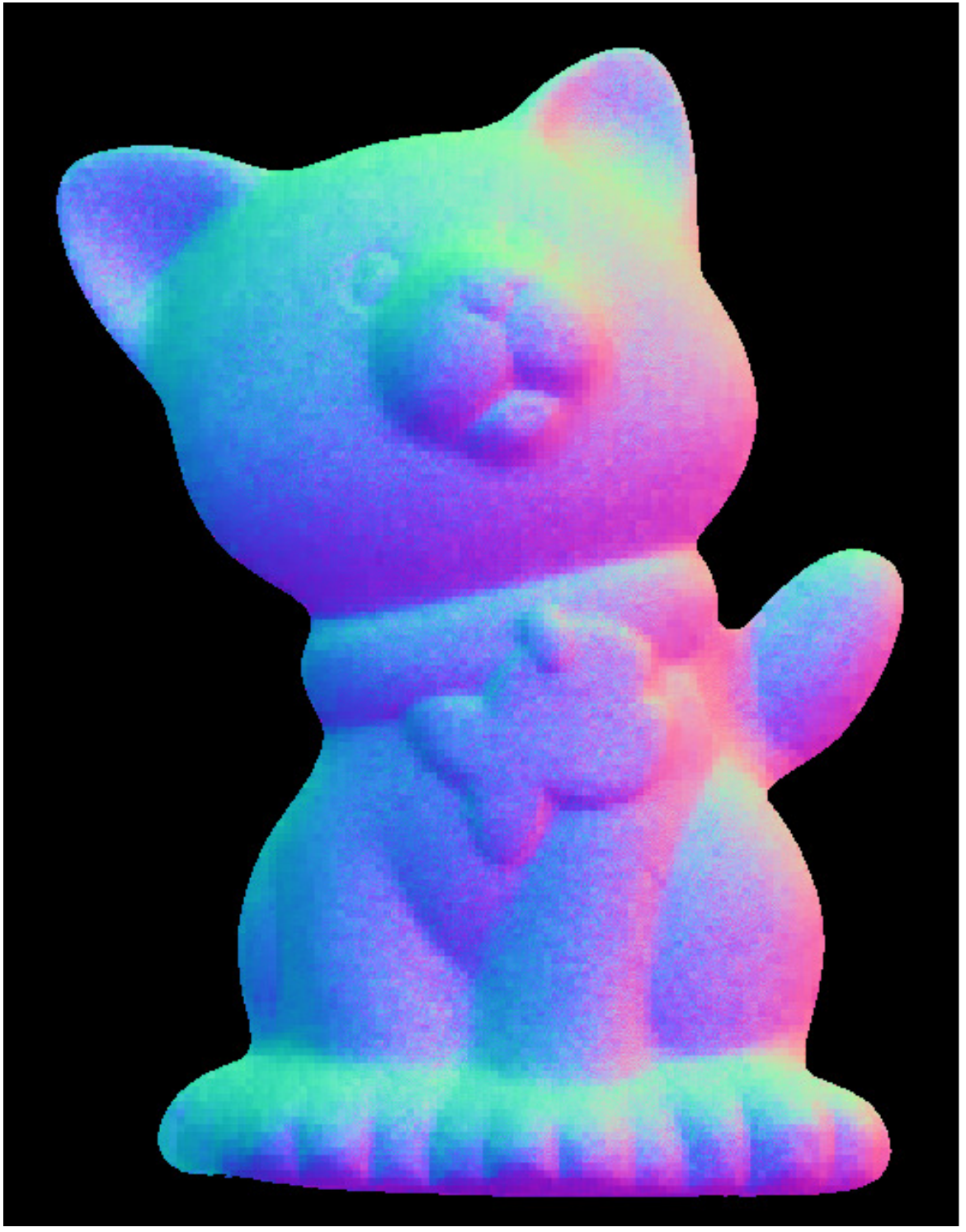}
  \caption{\textbf{PDLNV}}
\end{subfigure}
\hspace{-2mm} 
\begin{subfigure}[b]{0.14\textwidth}
  \includegraphics[width=\textwidth]{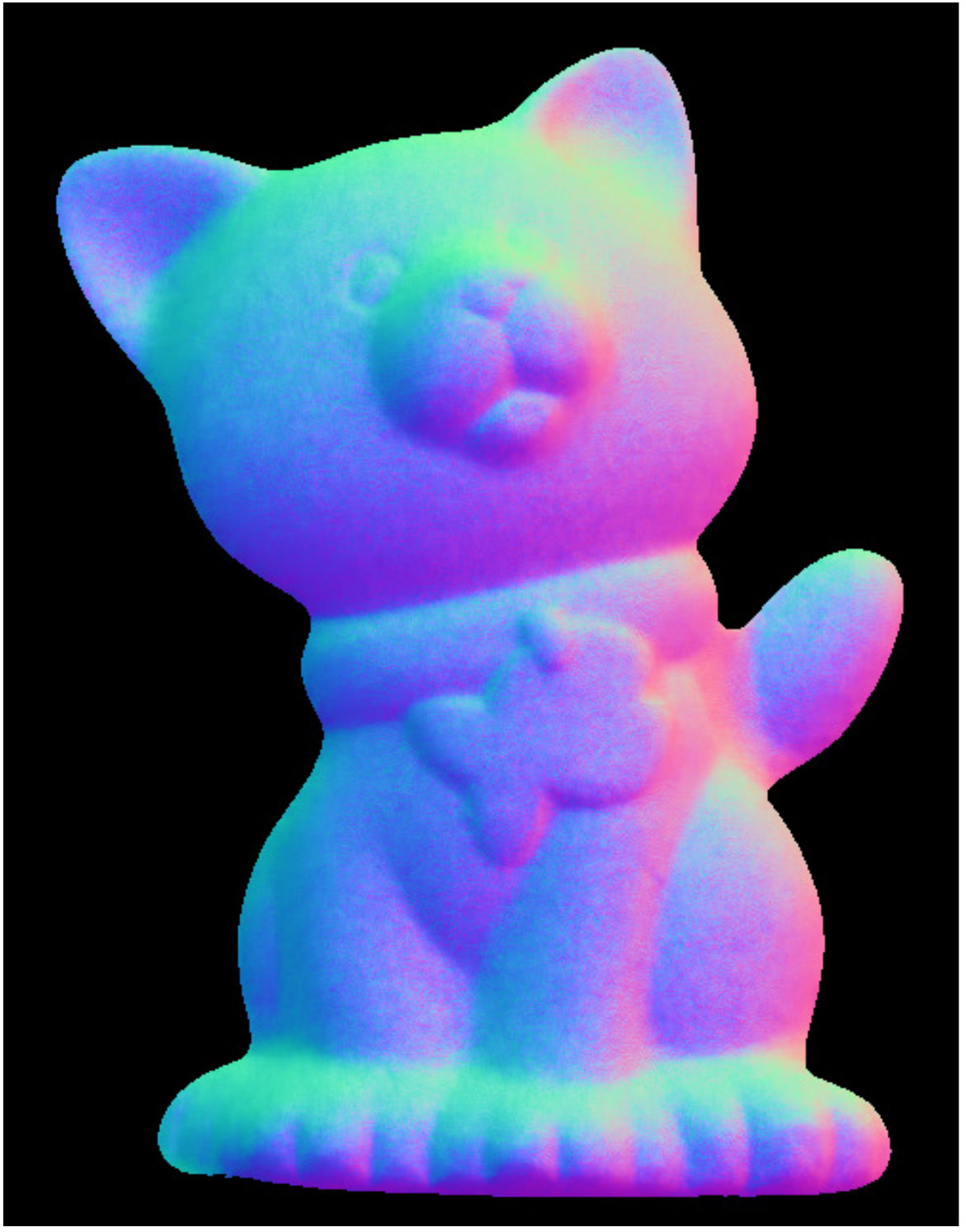}
  \caption{\textbf{DLNV}}
\end{subfigure}
\hspace{-2mm} 
\begin{subfigure}[b]{0.14\textwidth}
  \includegraphics[width=\textwidth]{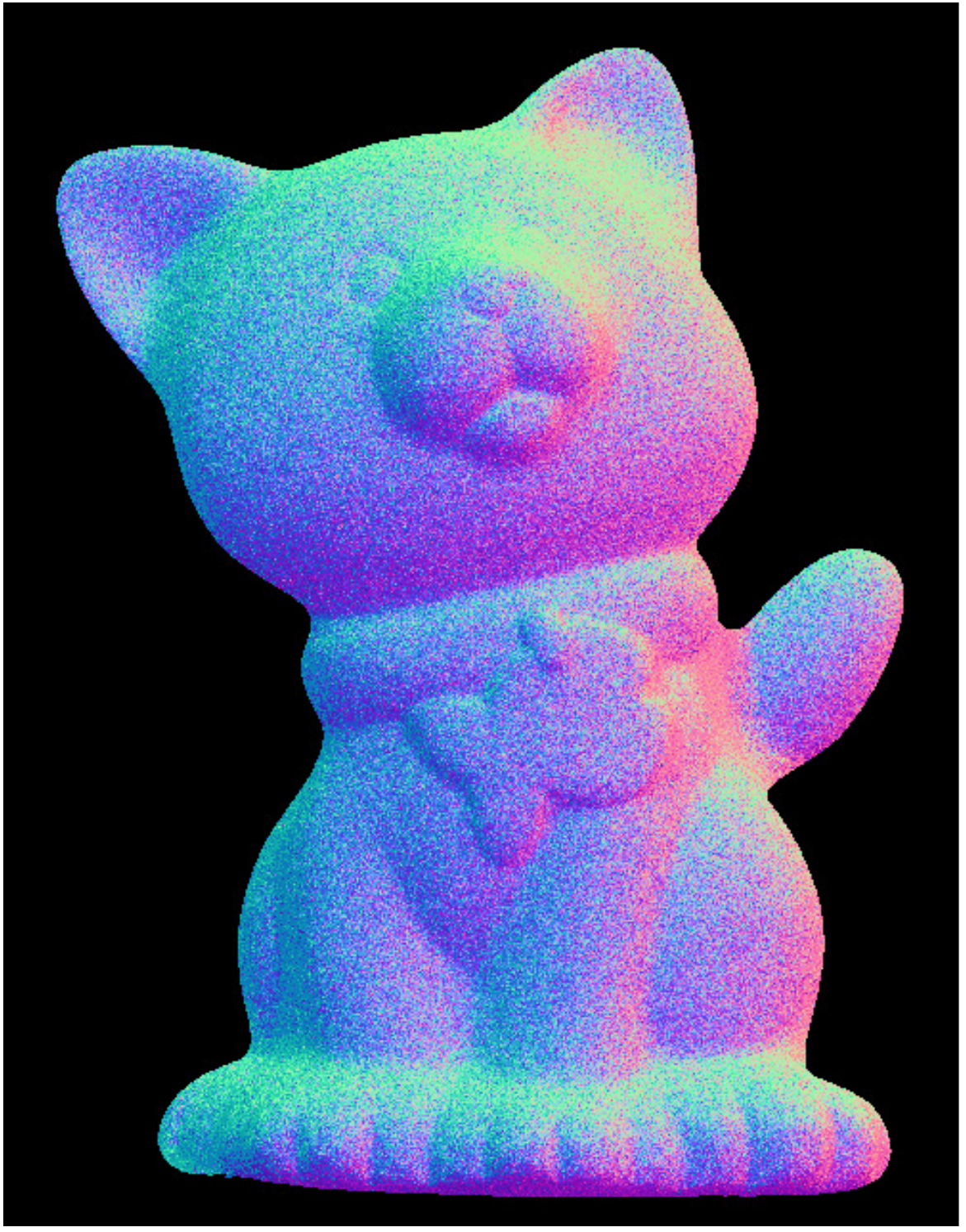}
  \caption{SR}
\end{subfigure}
\hspace{-2mm} 
\begin{subfigure}[b]{0.14\textwidth}
  \includegraphics[width=\textwidth]{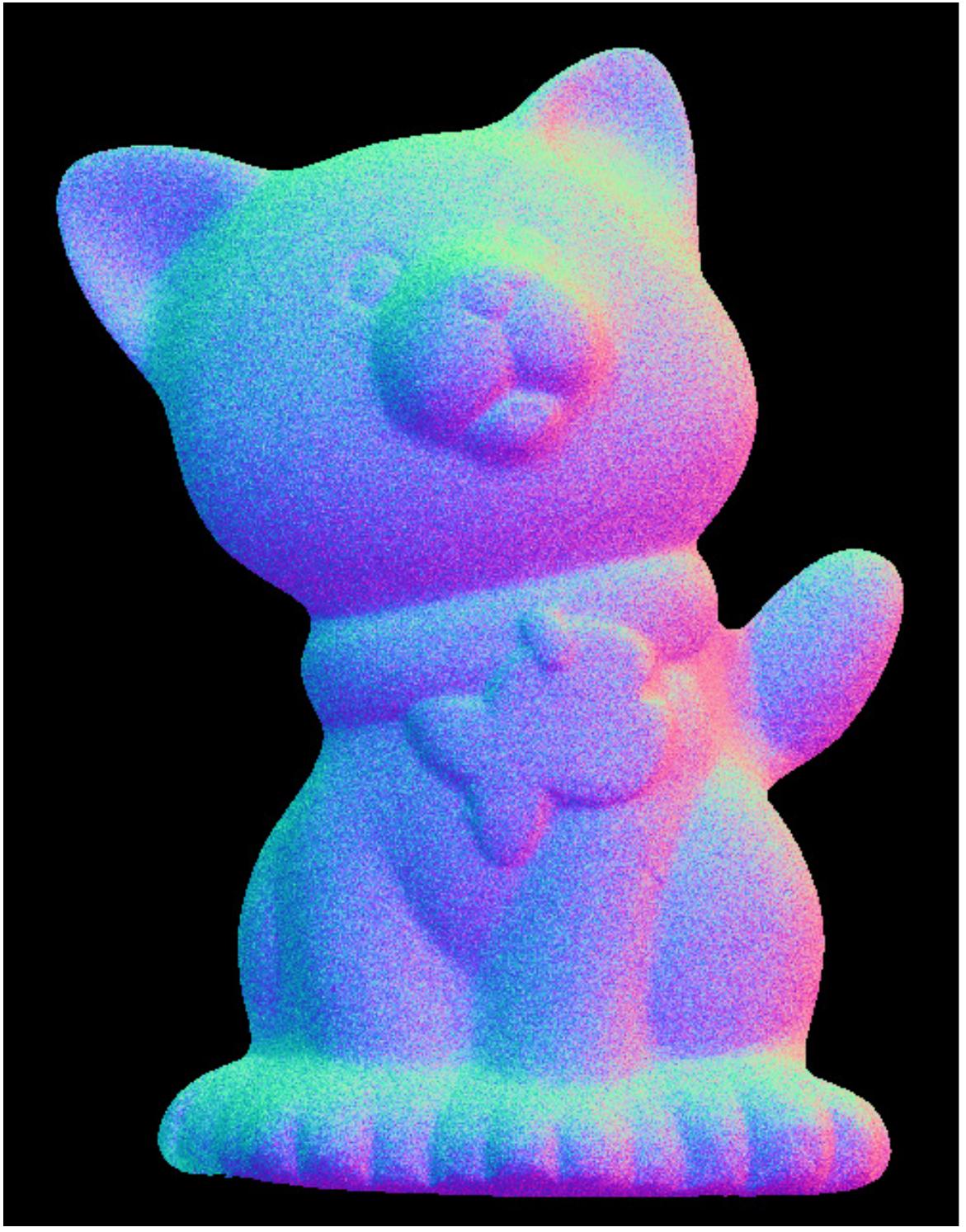}
  \caption{RPCA}
\end{subfigure}
\hspace{-2mm} 
\begin{subfigure}[b]{0.14\textwidth}
  \includegraphics[width=\textwidth]{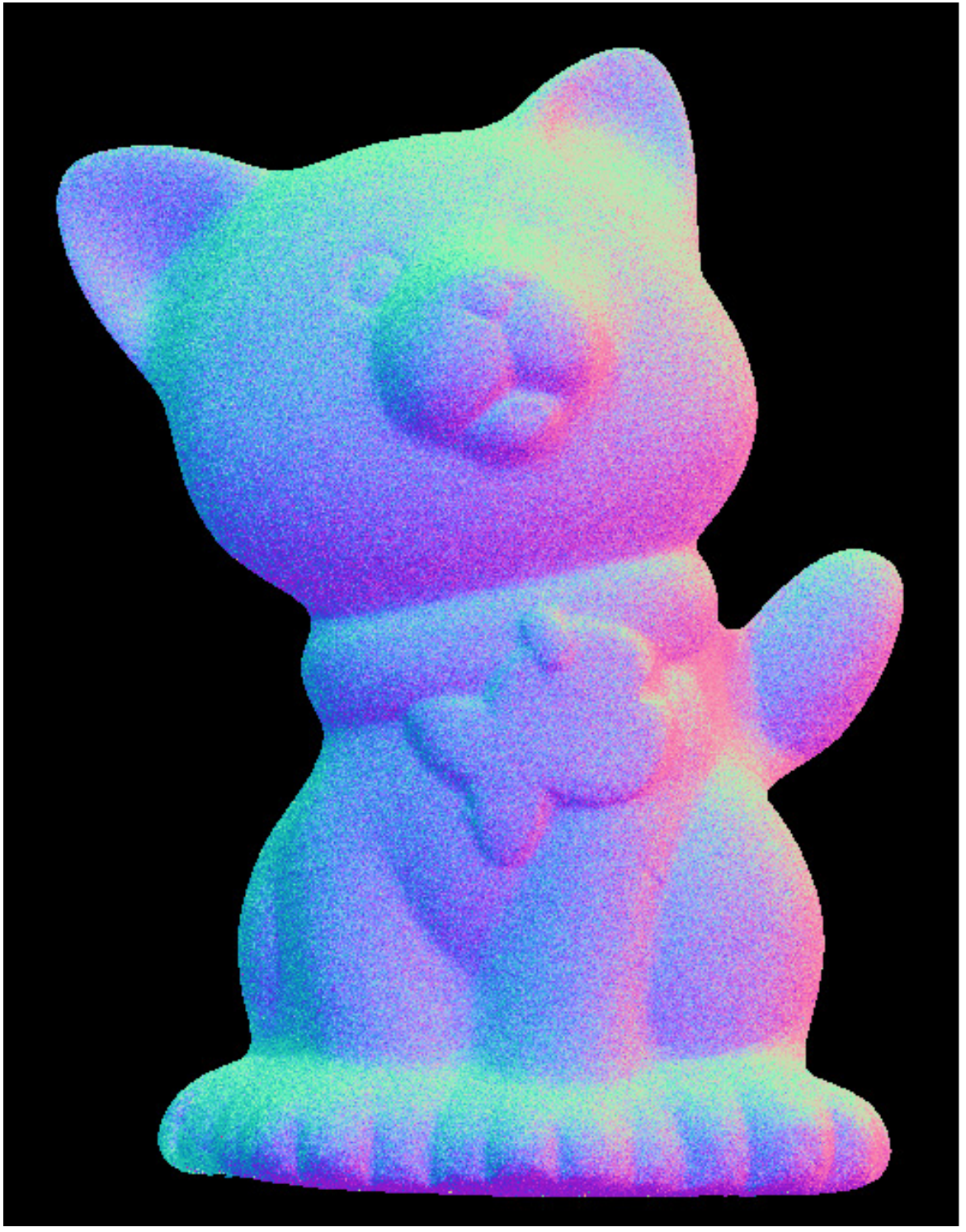}
  \caption{CBR}
\end{subfigure}
\hspace{-2mm} 
\begin{subfigure}[b]{0.14\textwidth}
  \includegraphics[width=\textwidth]{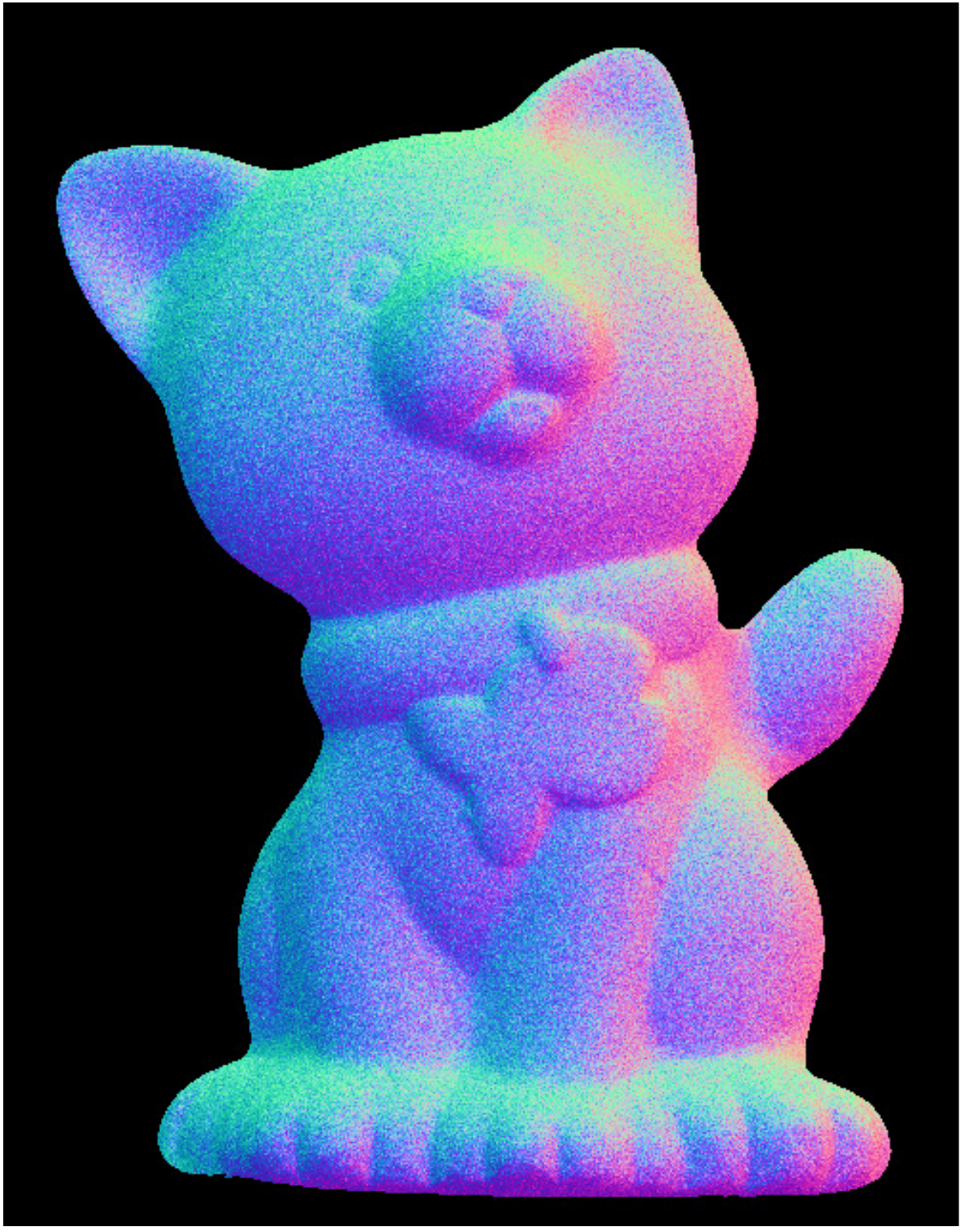}
  \caption{LS}
\end{subfigure}
\caption{Normal vector reconstructions for the Cat dataset with 20 images and \revised{Poisson noise added with resulting SNR 5 dB}.}
\label{fig:sweep_snr_20_na_0_cat_all_1_2}
\end{figure*}

\begin{figure*}[t!]
\centering
\begin{subfigure}[b]{0.14\textwidth}
  \includegraphics[width=\textwidth]{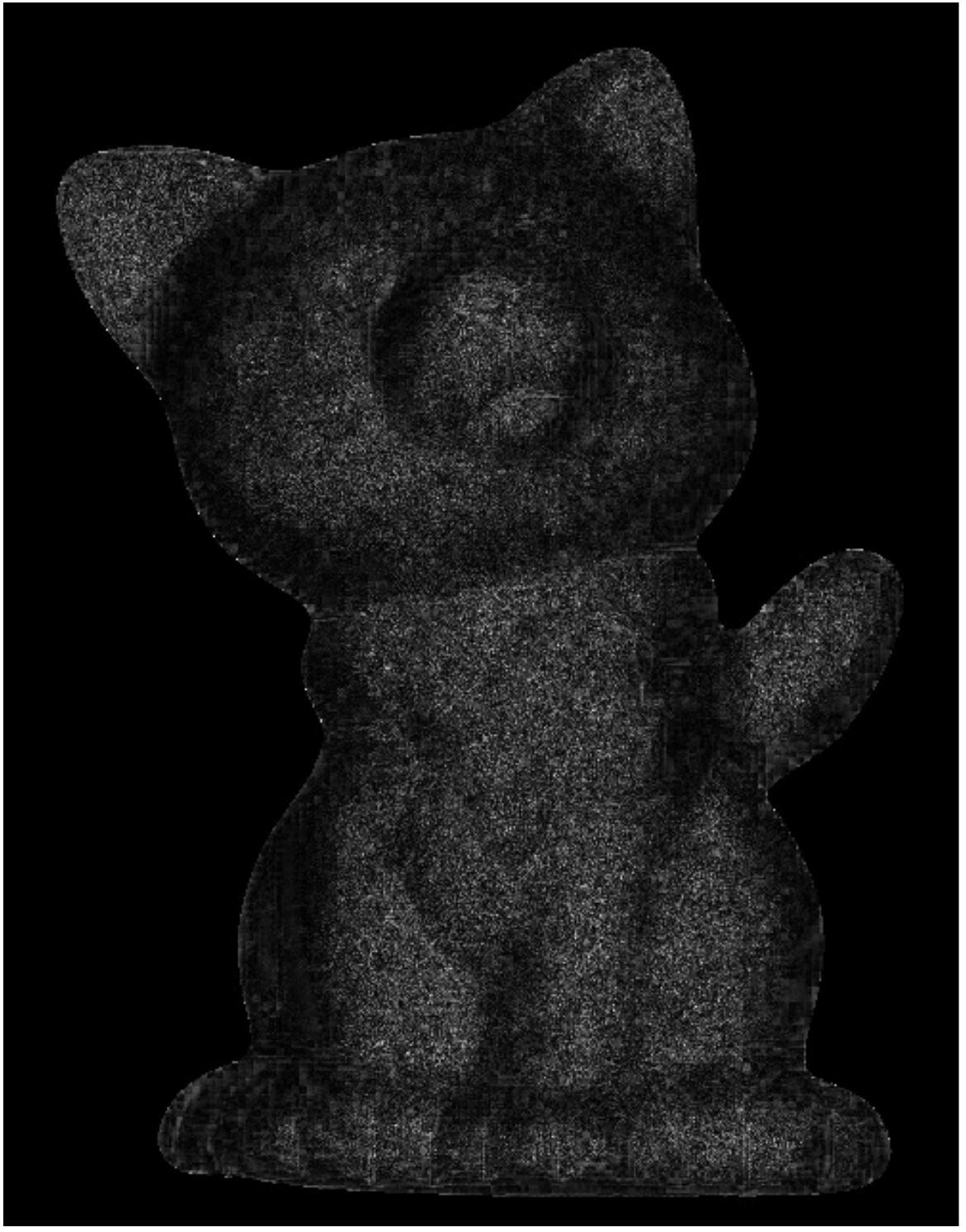}
  \caption{\textbf{PDLNV} (7.65)}
\end{subfigure}
\hspace{-2mm} 
\begin{subfigure}[b]{0.14\textwidth}
  \includegraphics[width=\textwidth]{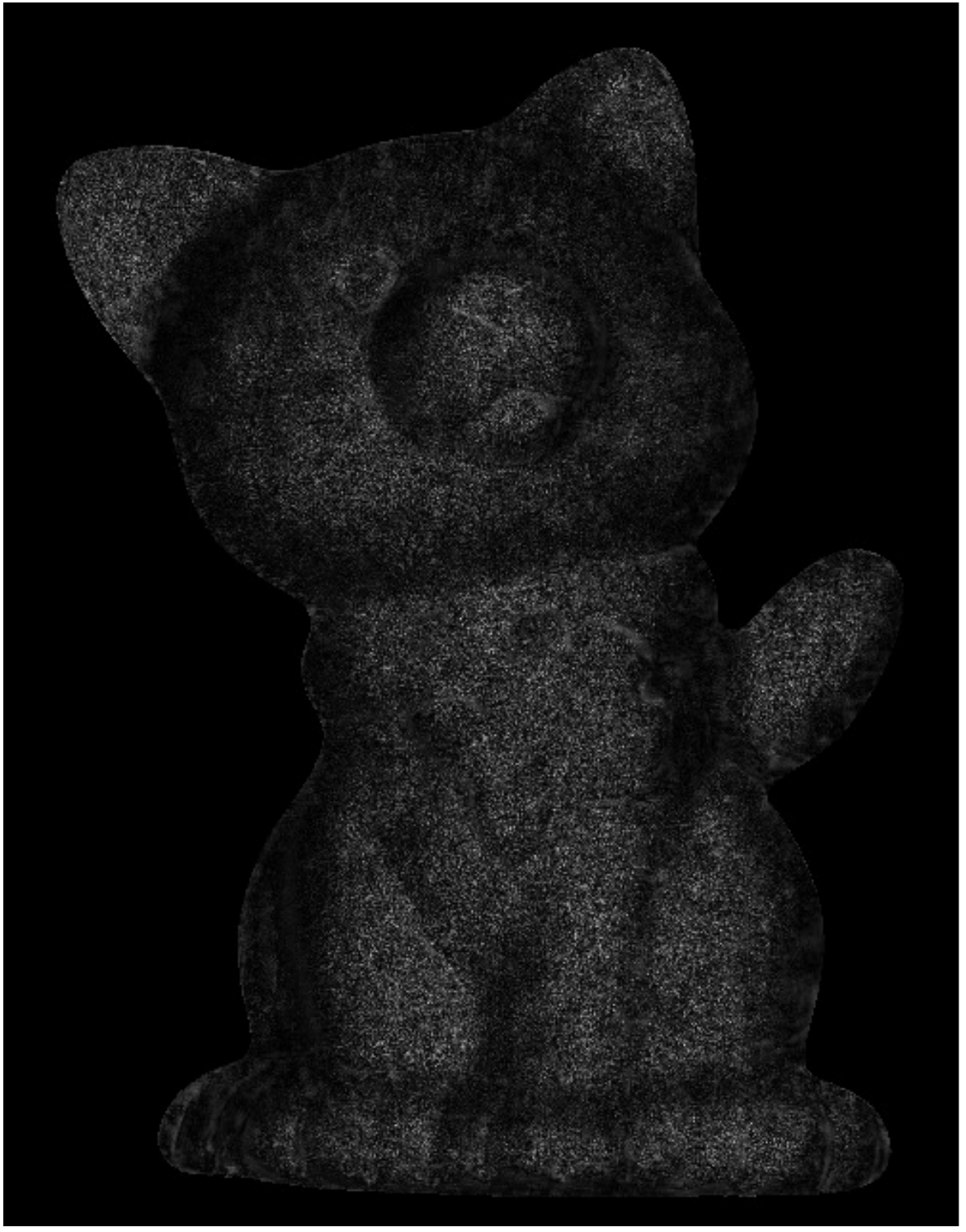}
  \caption{\textbf{DLNV} (6.64)}
\end{subfigure}
\hspace{-2mm} 
\begin{subfigure}[b]{0.14\textwidth}
  \includegraphics[width=\textwidth]{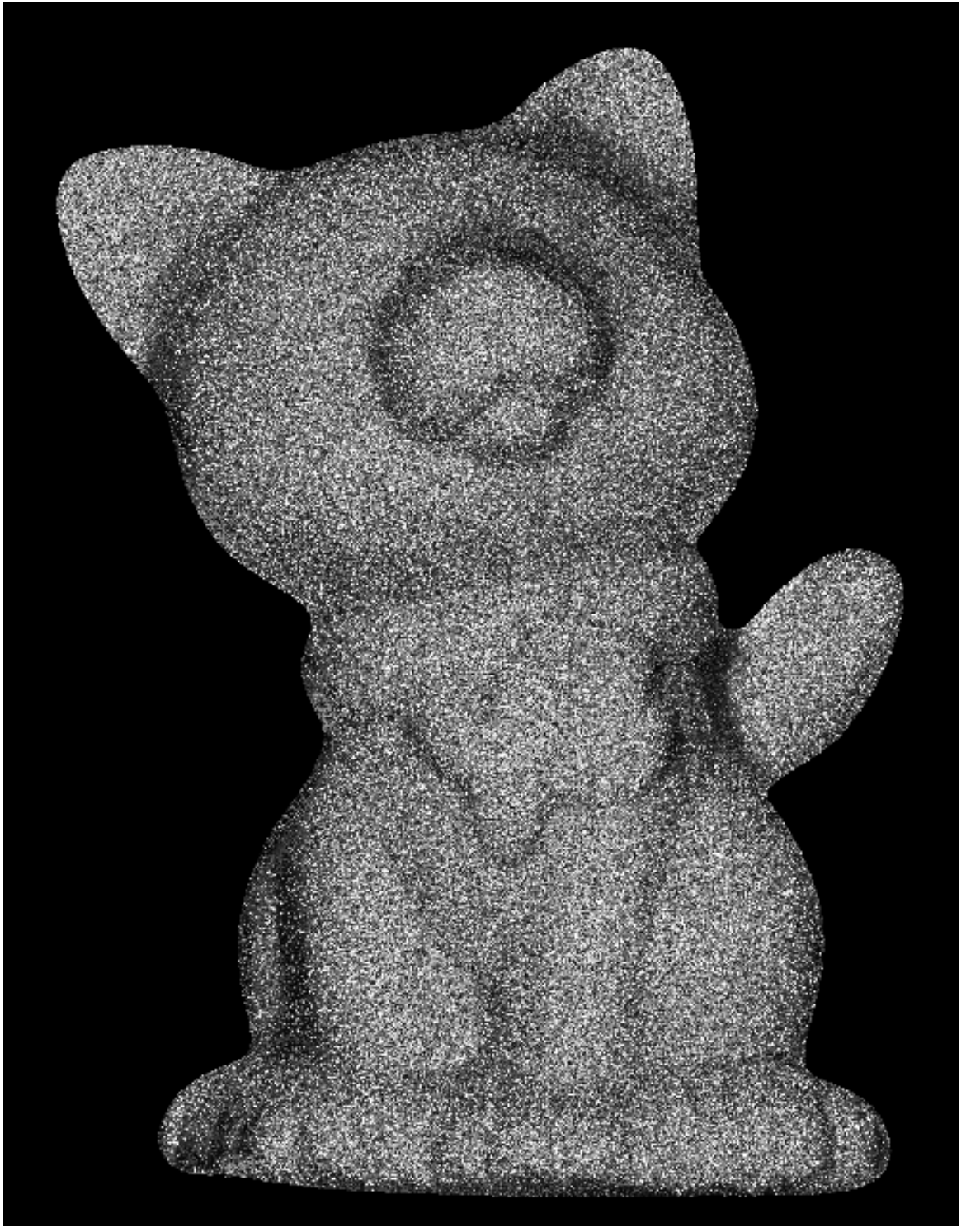}
  \caption{SR (24.46)}
\end{subfigure}
\hspace{-2mm} 
\begin{subfigure}[b]{0.14\textwidth}
  \includegraphics[width=\textwidth]{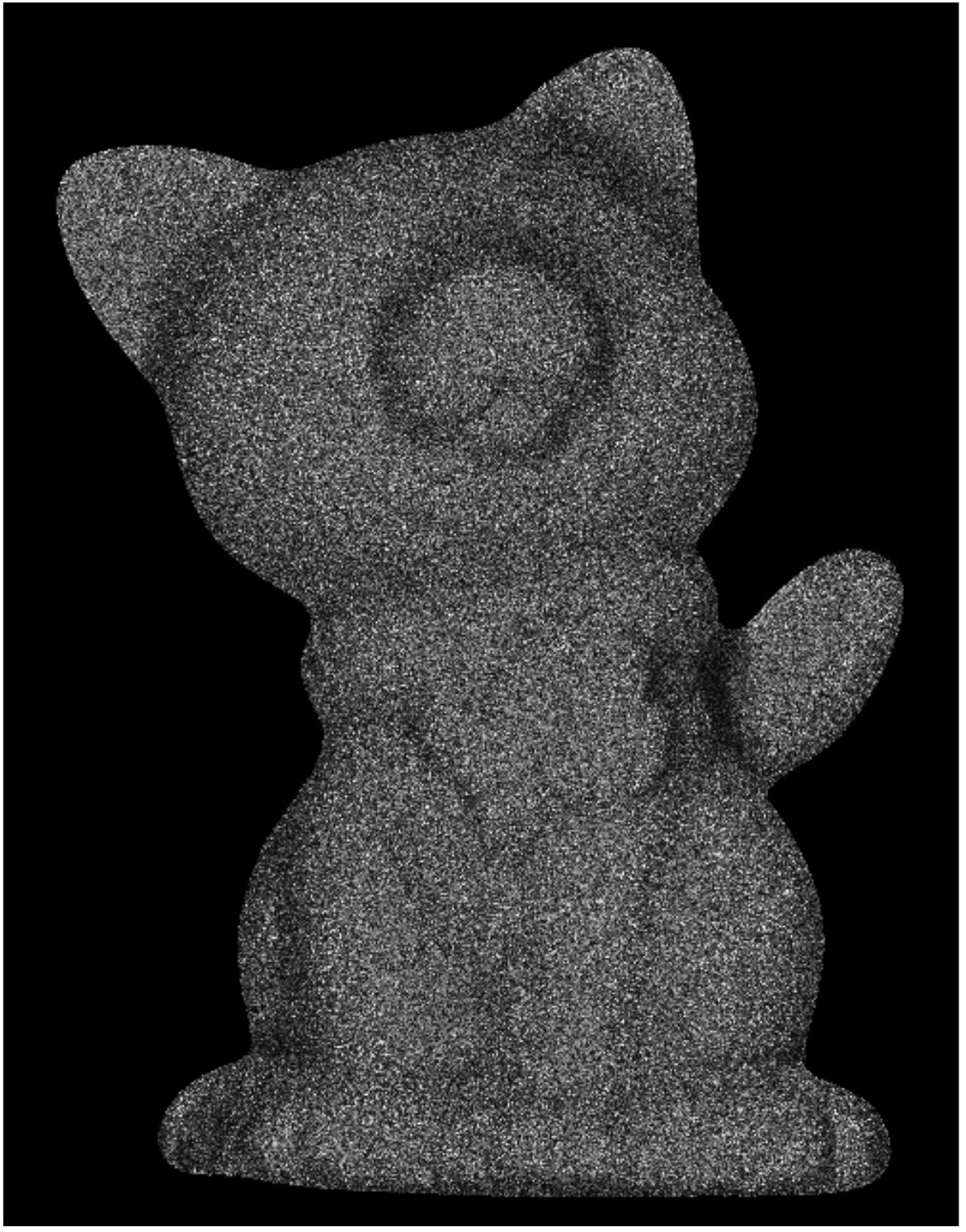}
  \caption{RPCA (17.35)}
\end{subfigure}
\hspace{-2mm} 
\begin{subfigure}[b]{0.14\textwidth}
  \includegraphics[width=\textwidth]{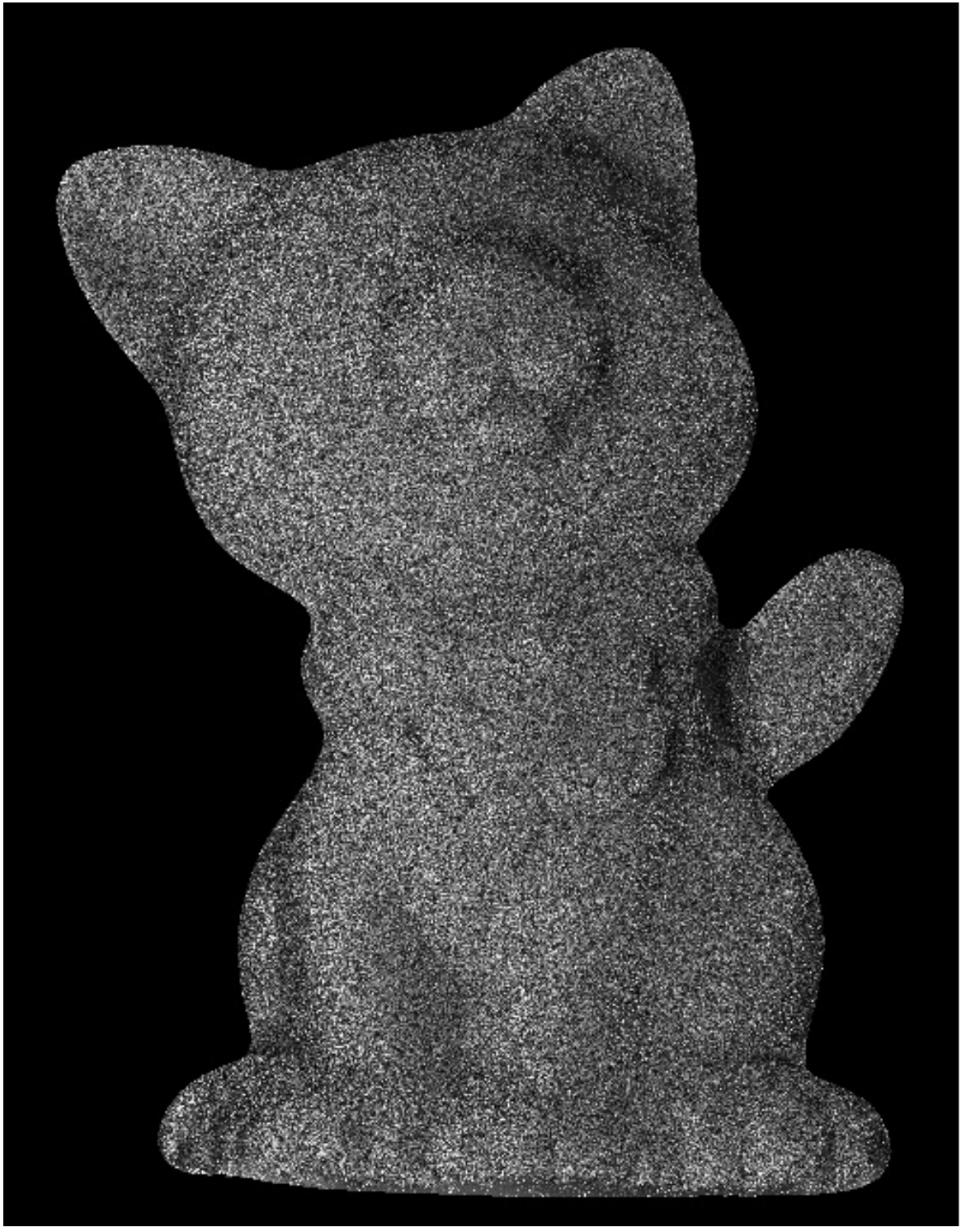}
  \caption{CBR (18.19)}
\end{subfigure}
\hspace{-2mm} 
\begin{subfigure}[b]{0.14\textwidth}
  \includegraphics[width=\textwidth]{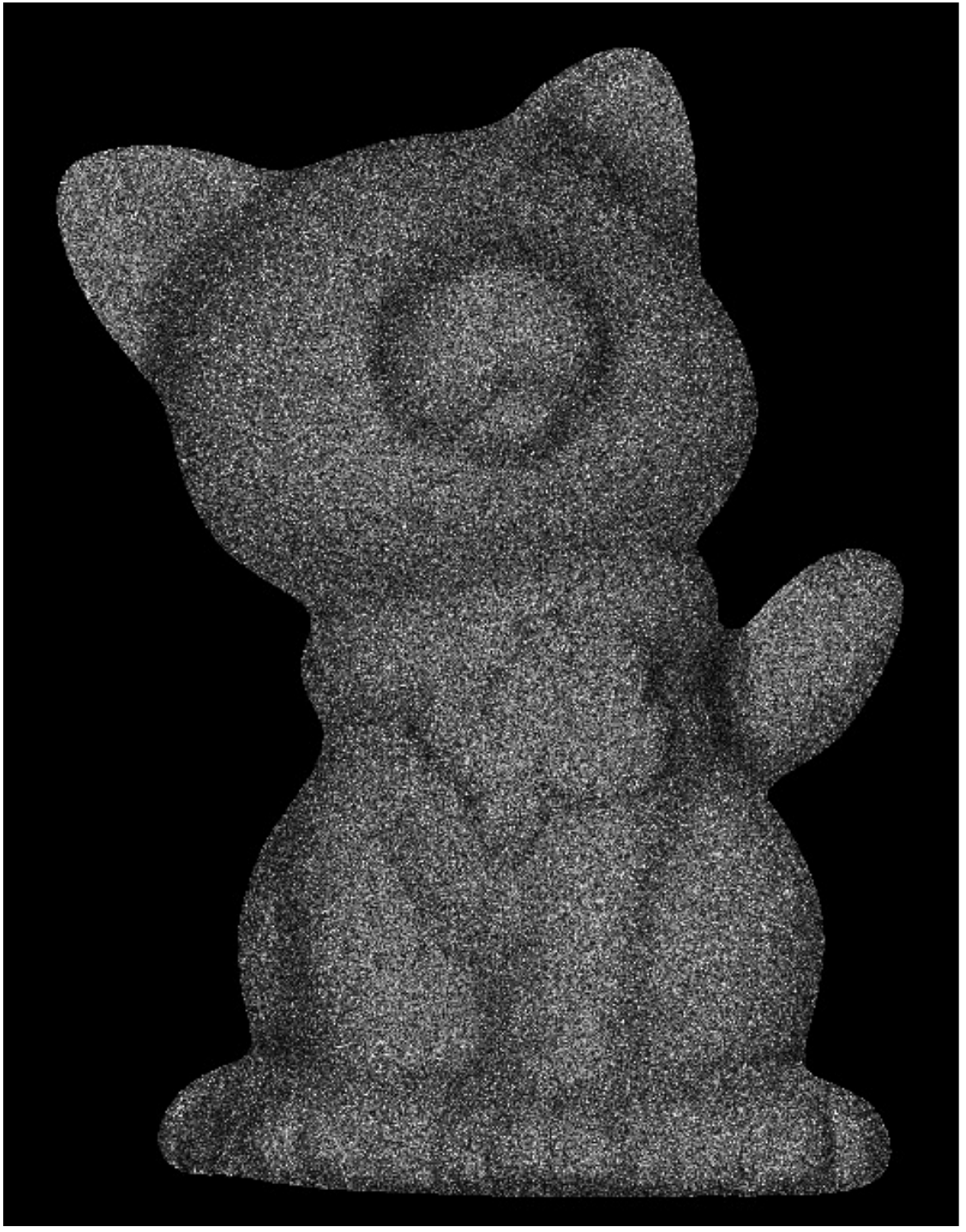}
  \caption{LS (17.87)}
\end{subfigure}
\begin{subfigure}[b]{0.04\textwidth} 
  \includegraphics[width=\textwidth]{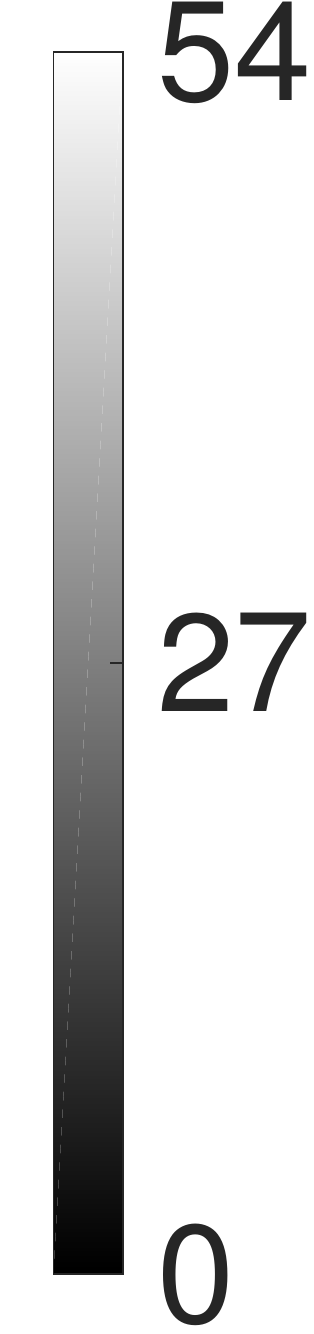}
  \caption*{}
\end{subfigure}
\caption{Normal vector error maps for the Cat dataset with 20 images and \revised{Poisson noise added with resulting SNR 5 dB. Mean angular errors (in degrees) for each reconstruction are shown in parentheses}.}
\label{fig:sweep_snr_20_na_0_cat_all_1_2_errors}
\end{figure*}

\begin{figure}[t!]
\centering
\begin{subfigure}[b]{0.09\textwidth}
  \includegraphics[height=1.75in]{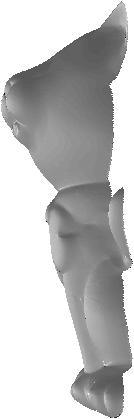}
  \caption{Ref.}
\end{subfigure}
\hspace{0.5mm}
\begin{subfigure}[b]{0.09\textwidth}
  \includegraphics[height=1.75in]{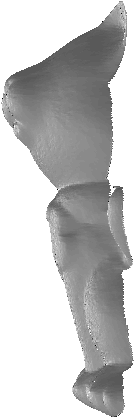}
  \caption{\textbf{PDLNV}}
\end{subfigure}
\hspace{0.5mm}
\begin{subfigure}[b]{0.09\textwidth}
  \includegraphics[height=1.75in]{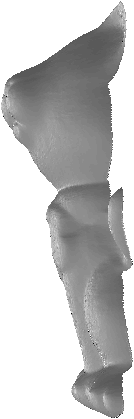}
  \caption{\textbf{DLNV}}
\end{subfigure}
\\[7pt]
\begin{subfigure}[b]{0.09\textwidth}
  \includegraphics[height=1.75in]{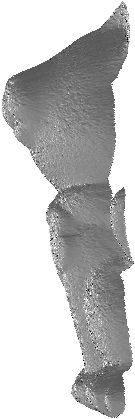}
  \caption{SR}
\end{subfigure}
\hspace{0.5mm}
\begin{subfigure}[b]{0.09\textwidth}
  \includegraphics[height=1.75in]{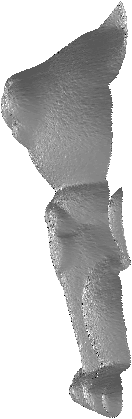}
  \caption{RPCA}
\end{subfigure}
\hspace{0.5mm}
\begin{subfigure}[b]{0.09\textwidth}
  \includegraphics[height=1.75in]{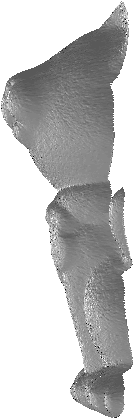}
  \caption{LS}
\end{subfigure}
\hspace{0.5mm}
\begin{subfigure}[b]{0.09\textwidth}
  \includegraphics[height=1.75in]{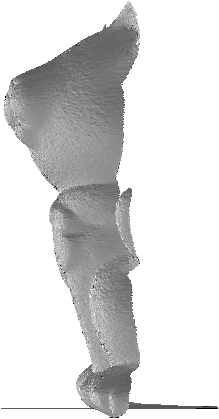}
  \caption{CBR}
\end{subfigure}
\caption{Surfaces computed from the estimated normal vectors of the Cat dataset with 20 images and \revised{Poisson noise added with resulting SNR 5 dB}.}
\label{fig:sweep_snr_20_na_0_cat_all_1_2_surface}
\end{figure}

In addition to the DiLiGenT dataset, we also consider the dataset\footnote{The data can be found at \url{http://vision.seas.harvard.edu/ qsfs/Data.html}.} from \cite{xiong2015shading}. This dataset contains images of several real objects with no corresponding \revised{ground truth} normal vectors. To obtain reference (ground truth) normal vectors for this dataset, we assume the objects are Lambertian. While this assumption does not hold exactly, the objects are matte in appearance and thus nearly Lambertian. We compute the reference normal vectors by applying the standard least squares method \eqref{eq:ls} to the raw images.

Our motivation for considering this (approximately) Lambertian dataset is as follows. Even when additional noise is added, the primary challenge with the DiLiGenT datasets is dealing with the fundamentally non-Lambertian properites of the data (specularities, shadows, etc.) As such, our experiments thus far do not necessarily evaluate the ability of each method to estimate a Lambertian surface in the presence of noise, despite the fact that the majority of the methods we are investigating are based on a Lambertian model. Therefore, in this section we assume our data is Lambertian, add corruptions, and then evaluate the ability of each method to reject the corruptions and produce normal vectors that agree with the underlying Lambertian model.

Figure~\ref{fig:sweep_snr_20_na_0_hippo_all} plots the mean angular errors of the estimated normal vectors for the Hippo dataset as a function of SNR. For high SNR, the errors approach zero, as expected since the uncorrupted data is Lambertian. However, in the high SNR regime, the proposed dictionary learning-based approaches are significantly more robust to imperfections compared to existing approaches.

We also evaluate the qualitative performance of each method. Figure~\ref{fig:sweep_snr_20_na_0_cat_all_1_2} shows the reference normal vectors for the Cat dataset (computed from the uncorrupted data using the least squares method) together with the normal vectors estimated by each method from data corrupted by Poisson noise with 5 dB SNR. Figure~\ref{fig:sweep_snr_20_na_0_cat_all_1_2_errors} shows the error maps for the estimated normal vectors with respect to the reference normal vectors. Clearly the proposed dictionary learning-based methods produce much more accurate normal vectors compared to the existing methods. 

Figure~\ref{fig:sweep_snr_20_na_0_cat_all_1_2_surface} plots the surfaces computed for the normal vectors from Figure~\ref{fig:sweep_snr_20_na_0_cat_all_1_2} using the method outlined in \cite{simchony1990}. Qualitatively, we see that the surfaces computed from the dictionary learning-based methods are much smoother and more accurate representations of the actual surface. In contrast, the surfaces obtained from the existing methods, though they preserve the general shape of the surface, are quite rough and/or contain significant spike artifacts.

\subsection{Algorithm Properties}\label{sec:alg_props}

\begin{figure*}[t!]
\centering
\begin{subfigure}[b]{0.3\textwidth}
  \includegraphics[width=\textwidth]{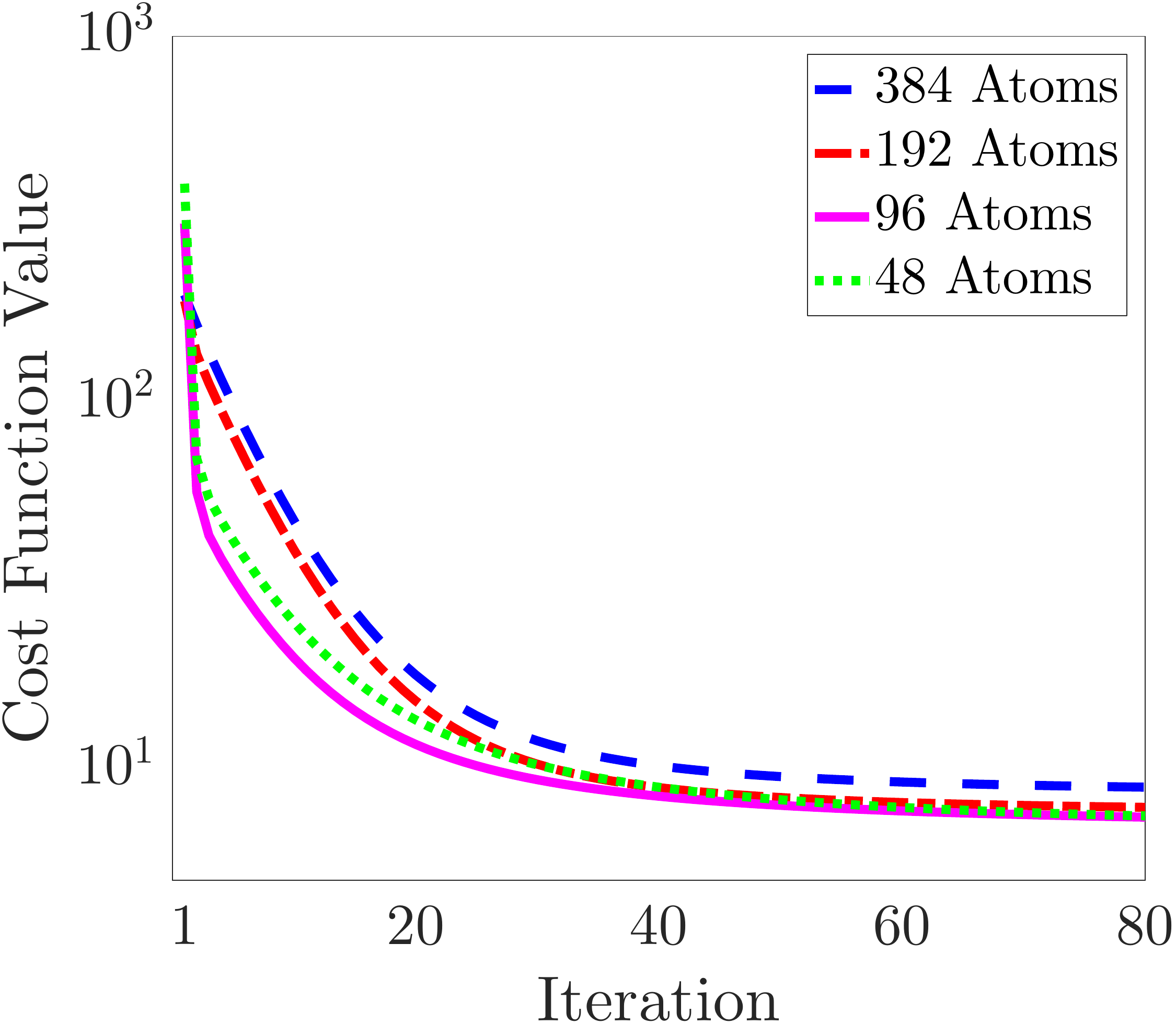}
  \caption{PDLNV cost function.}
  \label{fig:alg_cost}
\end{subfigure}
~
\begin{subfigure}[b]{0.3\textwidth}
  \includegraphics[width=\textwidth]{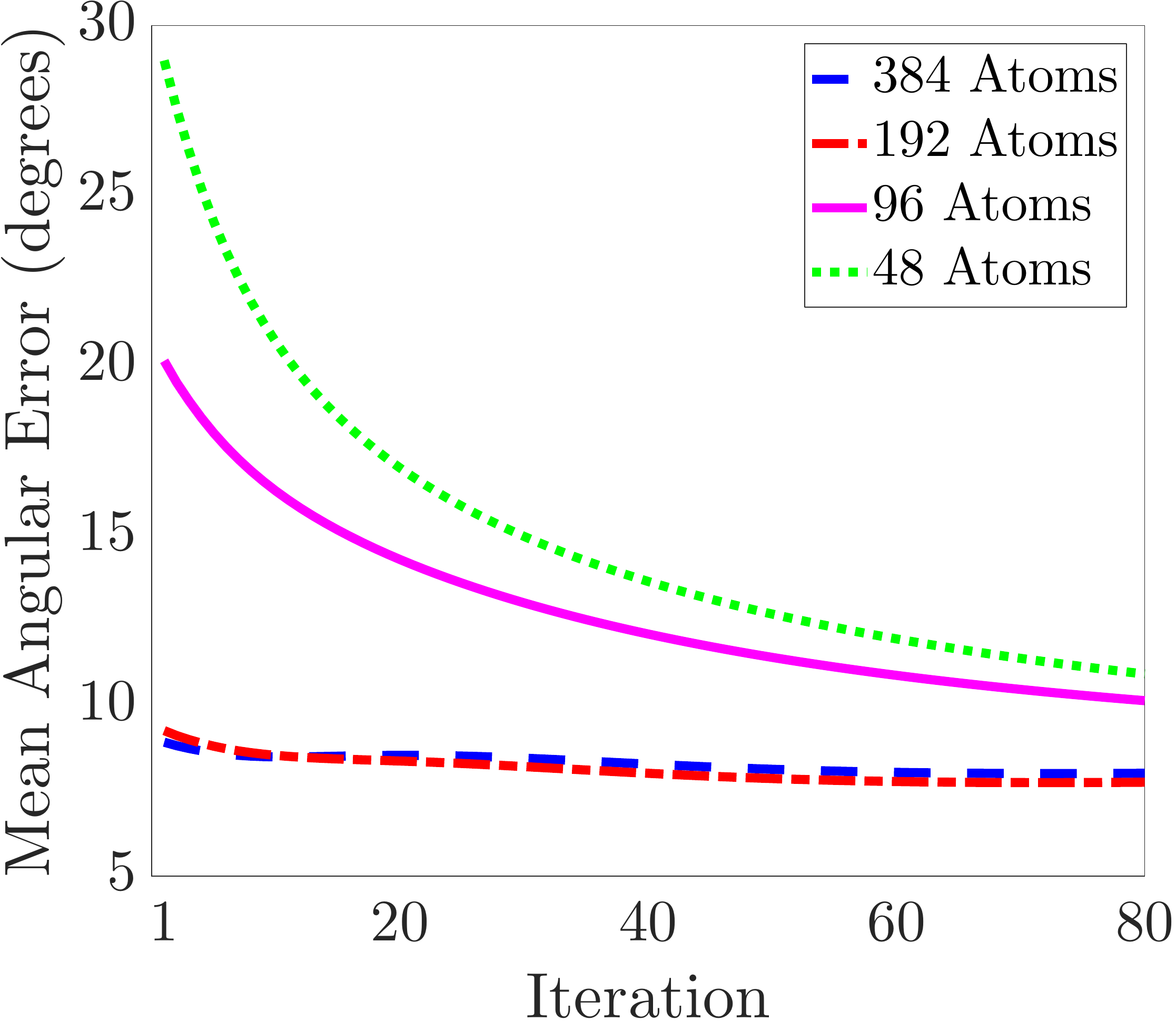}
  \caption{Normal vector angular errors.}
  \label{fig:alg_error}
\end{subfigure}
~
\begin{subfigure}[b]{0.3\textwidth}
  \includegraphics[width=\textwidth]{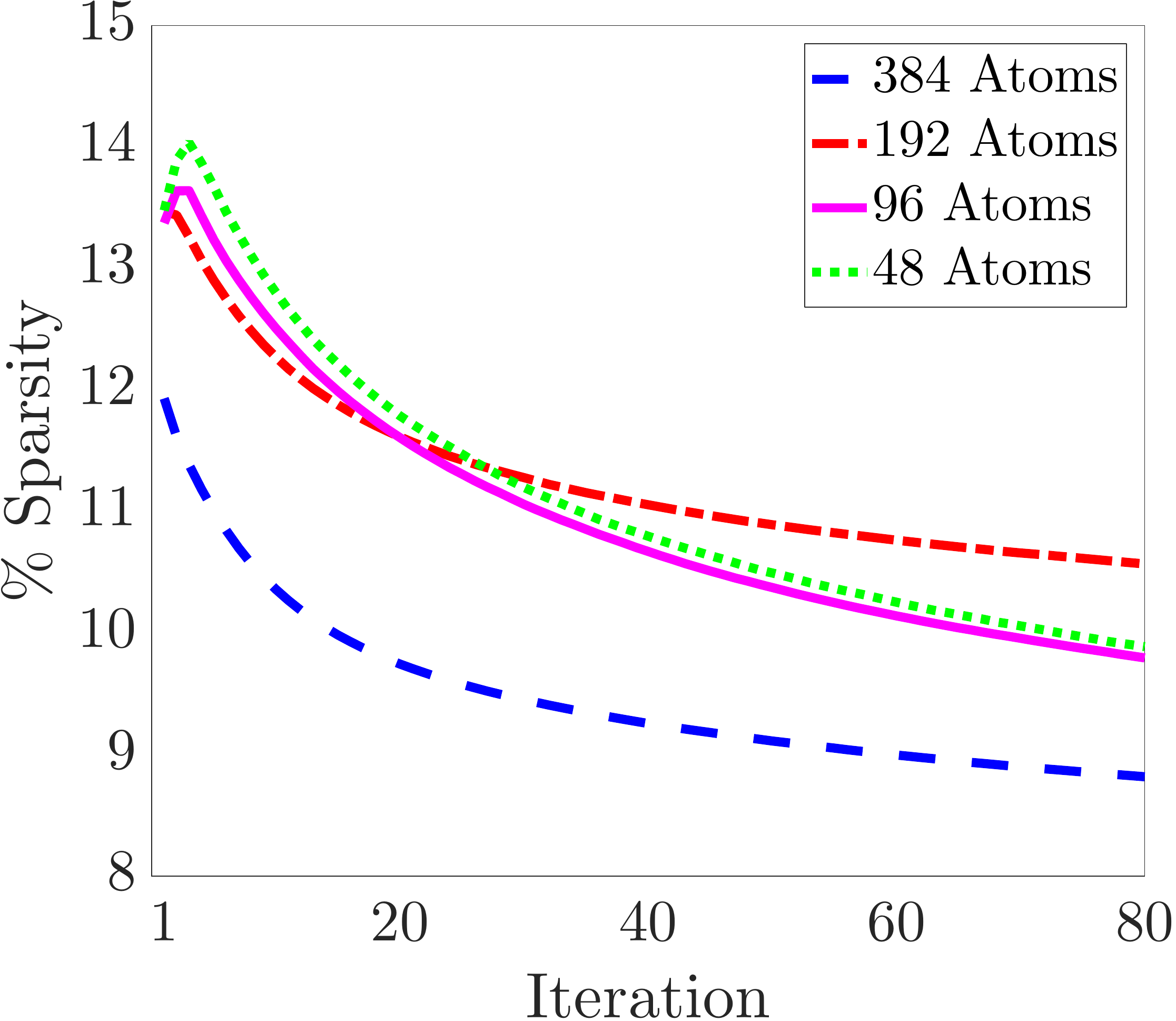}
  \caption{Sparsity of $B$.}
  \label{fig:alg_sparsity}
\end{subfigure}
\caption{Cost function, normal vector angular errors, and sparsity of the sparse coding matrix $B$ for the PDLNV method with $p = 2$ applied to the DiLiGenT Cat dataset with 20 images and 20 dB Poisson noise for several different dictionary sizes.}
\label{fig:sweepp_snrinf_nim96_Dpot1s_properties}
\end{figure*}

\begin{figure}[t!]
\centering
\includegraphics[width=0.45\textwidth]{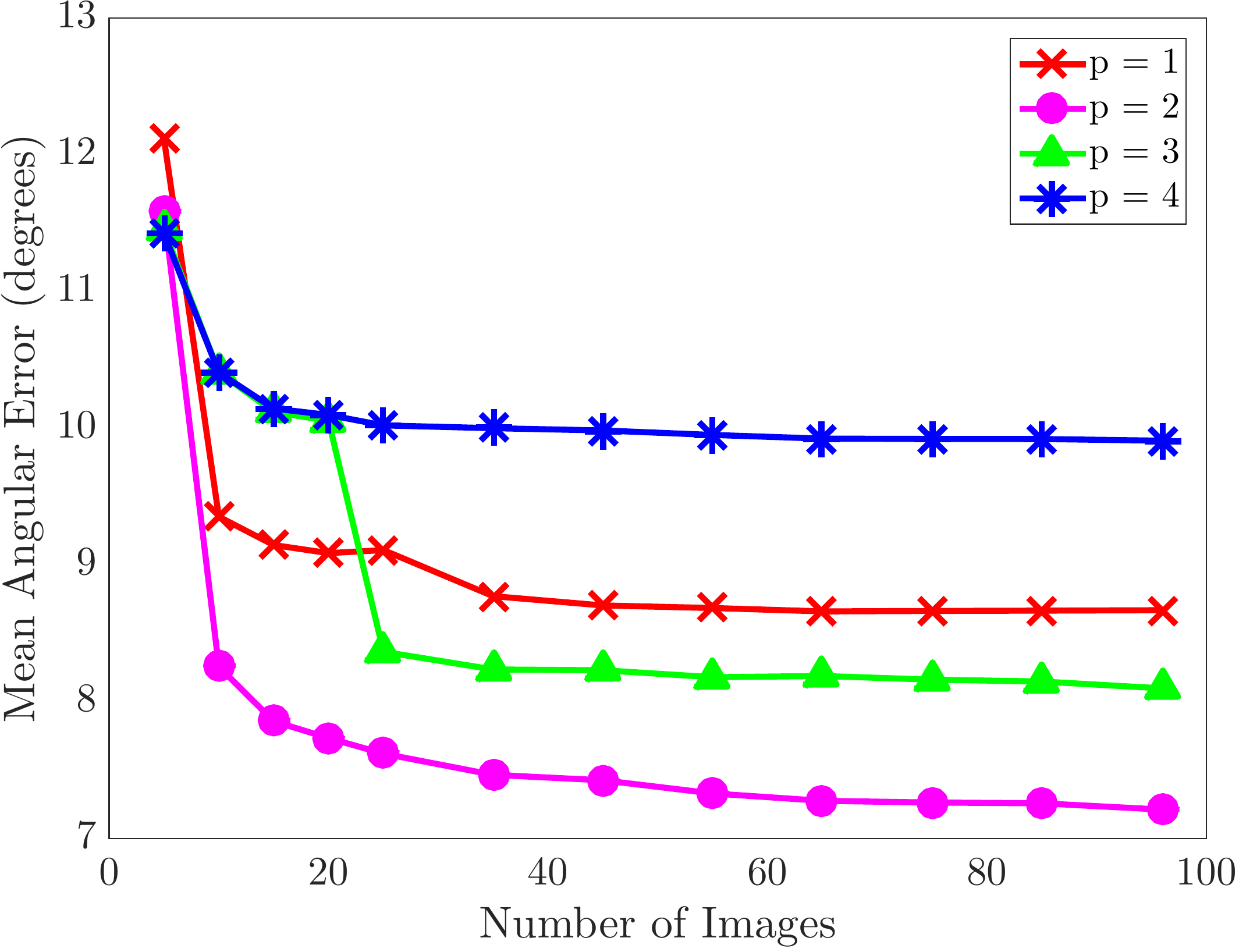}
\caption{Mean angular error of the estimated normal vectors for PDLNV with multiple values of $p$ on the DiLiGenT Pot1 dataset as a function of number of images used.}
\label{fig:sweep_p_snrinf_nim96_Dpot1s}
\end{figure}

Finally, we investigate the properties of our proposed dictionary learning-based methods and how the various model parameters affect the results. Figure~\ref{fig:sweepp_snrinf_nim96_Dpot1s_properties} shows the per-iteration properties of the PDLNV method with $p = 2$ from a representative trial on the DiLiGenT Cat dataset. Figure~\ref{fig:alg_cost} plots the cost function at each iteration, and Figure~\ref{fig:alg_error} shows the corresponding mean angular error of the normal vector estimates. While the cost is guaranteed to decrease at each iteration, angular error is not guaranteed to decrease and could increase. Empirically we observe, however, that angular error typically decreases with iteration. Figure~\ref{fig:alg_sparsity} plots the sparsity (percentage of nonzero elements) of the sparse coding matrix $B$ at each iteration. The sparsity of $B$ can be changed by varying the regularization parameter $\mu$. Empirically, we observe that sparsity values around 10\% often yield good results. Each plot in Figure~\ref{fig:sweepp_snrinf_nim96_Dpot1s_properties} includes multiple curves for several different dictionary sizes (number of columns). Of particular interest is how the size of the dictionary affects mean angular error. As Figure~\ref{fig:alg_error} illustrates, larger dictionaries typically perform better than smaller, undercomplete dictionaries. However, we do not observe a significant boost in performance when the dictionary becomes overcomplete. Note that $8 \times 8 \times 3$ dictionary atoms were used for PDLNV, so a size of 192 corresponds to a square dictionary. 

\revised{
\begin{figure}[t!]
\centering
\begin{subfigure}[b]{0.23\textwidth}
  \includegraphics[width=\textwidth]{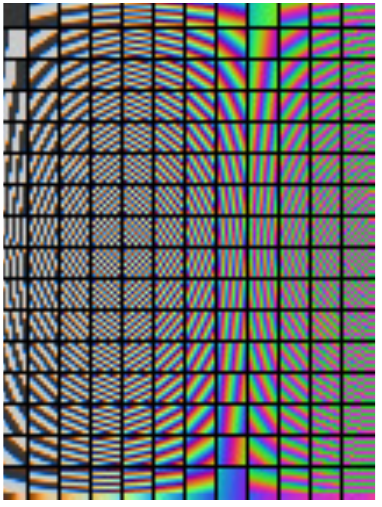}
  \caption{Initial DCT dictionary.}
\end{subfigure}
~
\begin{subfigure}[b]{0.23\textwidth}
  \includegraphics[width=\textwidth]{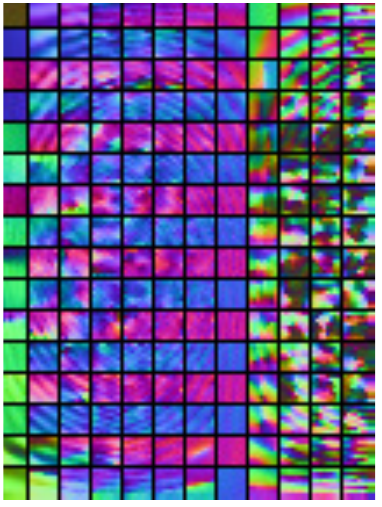}
  \caption{Final learned dictionary.}
\end{subfigure}
~
\caption{Initial and final learned dictionaries for the PDLNV method with $p=2$ applied to the full DiLiGenT Pot1 dataset.}
\label{fig:sweepp_snrinf_nim96_Dpot1s_dicts}
\end{figure}
}

For the experiments presented in this work, we terminated PDLNV after 50 iterations and terminated \revised{DLNV} after 20 iterations. At each iteration, we performed one pass over the columns of $(D,B)$ and performed 25 proximal gradient steps updating $n$ or $v$.

We next illustrate the effect of varying parameter $p$ on the performance of PDLNV. Figure~\ref{fig:sweep_p_snrinf_nim96_Dpot1s} shows the angular error of the estimated normal vectors as a function of the number of images in the DiLiGenT Pot1 dataset for several values of $p$. As illustrated, for this dataset $p = 2$ is the optimal choice. In general, we observe that $p = 2$ typically produces good results, but in some cases (e.g., the DiLiGenT Harvest dataset) $p = 3$ performs better.

Figure~\ref{fig:sweepp_snrinf_nim96_Dpot1s_dicts} illustrates the initial (DCT) and final (learned) dictionaries produced by PDLNV on the full DiLiGenT Pot1 dataset. \revised{Each dictionary atom is an $8 \times 8 \times 3$ tensor, so we visualize each atom as a single color image, analogous to how we visualized the normal vector maps from Figure~\ref{fig:pot2_normals}. The learned dictionary exhibits interesting structure. Some atoms have learned structure across all three normal vector dimensions, while other atoms have learned structure in one or two dimensions and are constant in the other dimension(s). Note that this behavior has emerged organically---the dictionary learning methods automatically adapt to the underlying structure on a per-dataset basis.}

\section{Conclusion} \label{sec:conclusion}
In this work, we proposed \revised{two} methods for applying dictionary learning to photometric stereo. \revised{Both methods seek} to represent the reconstructed normal vectors as sparse with respect to an adaptive dictionary. We showed through extensive numerical studies that our proposed methods are significantly more robust than existing methods in the high-noise regime while preserving accuracy in the low-noise regime. Dictionary learning is a general purpose adaptive regularization framework, and, as such, it could be coupled with more complex reflectance models from the photometric stereo literature to further improve reconstruction quality. We plan to investigate this line of inquiry in future work.

\bibliographystyle{IEEEtran}
\bibliography{PSviaDL}

\begin{IEEEbiography}[{\includegraphics[width=1in,height=1.25in,clip,keepaspectratio]{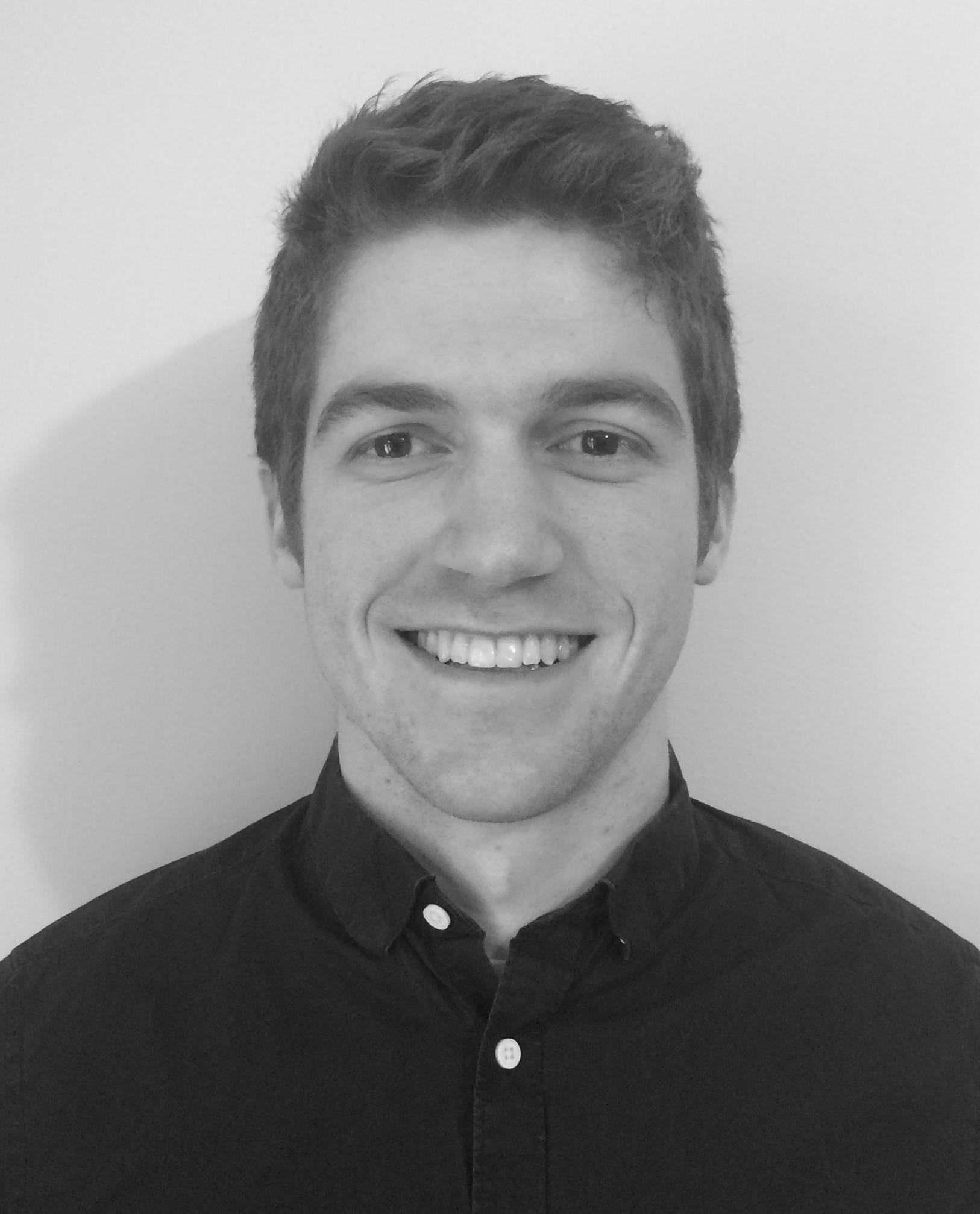}}]
{Andrew J. Wagenmaker} received the B.S.E. and M.S. degrees in electrical engineering from the University of Michigan, Ann Arbor, MI, USA, in 2016 and 2017, respectively. He is currently pursuing a Ph.D. in computer science at the University of Washington, Seattle, WA, USA, where he is supported by a National Science Foundation Graduate Fellowship. His research interests are broadly in the area of statistical machine learning.
\end{IEEEbiography}

\begin{IEEEbiography}[{\includegraphics[width=1in,height=1.25in,clip,keepaspectratio]{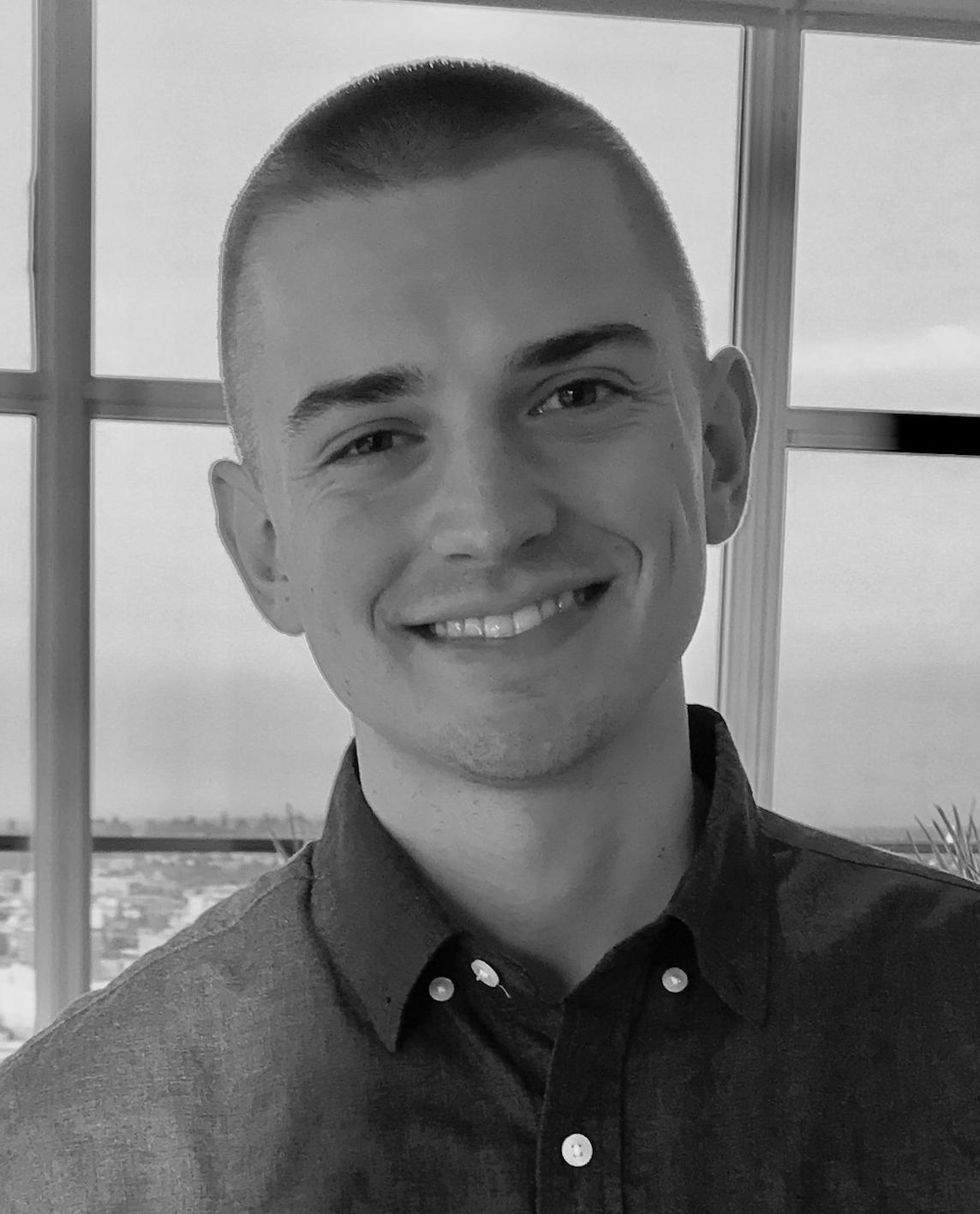}}]
{Brian E. Moore} received the B.S. degree in electrical engineering from Kansas State University, Manhattan, KS, USA, in 2012, and the M.S. and Ph.D. degrees in electrical engineering from the University of Michigan, Ann Arbor, MI, USA, in 2014 and 2018, respectively. He is currently the co-founder of Voxel51, LLC, Ann Arbor, MI, USA, a startup founded in late 2016 focused on cutting edge problems in computer vision and machine learning with applications in public safety and automotive sensing. His research interests include efficient algorithms for large-scale machine learning problems with a particular emphasis on computer vision applications.
\end{IEEEbiography}

\begin{IEEEbiography}[{\includegraphics[width=1in,height=1.25in,clip,keepaspectratio]{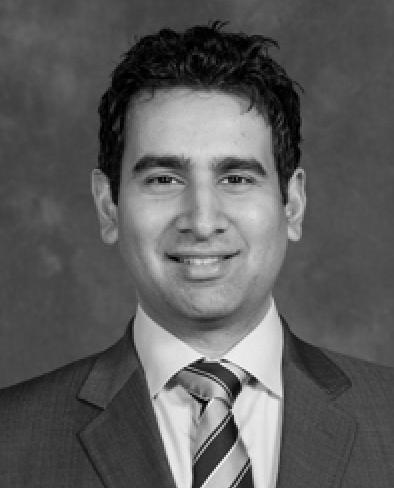}}]
{Raj Rao Nadakuditi} received the Ph.D. degree in 2007 from the Massachusetts Institute of Technology, Cambridge, MA, USA, and the Woods Hole Oceanographic Institution, Woods Hole, MA, USA. He is currently an Associate Professor in the Department of Electrical Engineering and Computer Science, University of Michigan, Ann Arbor, MI, USA. His research focuses on developing theory for random matrices for applications in signal processing, machine learning, queuing theory, and scattering theory. He received an Office of Naval Research Young Investigator Award in 2011, an Air Force Office of Scientific Research Young Investigator Award in 2012, the Signal Processing Society Young Author Best Paper Award in 2012, and the DARPA Young Faculty Award in 2014.
\end{IEEEbiography}

\end{document}